\documentclass[10pt]{article}

\usepackage[left=3cm,right=3cm,top=2.25cm,bottom=2.25cm]{geometry} 
\usepackage{bbm}
\usepackage{epsfig,graphicx}
\usepackage{xcolor}

\newcommand{\thetaTV}{\theta^\textrm{TV}}
\newcommand{\hatthetaTV}{\thetaTV}%\widehat{}
\newcommand{\rd}{\mathrm{d}}
\newcommand{\iid}{\mathrm{iid}}
\newcommand{\cov}{\text{cov}}

\makeatletter
\newcommand{\objref}[4]{\def\obj@rg{#4}%
  #1\ifx\obj@rg\empty#2\else#3\xspace\ref{#4}--\fi\ref}
\makeatother

\usepackage{amsmath,amsfonts,mathrsfs,amsfonts, bbold,bm,blindtext}%,color
\usepackage{algorithmicx,algorithm,algpseudocode}
\allowdisplaybreaks
\usepackage{setspace}
\usepackage{multirow}
\usepackage{xr-hyper}
\usepackage{hyperref}

\usepackage[normalem]{ulem} %otherwise \emph is underlined in references!

\begin{document}
	
\title{ \bf Reinforcement learning and Bayesian data assimilation for model-informed precision dosing in oncology}
\author{ Corinna Maier$^{1,2}$, Niklas Hartung$^{1}$, Charlotte Kloft$^{3}$,\\[1ex] Wilhelm Huisinga$^{1,\dagger,\ast}$, and Jana de Wiljes$^{1,\dagger}$}
\date{\relax}
\maketitle

\noindent
%\textbf{Word count: }\\
%\textbf{Version: \today}\\
$^1$Institute of Mathematics, University of Potsdam, Germany\\[1ex]
$^2$Graduate Research Training Program PharMetrX: Pharmacometrics \& Computational Disease Modelling, Freie Universit\"at Berlin and University of Potsdam, Germany\\[1ex]
$^3$Department of Clinical Pharmacy and Biochemistry, Institute of Pharmacy, Freie Universit\"at Berlin, Germany\\[1ex]
$^\ast$ corresponding author\\
$^\dagger$ these authors contributed equally to this work\\
Institute of Mathematics, Universit\"at Potsdam\\
Karl-Liebknecht-Str. 24-25, 14476 Potsdam/Golm, Germany\\
Tel.: +49-977-59 33, Email: huisinga@uni-potsdam.de, wiljes@uni-potsdam.de 
%\subsubsection*{Word count: 4037}

% -----------------------------------------------------------------------------------------------------------------------------------------------------
\subsubsection*{Conflict of Interest/Disclosure}
% -----------------------------------------------------------------------------------------------------------------------------------------------------
CK and WH report research grants from an industry consortium (AbbVie Deutschland GmbH \& Co. KG, AstraZeneca, Boehringer Ingelheim Pharma GmbH \& Co. KG, Gr\"unenthal GmbH, F. Hoffmann-La Roche Ltd., Merck KGaA and Sanofi) for the PharMetrX program. In addition CK reports research grants from the Innovative Medicines Initiative-Joint Undertaking (`DDMoRe'), from H2020-EU.3.1.3 ('FAIR') and Diurnal Ltd. All other authors declare no competing interests for this work.

% -----------------------------------------------------------------------------------------------------------------------------------------------------
\subsubsection*{Funding information}
% -----------------------------------------------------------------------------------------------------------------------------------------------------
\begin{itemize}
\item Graduate Research Training Program PharMetrX: Pharmacometrics \& Computational Disease Modelling, Berlin/Potsdam, Germany,
\item Deutsche Forschungsgemeinschaft (DFG) - SFB1294/1 - 318763901 (associated project).
\item Deutsche Forschungsgemeinschaft and Open Access Publishing Fund of University of Potsdam.
\end{itemize}

% -----------------------------------------------------------------------------------------------------------------------------------------------------
\subsubsection*{Keywords}
% -----------------------------------------------------------------------------------------------------------------------------------------------------
Data assimilation, Oncology, Chemotherapy, Therapeutic Drug Monitoring, Model-Informed Precision Dosing

% -----------------------------------------------------------------------------------------------------------------------------------------------------
% Diskussionspunkte
% -----------------------------------------------------------------------------------------------------------------------------------------------------

\newpage	
% =====================================================================================
% Abstract
% =====================================================================================
\begin{abstract} 
Model-informed precision dosing (MIPD) using therapeutic drug/biomarker monitoring offers the opportunity to significantly improve the efficacy and safety of drug therapies.
Current strategies comprise model-informed dosing tables or are based on maximum a-posteriori estimates. 
These approaches, however, lack a quantification of uncertainty and/or consider only part of the available patient-specific information. 
We propose three novel approaches for MIPD employing Bayesian data assimilation (DA) and/or reinforcement learning (RL) to control neutropenia, the major dose-limiting side effect in anticancer chemotherapy.
These approaches have the potential to substantially reduce the incidence of life-threatening grade 4 and subtherapeutic grade 0 neutropenia compared to existing approaches. 
We further show that RL allows to gain further insights by identifying patient factors that drive dose decisions.
Due to its flexibility, the proposed combined DA-RL approach can easily be extended to integrate multiple endpoints or patient-reported outcomes, thereby promising important benefits for future personalized therapies.
% 152 words / 150 words
\end{abstract}
%to discuss 
%wording: investigate (propose?)
%The integration of DA and RL allows to make use of the individualized model parameter uncertainties (DA) to improve the individual dose adjustments (RL). 		
		
% =====================================================================================
\section{INTRODUCTION}
% =====================================================================================

%general setting when standard treatment protocols are not successful
Personalized dosing offers the opportunity to improve safety and efficacy of drugs beyond the current practice \cite{Peck2015}. 
This is particularly crucial for drugs that exhibit narrow therapeutic indices relative to the variability between patients. 
%In this context, adherence to standardized treatment protocols may lead to serious adverse events in some patients and subtherapeutic exposure in others. 
Patient-specific dose adaptations during ongoing treatments are, however, difficult due to the need to integrate multiple sources of information, the lack of precise guidelines for dose adaptations in the label and limited time resources \cite{deJonge2005,Darwich2017}.
%107 words
	
%chemotherapy - neutropenia
A particularly critical case is cytotoxic anticancer chemotherapy with neutropenia as major dose-limiting toxicity \cite{Crawford2004}. 
Patients with severe neutropenia experience a drastic reduction of neutrophil granulocytes and are thus highly susceptible to potentially life-threatening infections.
Depending on the lowest neutrophil concentration (nadir), the different grades $g$ of neutropenia range from no neutropenia ($g=0$) to life-threatening ($g=4$) \cite{CTCAE}.
At the same time, neutropenia serves as a surrogate for efficacy (in terms of median survival) \cite{Cameron2003,DiMaio2005,DiMaio2006}. 
Neutrophil counts can therefore be used as a biomarker to guide dosing \cite{Wallin2009a,Hansson2010,Netterberg2017a}. 
%98 words
	
%current practice - Paclitaxel - CEPAC-TDM -Joerger	
In this article, we consider paclitaxel-induced neutropenia as an illustrative and therapeutically relevant application. Paclitaxel is used as first-line treatment against non-small-cell lung cancer in platinum-based combination therapy \cite{Peters2012}. 
The standard dosing of paclitaxel is based on the patient's body surface area (BSA). 
To individualize treatment, a dosing table based on sex, age, BSA, drug exposure and toxicity was developed  \cite{Joerger2012a} and evaluated in a clinical trial (hereafter ``CEPAC-TDM study'') \cite{Joerger2016a}. 
%Although this study showed that occurrence of another side effect could be decreased, it did not succeed in reducing severe neutropenia (grade 4) \cite{Joerger2016a}.
%105 words

%literature review: MIPD
Model-informed precision dosing (MIPD) describes approaches for dose individualization that take into account prior knowledge on the drug-disease-patient system and associated variability, e.g., from a nonlinear mixed effects (NLME) analysis as well as patient-specific therapeutic drug/biomarker monitoring (TDM) data \cite{Keizer2018}.
A popular approach is based on maximum a-posteriori (MAP) estimation \cite{Sheiner1979,Wallin2009,Bleyzac2001}, which infers the individual model parameters of the pharmacokinetic/pharmacodynamic (PK/PD) model. 
MAP-based outcomes are typically evaluated with respect to a utility function or a target concentration to determine the next dose (MAP-guided dosing) \cite{WallinPage2009,Wallin2009}. 
The definition of a target concentration or utility function is, however, difficult since in many therapies rather subtherapeutic or toxic \textit{ranges} are known. 
For therapeutic ranges MAP-guided dosing is inaccurate \cite{Holford2018}, since only a point estimate is used, neglecting associated uncertainties \cite{Maier2020}.
%176
%(concentrations at the end of the range are evaluated the same as those in the middle of the range)

%literature review: DA
Recently, we have shown that Bayesian data assimilation (DA) approaches provide more informative clinical decision support, fully exploiting patient-specific information \cite{Maier2020}.
DA allows for individualized uncertainty quantification, which is a necessity (i) to integrate both, safety and efficacy aspects into the objective function of finding the optimal dose, or (ii) to compute the probability of being within/outside the target range. 
However, optimizing across a whole therapy time frame can be hard and potentially too costly for real-time decision support. 
%113

%literature review: RL
Reinforcement learning (RL) has been applied to various fields in health care, however, mainly focusing on clinical trial design \cite{Zhao2011,Yu2019}, and only few studies relate to optimal dosing in a PK/PD context \cite{Escandell2014,YauneyShah2018}.
In model-based RL, it is learned how to act best in an uncertain environment using model simulations. 
A key aspect of learning is to make successively use of knowledge already acquired, while also exploring yet unknown sequences of actions.
The result is typically a decision tree (or some functional relationship). 
In other words, the physician's decision is supported via a pre-calculated, extensive and detailed look-up table without additional online computation. 
So far, RL approaches in health care are limited to rather simple exploration strategies (so-called $\epsilon$-greedy approaches) with one time step ahead approximations of the look-up table (Q-learning) \cite{Yu2019}.

%and surprisingly consider only approaches, which are intended for tasks with no clearly defined or foreseeable end (in contrast to many therapies with finite treatment time frame).
%110 words

%What do we do here
In this article, we demonstrate how DA and RL can be very beneficially exploited to develop new approaches to MIPD. 
The first approach, referred to as DA-guided dosing, improves existing online MIPD by integrating model uncertainties into the dose selection process. 
For the second approach (RL-guided dosing) we propose Monte Carlo tree search (MCTS) in conjunction with the upper confidence bound applied to trees (UCT) \cite{Silver2016,Rosin2011} as sophisticated learning strategy. %instead of simple random exploration of possible actions.
The third approach combines DA and RL (DA-RL-guided) to make full use of patient TDM data and to provide a flexible, interpretable and extendable framework. 
We compared the three proposed approaches with current dosing strategies in terms of dosing performance and their ability to provide insights into the factors driving dose selection.

% =====================================================================================
\section{METHODS}
% =====================================================================================

%We outline a general framework for dose individualization that is subsequently used to introduce three novel approaches for MIPD. 
%To illustrate this framework, we use the example of paclitaxel-induced neutropenia as dose-limiting side effect.
%34
%\\

\noindent 
We consider a single dose every three weeks schedule for paclitaxel-based chemotherapy, usually termed a cycle $c=1,\dots,C$, for a total of six cycles ($C=6$). 
We denote the decision time point for the dose of cycle $c$ by $T_c$, and assume $T_1=0$ (therapy start). 
For dose selection, the physician has different sources of information available, such as the patient's covariates `cov' (sex, age, etc), the treatment history (drug, dosing regimen, etc), TDM data related to PK/PD (drug concentrations, response, toxicity, etc). 
Despite these multiple sources of information, it remains a partial and imperfect information problem, as only noisy measurements of few quantities of interest at certain time points are available. 
%(e.g., measurements taken outside the expected nadir time window to infer the neutropenia grade).
MIPD aims to provide decision support by linking prior information on the drug-patient-disease system with patient-specific TDM data.  
%190

The \textit{standard dosing} for 3-weekly paclitaxel, as applied in the CEPAC-TDM study arm A, is $200\,\text{mg}/\text{m}^2 \, \text{BSA}$ and a 20\,\% dose reduction if neutropenia grade 4 ($g_c=4$) was observed  \cite{Joerger2016a}.
The aforementioned dosing table (termed \textit{PK-guided dosing} \cite{Joerger2012a}) was evaluated in study arm B, see Section~S~3.
%additionally taking into account the patient's sex, age, and drug exposure
For dose selection at cycle start $T_c$, we chose the patient state
\begin{equation}\label{eq:Patientstate}
s_{c-1}  = \Big(\text{sex}, \text{age}; \; \text{ANC}_0, g_1,\ldots,g_{c-1} \Big)\,,
\end{equation}
with $s_0 = (\text{sex}, \text{age}; \; \text{ANC}_0)$.
The covariates sex, age, have previously been identified as important predictors of exposure \cite{Joerger2012a}, and baseline absolute neutrophil counts $\text{ANC}_0$, as a crucial parameter in the drug-effect model \cite{Friberg2002,Henrich2017a}. 
We included the neutropenia grades of all previous cycles $g_{1:c-1}=(g_1,\ldots,g_{c-1})$ to account for the observed cumulative behavior of neutropenia \cite{Henrich2017a,Huizing1997}.
%Methods intro:
%24+190+120=334

% -----------------------------------------------------------------------------------------------------------------------------------------------------
\subsection{MIPD framework}
% -----------------------------------------------------------------------------------------------------------------------------------------------------

MIPD builds on prior knowledge from NLME analyses of clinical studies \cite{Maier2020}. 
The structural and observational models are generally given as
\begin{align}
\frac{\rd x}{\rd t}(t) &= f(x(t);\theta,d), \qquad x(0) =x_0(\theta)  \label{eq:ode} \\
h(t) &= h(x(t),\theta) \label{eq:observable} %
\end{align}
with state vector $x=x(t)$ (e.g., neutrophil concentration), parameter values $\theta$ (e.g., mean transition time) and rates of change $f(x;\theta, d)$ for given doses $d$. 
The initial conditions $x_0$ are given by the pre-treatment levels (e.g., $\text{ANC}_0$). 
A statistical model links the observables, the quantities that can be measured $h_j(\theta) = h(x(t_j),\theta)$ at time points $t_j$ to observations $(t_j, y_j)_{j=1,\ldots,n}$ taking into account measurement errors and potential model misspecifications, e.g.,
\begin{equation}
Y_{j|\Theta=\theta} = h_j(\theta) + \epsilon_j\,
%\label{eq:ResidualModel}\\
\end{equation}
with $\epsilon_j \sim_\iid \mathcal{N}(0,\Sigma)$.
In more general terms, $Y_{j|\Theta=\theta} \sim p\big(\,\cdot\, | \theta; h_j(\theta),\Sigma\big)\,,$ with $j=1,\dots,n$ independent.
%The dot (`$\cdot$') serves as a placeholder for the argument in a probability distribution. 
The prior distribution for the individual parameters is given by a covariate and statistical model,
\begin{equation}
\Theta \sim p_\Theta\big(\hatthetaTV(\cov),\Omega\big) \label{eq:IIVModel}%
\end{equation}
with $\hatthetaTV(\cov)$ denoting the typical values (TV), which generally depend on covariates `cov', and $\Omega$ the inter-individual variability . 
We used the term `model' to refer to eqs.~\eqref{eq:ode}-\eqref{eq:IIVModel}, and the term `model state of the patient' to refer to a model-based representation of the state of the patient, i.e., a distribution of state-parameter pairs $(x,\theta)$, or just a single (reference) state-parameter pair. 
In the proposed approaches, the model is used to simulate treatment outcomes (in RL called ``simulated experience"), or to assimilate TDM data and infer the model state of the patient, or both. 
To infer the patient state \eqref{eq:Patientstate}, the grade of neutropenia of the previous cycle $g_{c-1}$ needs to be determined; either directly from the TDM data ($y_{c-1} \mapsto g_{c-1} \mapsto s_{c-1}$) or based on a model simulation of the model state of the patient ($(x,\theta) \mapsto c_\text{nadir} \mapsto g_{c-1} \mapsto s_{c-1}$).

\begin{figure}[H]
\centering
\includegraphics[width =\linewidth]{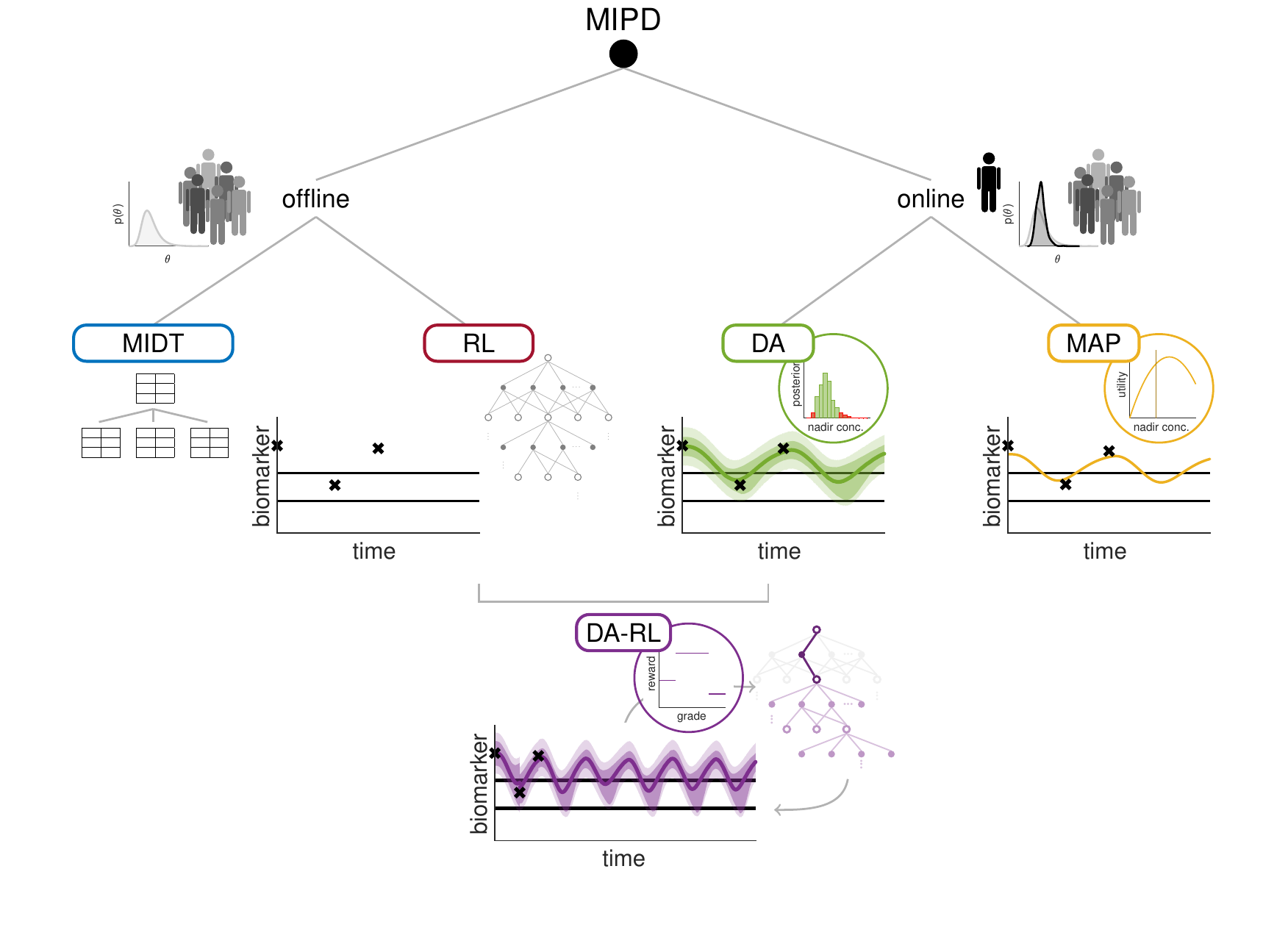}
\caption{\textbf{Overview of different approaches for model-informed precision dosing (MIPD)}. The different methods can be categorized according to the time when the computational effort to calculate the optimal dose must be made. \textit{Offline approaches} calculate optimal doses for all possible covariates and state combinations prior to any treatment, like in model-informed dosing tables (MIDT) and reinforcement learning (RL). The physician selects the dosing recommendation in the table/tree based on specific patient information (covariates, observations). While the TDM data (measured biomarker) are used to determine the entry in the table/tree, the table/tree itself is static. \textit{Online approaches} solve an optimization problem at any decision time point, i.e., when a dose has to be given. They integrate patient-specific TDM data using Bayesian data assimilation (DA) or maximum a-posteriori (MAP) estimation. \textit{Offline--online approaches} allocate computational resources between offline and online. Pre-calculated dosing decision-trees are individualized during treatment, based on TDM data.}
\label{fig:OnlineOfflineFig}
\end{figure}

\noindent
We considered three different approaches towards MIPD, see Figure~\ref{fig:OnlineOfflineFig}:

(i) \textit{Offline approaches} support dose individualization based on model-informed dosing tables (MIDT) or dosing decision trees (RL-guided dosing).
At the start of therapy a dose based on the patient's covariates and baseline measurements is recommended.
During therapy, the observed TDM data are used to determine a path through the table/tree;
While the treatment is individualized to the patient (based on a-priori uncertainties), the procedure of dose individualization itself does not change, i.e., the tree/table is static. 
As such, it can be communicated to the physician before the start of therapy. 
%A common practice and simplified version of this dose individualization is to stratify dosing solely based on the patient's covariates, e.g. BSA-guided dosing. 

(ii) \textit{Online approaches} determine dose recommendations based on a model state of the patient and its simulated outcome. 
Bayesian DA or MAP estimation assimilate individual TDM data to infer the posterior distribution or MAP point-estimate as model state of the patient, respectively. 
While online approaches tailor the model (more precisely, the parameters) to the patient, clinical implementation requires an IT infrastructure and expert personnel, which constitutes a current challenging problem and hinders broad application \cite{Keizer2020}.

%lack an analogue of the dosing decision trees/tables that can be communicated to the physician. 

(iii) \textit{Offline--Online approaches} combine the advantages of dosing decision trees and an individualised model. 
The individualised model is used in two ways, to infer the patient state more reliably than sparsely observed TDM data and to individualize the dosing decision tree (using individualized uncertainties, rather than population-based uncertainties).   
%306
\\

\noindent
Key to all approaches is the so-called reward function $R$ (RL terminology), also termed cost or utility function, defined on the set $\mathcal{S}$ of patient states
\begin{equation}\label{eq:evalfunc}
R:\mathcal{S} \rightarrow \mathbb{R}.
\end{equation}
Ideally, the reward corresponds to the net utility of beneficial and noxious effects in a patient given the \textit{current} state \cite{Sheiner1997}. 
%In practice, however, the reward only takes some of the desired aspects into account. We assumed that the larger the reward, the better.
%86
For neutrophil-guided dosing, a reward function was suggested that maps (MAP-based) nadir concentrations to a continuous score \cite{WallinPage2009} or penalizes the deviation from a target nadir concentration \cite{Wallin2009}, see also Section~S~8.5 and Figure~S~8. 
The individualized uncertainties quantified via DA allow to consider the probability of being within/outside the target range in the reward function \cite{Maier2020}, which is more closely related to clinical reality.
For the patient state \eqref{eq:Patientstate} used in RL we also designed the reward function to account for efficacy and toxicity. 
 We chose to penalize the short-term goal (avoiding life-threatening grade 4) more than the long-term goal (increased median survival associated with neutropenia grades 1-4 \cite{DiMaio2006}) :
\begin{equation}\label{eq:Rewardfunc}
R(s_c) =
\begin{cases}
-1 \qquad \text{if $g_c=0$}\,,\\
\hphantom{-}1 \qquad \text{if $g_c=1,2,3$}\,,\\
-2 \qquad \text{if $g_c=4$}\,.
\end{cases}
\end{equation}
%We chose to penalize the immediate life-threatening grade 4 more than the lack of long-term efficacy associated with grade 0.
%The degree of penalization could also be related to other therapeutic goals, the experience of the physician and potentially also to the personal preference of the patient. 
%In principle, the reward might also change over the course of the therapy, but we assumed for the sake of simplicity a constant reward. 
%200
\\

%MIPD framework:
%263+306+86+200= 855

% -----------------------------------------------------------------------------------------------------------------------------------------------------
\subsection{RL-guided dosing}\label{sec:RL}
% -----------------------------------------------------------------------------------------------------------------------------------------------------

\begin{figure}[H]
\centering
\includegraphics[width =\linewidth]{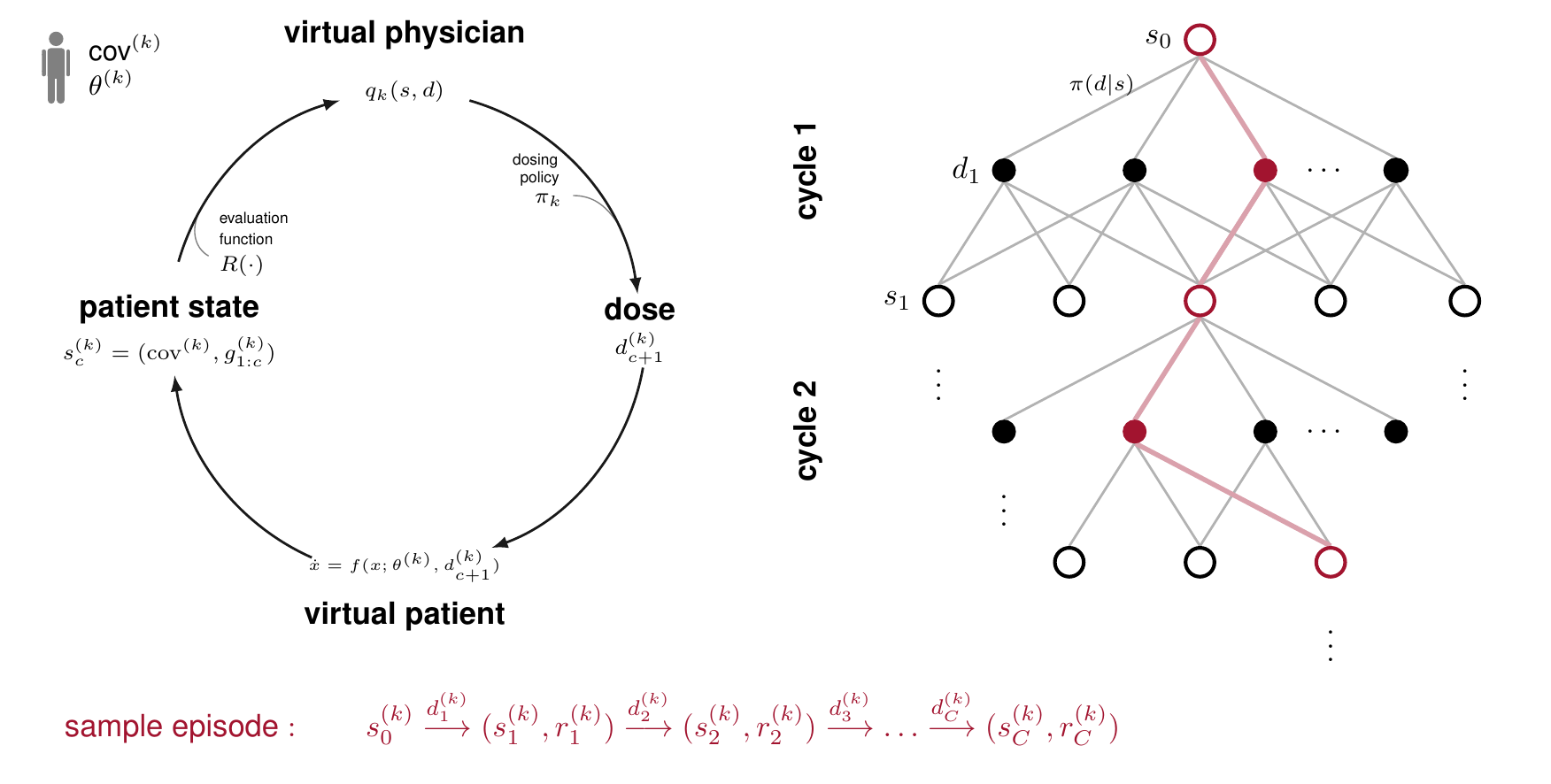}
\caption{\textbf{Model-based reinforcement learning (planning)}. The expected long-term return (action-value function) is estimated based on simulated experience (sample approximation Eq.~\eqref{eq:qest}). For simulating experience, an ensemble of virtual patients is generated, $k=1\dots,K$ (for all covariate classes $\mathcal{COV}_l$, $l=1,\dots,L$, covariates cov$^{(k)}$ are sampled within the covariate class and model parameters $\theta^{(k)}$ are sampled from the prior distribution). At start of each cycle $c$, a dose $d_{c+1}^{(k)}$ is chosen according to the current policy $\pi_k$, and the outcome (grade of neutropenia) is predicted based on the model $\dot x = f(x; \theta^{(k)},d_c^{(k)})$ for the sample parameter vector $\theta^{(k)}$ and chosen dose. The updated patient state $s_c^{(k)}$ is assessed using the reward function $R$. The sequential dose selections (going through the circle $C$ times (left part)) lead to so-called sample episodes; the entirety of episodes to a tree structure.
}
\label{fig:Planning}
\end{figure}

RL problems can be formalized as Markov decision processes, modeling sequential decision making under uncertainty, and are closely related to stochastic optimal control \cite{Bertsekas2019}. 
%RL has been applied very successfully to games, e.g., AlphaGo Zero \cite{Silver2017GO}, but also to a model-informed therapeutical setting with the aim of finding an optimal dosing strategy \cite{YauneyShah2018,Yu2019,Magni2019PAGE}. 
In RL, the goal of a so-called agent (here: the virtual physician) is to learn a policy (strategy) of how to act (dose) best with respect to optimizing a specific expected long-term return (response), given an uncertain and delayed feedback environment (virtual patient) \cite{Zhavoronkov2020,SuttonBarto2018,Magni2019PAGE,Ribba2020}, see Figure~\ref{fig:Planning}. 
%It can be model-free (so-called direct RL), just relying on rich data, or model-based and suitable for sparse data. 
%CM: Ich hab den Satz mal rausgenommen, das er verwirrend sein könnte weil wir model-based RL machen in dem wir model-free RL Methoden mit simulierten Daten verwenden.
%Translated to a therapeutical setting, a virtual physician (agent) interacts with a virtual patient (environment) by choosing a dose (action) based on the patient status (state) according to a dosing strategy (policy) with the goal to optimize a specific long-term response (return) \cite{}, see Figure~\ref{fig:Planning}.
%142
\\

\noindent
A Markov decision process comprises a sequence including states $S_c$, actions $D_c$ and rewards $R_c$ with the subscript $c$ referring to time (e.g., treatment cycle). 
If there is a natural notion of a final time $c=C$ (e.g., therapy of six cycles), the sequence is called an episode.
Every episode corresponds to a path in the tree of possibilities (Figure~\ref{fig:Planning}). 
Due to unexplained variability between patients (and occasions), transitions between states are characterized by transition probabilities \mbox{$\mathbb{P}[S_{c+1}=s_{c+1}|S_c=s_c,D_{c+1}=d_{c+1}]$}.
The reward is defined via the reward function, i.e., $R_c = R(S_{c})$, while a so-called dosing policy $\pi$ models how to choose the next dose  
\begin{equation}
\pi(d|s)=\mathbb{P}[D_{c+1}=d | S_c=s].
\end{equation}
A dosing policy is evaluated based on the so-called return $G_c$ at time step $c$, defined as the weighted sum of rewards over the \textit{remaining} course of therapy 
\begin{equation}\label{eq:return}
G_c = R_{c+1} + \gamma R_{c+2} + \dots + \gamma^{C-(c+1)} R_{C} = \sum_{k=c+1}^{C} \gamma^{k-(c+1)} R_k\,.
\end{equation}
The discount factor $\gamma \in [0,1]$ balances between short-term ($\gamma \rightarrow0$) and long-term ($\gamma \rightarrow1$) therapeutic goals (see Sections~S~6 and S~8.7.1). 
Ultimately, the objective is to maximize the expected long-term return
\begin{equation} \label{eq:av_fct}
q_\pi(s,d) := \mathbb{E}_\pi[G_c|S_c=s,D_{c+1}=d] \,, 
\end{equation}
given the current state $S_c=s$ and dose $D_{c+1}=d$ over the space of dosing policies $\pi$.  
The function $q_\pi$ is called the \textit{action-value function} \cite{SuttonBarto2018}.
Due to the close interplay between $q_\pi$ and $\pi$, learning an optimal policy involves an iterative process of value estimation and policy improvement  \cite{SuttonBarto2018}. 

Model-based RL methods that rely on sampling (sample-based planning) estimate the expected value in eq.~\eqref{eq:av_fct} via a sample approximation. 
To simplify the calculations we have discretized `age' and ANC$_0$ into covariate classes $\mathcal{COV}_l$, $l=1,\dots,L$.
For each class $\mathcal{COV}_l$ consider the ensemble 
\begin{displaymath}
\mathcal{E}_\text{RL} (\mathcal{COV}_l):= \Big \lbrace \big(x_0(\theta_c^{(k)}),\theta_c^{(k)},\text{cov}^{(k)}\big) \Big \rbrace_{k=1}^K
\end{displaymath}
with $\text{cov}^{(k)}$ sampled within $\mathcal{COV}_l$ according to the covariate distributions in the CEPAC-TDM study \cite{Joerger2016a,Henrich2017Thesis}, parameter values sampled from $p_\Theta\big(\hatthetaTV(\cov^{(k)}),\Omega\big)$ and initial states according to \eqref{eq:ode}. 
Then, for each $k=1,\ldots,K$ with $K$ large, a sample episode
\begin{equation}
s_0^{(k)} \overset{d_1^{(k)}}{\longrightarrow} (s_1^{(k)},r_1^{(k)}) \overset{d^{(k)}_2}{\longrightarrow} (s^{(k)}_2, r^{(k)}_2) \overset{d^{(k)}_3}{\longrightarrow}\ldots \overset{d^{(k)}_{C}}{\longrightarrow} (s^{(k)}_{C}, r^{(k)}_{C})
\end{equation}
using policy $\pi_{k}$ 
%\cm{Ich habe hier tree policy entfernt da eigentlich nur ein Teil der sample episode in MCTS mit der tree policy ausgewaehlt wird der andere Teil wird mit der roll-out policy (random policy) ausgewaehlt}  
is determined and 
\begin{equation}\label{eq:qest}
q_{k}(s,d) = \frac{1}{N_k(s,d)} \sum_{k'=1}^k \sum_{c=1}^{n_c}  \mathbbm{1}_{(s_c^{(k')}=s,d^{(k')}_{c+1}=d)} G_c^{(k')}
\end{equation}
computed.
Here, $N_{k}(s,d)$ denotes the number of times that dose $d$ was chosen in patient state $s$ amongst the first $k$ episodes, and  $G_c^{(k)} = r^{(k)}_{c+1} + \gamma r^{(k)}_{c+2} + \ldots$.  
Ideally, $N_{k}(s,d)$ should be large for each state-dose combination to guarantee a good approximation of the expected return (law of large numbers).
This, however, is infeasible for most applications (curse of dimensionality).

%Since it is based on simulated data, there is no harm to the patient. 
%Once, the action-value function in eq.~\eqref{eq:av_fct} has been well approximated, the optimal dosing policy $\pi^*$ for clinical use selects/suggests/recommends  at each step the dose with maximal expected long-term response. 
%\cm{Ich finde den oberen Teil etwas repetitiv zu dem unteren Abschnitt nach Gl 12 und nach Gl 14. Word count momentan $>4300$}
%\hui{es scheint zwei optimale Policies zu geben: jene, die als Ergebnis von MCTC+UCT im Iterationsprocess approximiert wird, nennen wir die Approximation mal $\pi_K$, und die, welche wir fuer die klinische Praxis vorschlagen (greedy-policy), $\pi^*$ mit $\pi^*=\text{arg-max } q_{\pi_K}$. Nach meinem Verstaendnis ist $q_{\pi^*}\neq  q_{\pi_K}$. Eigentlich brauchen wir ja $q_{\pi^*}$ nicht, sodass ich optimal policy nur im Zusammenhang mit $\pi_K$ verwenden wuerde, genauer: $\pi_K$ ist unser Schaetzer fuer $\pi_\text{opt}$, also $\widehat{\pi}_\text{opt} = \pi_K$ [eigentlich genauer $\widehat{\pi_\text{opt}} = \pi_K$], siehe auch unten) und wir $\pi^*$ vielleicht policy for clinical practice oder clinical policy nennen sollten. Zudem ist die 'Lage' von $k$ uneinheitlich: $\cdot^{(k)}$ und manchmal $\cdot_k$.}
%\cm{Das mit der Lage von $k$ ist auch Absicht! $\cdot^{(k)}$ steht fuer gesamplete Groessen und $\cdot_k$ sind cumulativ berechnete Groessen die alle $k'=1,\dots,k$ beinhaltet z.B. running mean. Also wir unterscheiden in der $k$-ten Iteration gesamplete Werte von in der $k$-ten Iteration berechneten Werten.}
%53+4+30+2+30+30+65+50=264

%130
Therefore, we employed MCTS in conjunction with UCT as policy in the iterative training process
\cite{Auer2002,Coulom2006,Silver2016,KocsisSzepesvari2006,Rosin2011}:\\
%\hui{Die wechselseitige Abhaengigkeit von $\pi_k$ und $\hat{q}_k$ war immer noch nicht ganz einfach zu verstehen. Auch nicht, warum $q_k$ einen Hut $\hat{.}$ hatte und $\pi_k$ nicht (obwohl, dafuer habe ich eine Idee). Ist doch eher so, dass final $\widehat{\pi_\text{opt}} = \pi_K$ ist. Irgendwo muss eine Gleichung auftauchen, bei der links $k+1$ und rechts $k$ als Index auftaucht, damit man das Fortschreiten der Iteration verstehen kann. Passt das so (auch gleich mit Iterationsindex oben)?}
\begin{equation}\label{eq:UCT}
\pi_{k+1}(d_{c+1}|s_c) = 
\begin{cases}
1 \qquad \text{if} \ d_{c+1} = \underset{d \in \mathcal{D}}{\arg \max} \ 
\text{UCT}_k(s_c,d)\,,\\
%q_{{k}}(s_c,d) + U_{k}(s_c,d)\,,\\
0 \qquad \text{else}
\end{cases}
\end{equation}
with $\text{UCT}_{k}$ defined based on the current sample estimate $q_{k}(s_c,d)$
\begin{equation}\label{eq:U}
\text{UCT}_k(s_c,d) = \underbrace{q_{{k}}(s_c,d) \vphantom{\frac{\sqrt{\sum_{d'}N_{k}(s_c,d')}}{N_{k}(s_c,d)+1}}}_\text{exploitation} +\epsilon_c \underbrace{\frac{\sqrt{N_{k}(s_c)}}{N_{k}(s_c,d)+1}}_\text{exploration}.
\end{equation}
It successively expands the search tree (Figure~\ref{fig:Planning}) by focusing on promising doses (exploitation, large $q_{k}(s_c,d)$), while also encouraging exploration of doses that have not yet been tested exhaustively (small $N_{k}(s_c,d)$ relative to the total number of visits $N_{k}(s_c):=\sum_{d'}N_{k}(s_c,d')$ to state $s_c$). 
The parameter $\epsilon_c$ balances exploration vs.\ exploitation; it depends on the range of possible values of the return and current state of the therapy (cycle $c$), see eq.~(S 10). 
Finally, we define $\widehat{\pi}_\text{UCT} = \pi_{K}$ as estimate of the optimal dosing policy \textit{in the training setting} (learning with virtual patients), and $\hat{q}_{\pi_\text{UCT}}=q_K$ as an estimate of the associated expected long term return. 
\textit{In a clinical TDM setting} (RL-guided dosing), we finally use $\pi^*=\text{arg max } \hat{q}_{\pi_\text{UCT}}$, i.e., $\epsilon_c=0$ (no exploration) in Eq.~\eqref{eq:U}.
See Section~S~6.1 for details.

%75

%RL-guided dosing:
%

% -----------------------------------------------------------------------------------------------------------------------------------------------------
\subsection{DA-guided dosing}\label{sec:DA}
% -----------------------------------------------------------------------------------------------------------------------------------------------------
Sequential DA approaches have been introduced as more informative and unbiased alternatives to MAP-based predictions of the therapy outcome, since they more comprehensively make use of patient-specific TDM data \cite{Maier2020}. 
The individualized uncertainty in the model state of the patient is inferred and propagated to the predicted therapy time course, allowing to predict the probability of possible outcomes. 
For this, the uncertainty in the individual model parameters is sequentially updated via Bayes' formula, i.e.,
\begin{equation}\label{eq:seqBayes}
p(\theta|y_{1:c}) \propto p(y_{c}|\theta)\cdot p(\theta|y_{1:c-1})\,,
\end{equation}
where $y_{1:c}=(y_1,\dots,y_c)^T$ denotes the TDM data up to and including cycle $c$, and $y_c=(y_{c1},\dots,y_{cn_c})^T$ the measurements taken in cycle $c$. 
Since the posterior distribution $p(\theta|y_{1:c})$ generally cannot be determined analytically, DA approaches approximate it by an ensemble of so-called particles: 
\begin{displaymath}
\mathcal{E}_{1:c}:=\Big \lbrace \big(x_{1:c}^{(m)},\theta_c^{(m)},w_c^{(m)} \big) \Big \rbrace_{m=1}^M\,.
\end{displaymath}
In our context, a particle represents a potential model state of the patient (for the specific patient covariates $\text{cov}$) with a weighting factor  $w_c^{(m)}$ characterizing how probable the state is (given prior knowledge and TDM data up to $c$). 
As more TDM data is gathered, the Bayesian updates reduce the uncertainty in the model parameters and consequently in the therapeutic outcome, see Figure~\ref{fig:DARL} (DA part, reduced width of CrI/PI) and Section~S~5. 
Since subtherapeutic as well as toxic ranges, i.e., very low or high drug/biomarker concentrations, are described by the tails of the posterior distribution, the uncertainties provide crucial additional information compared to the mode (MAP estimate) for dose selection. 

We chose the \textit{optimal dose} to be the dose that minimizes the weighted risk of being outside the target range, i.e., the a-posteriori probability of $g_c=0$ or $g_c=4$:
\begin{equation}\label{eq:DAOpt}
 d_{c+1}^*=\underset{d \in \mathcal{D}}{\arg \min} \ \lambda_0 \sum_{m=1}^M w_c^{(m)} \mathrm{1}_{\lbrace g(\theta_c^{(m)},d)=0\rbrace} + \lambda_4 \sum_{m=1}^M w_c^{(m)} \mathrm{1}_{\lbrace g(\theta_c^{(m)},d)=4\rbrace}
\end{equation}
with $g(\theta_c^{(m)},d)$ denoting the predicted neutropenia grade by forward simulation of the $m$-th particle for dose $d$.
We penalized grade 4 more severely than grade 0, i.e., $\lambda_4=2/3$ and $\lambda_0=1/3$, similarly as in \eqref{eq:Rewardfunc}.

The integration of an ensemble of particles into the optimization problem, instead of a point estimate (as in MAP-guided dosing), increases the computational effort and complexity of the problem. 
%In addition, we are confronted with a multi-objective optimization problem leading to pareto optimal solutions, e.g. we cannot decrease the probability of grade 4 without increasing the probability of grade 0. 
If time or computing power is limited, approximations have to be used, e.g., by solving only for the next cycle dose rather than all remaining cycles at the cost of neglecting long-term effects. 
Alternatively, the number of particles $M$ could be reduced (we used both approximations in this study); see also Section~S~8.6.
The DA optimization problem is stated in the space of actions (doses), while RL optimizes in the space of states by estimating the expected long-term return as intermediate step (eq.~\ref{eq:qest}) thereby promising efficient solutions to the sequential decision-making problem under uncertainty \cite{SuttonBarto2018}.
%\hui{hier bin ich mir nicht sicher, was du genau sagen willst, da der Begriff 'stochastic sequential decision problem' nirgendwo anders im Manuscript auftaucht: ... thereby promising efficient solutions to the stochastic sequential decision problem \cite{SuttonBarto2018}}
%\cm{ sequ decision-making under uncertainty wurde bereits zuvor verwendet}
%300

%Yet it is important to point out that an online learning strategy opposed to an offline approach allows to continuously improve with respect to specific individuals. As a result online learning can often significantly improve the performance of a dosing scheme for a specific patient, which in return justifies associated additional computational costs.
%

% -----------------------------------------------------------------------------------------------------------------------------------------------------
\subsection{DA-RL-guided dosing}\label{sec:DAplan}
% -----------------------------------------------------------------------------------------------------------------------------------------------------

% \cm{
%Two aspects: (1) patient status = DA state
%(2) fixed numbers in tables are for patient groups rather than individuals
%}

\begin{figure}[H]
\centering
\includegraphics[width =0.95\linewidth]{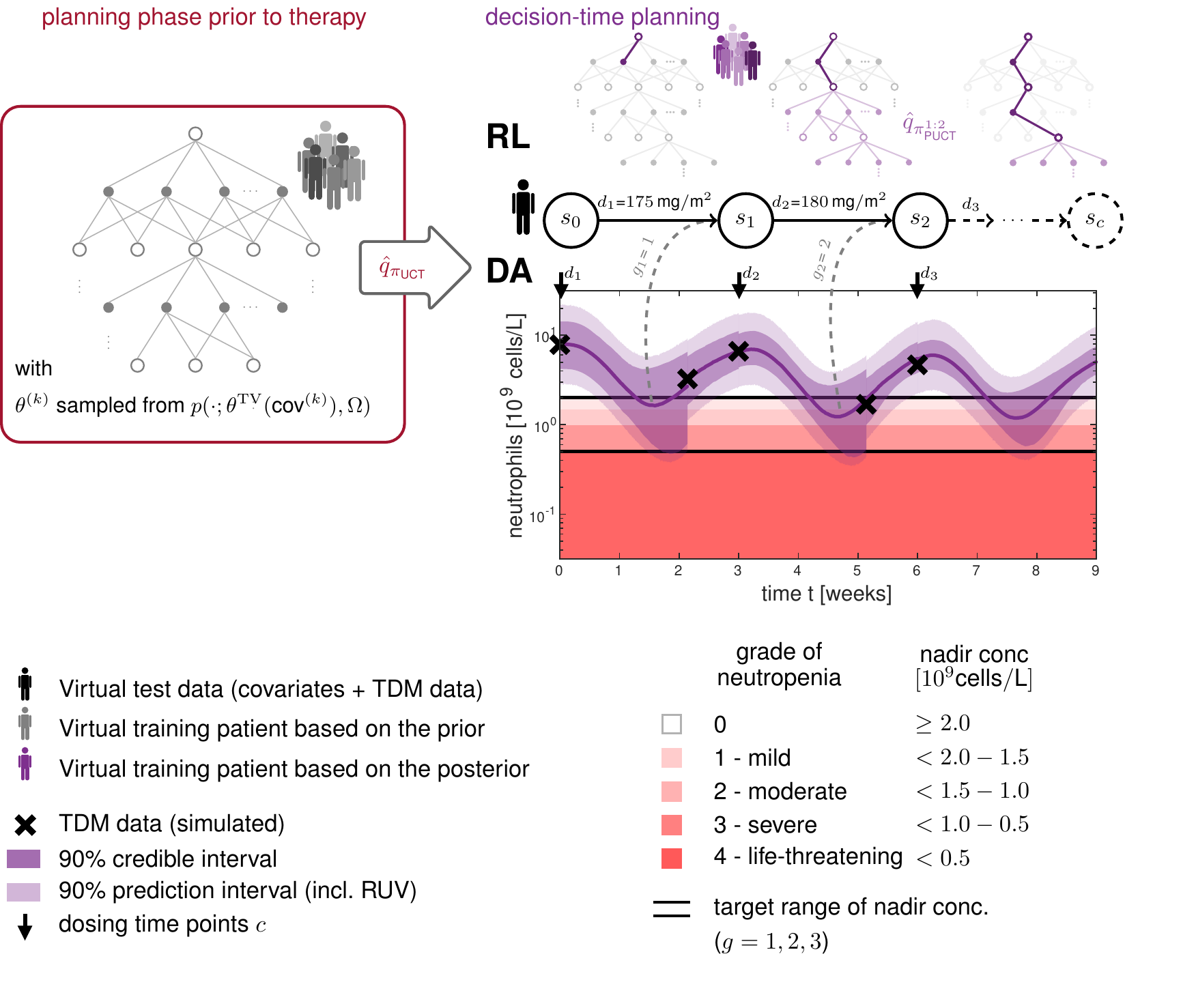}
\caption{\textbf{The interplay of data assimilation (DA) and reinforcement learning (RL).} 
In the planning phase prior to therapy, the expected long-term return $q_{\pi_0}:=\hat{q}_{\pi_\text{UCT}}$ is estimated in Monte Carlo Tree search (MCTS) with upper confidence bound applied to trees (UCT) using an ensemble of covariates $\text{cov}^{(k)}$ and parameter values $\theta^{(k)}\sim p(\cdot|\theta^\textrm{TV}(\text{cov}^{(k)}),\Omega)$. 
The first dose is selected based on $q_{\pi_0} :=\hat{q}_{\pi_\text{UCT}}$ for the patient specific covariate class. 
The DA algorithm initializes a particle ensemble given the patient's covariates $\text{cov}$. The ensemble is propagated forward continuously in time, and observed patient TDM data (black crosses) is assimilated when it becomes available. This results in updated uncertainty, visible as 'cuts' in the credible/prediction intervals. In contrast, the RL state evolves in discrete time steps $c$ according to the decision time points and only considers selected features/summaries of the model state of the patient, e.g., smoothed posterior expectation of nadir concentrations translated into neutropenia grades. At each decision time point, the posterior model state of the patient is used to refine the prior computed $\hat{q}_{\pi_\text{UCT}}$ (grey tree) for future reachable states (light purple tree).  This individualizes the tree based on individualized uncertainties ($\mathcal{E}_{1:c}$). }
\label{fig:DARL}
\end{figure}

The particle-based DA scheme and the model-based RL scheme address the problem of personalized dosing from different angles. 
A combined DA-RL approach therefore offers several advantages by integrating individualized uncertainties provided by DA within RL, see Figure~\ref{fig:DARL}. 
First, instead of the observed grade (e.g., measured neutrophil concentration on a given day, translated into the neutropenia grade),  we may use the smoothed posterior expectation of the quantity of interest (e.g., predicted nadir concentration), see Section~S~7. 
This reduces the impact of measurement noise and the dependence on the sampling day.
%, and provides the best state estimate given the prior, the TDM data and the model. 
Second, for model simulations within the RL scheme, we can sample from the posterior $p(\theta|y_{1:c})$ represented by the ensemble $\mathcal{E}_{1:c}$, i.e., from individualized uncertainties, instead of the prior $p(\theta)$, i.e. population-based uncertainties. 
During the course of the treatment, the ensemble of potential model states of the patient is continuously updated when new patient-specific data are obtained (see eq.~\eqref{eq:seqBayes}). 
This allows to individualize the expected long-term return during treatment as new patient data is observed, see Figure~\ref{fig:DARL}, i.e., the dosing decision tree in RL is updated prior to the next dosing decision.  

\begin{figure}[H]
\centering
\includegraphics[width =1\linewidth]{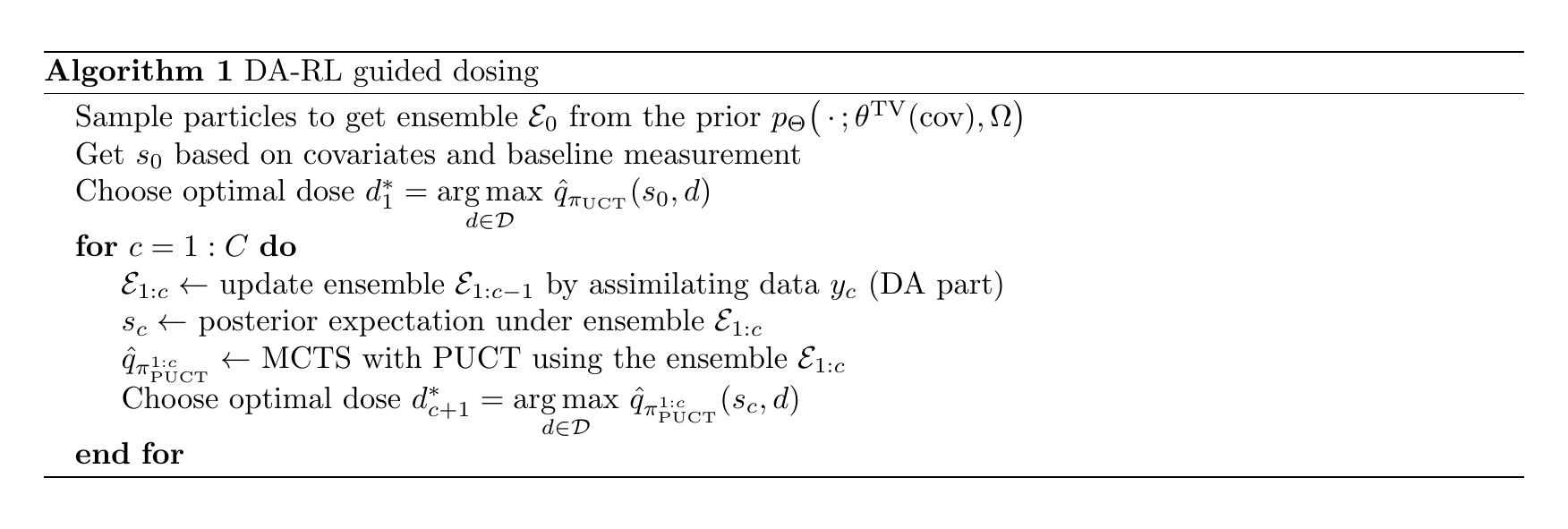}
\caption{\textbf{Pseudo Code of DA-RL-guided dosing}. 
At therapy start a particle ensemble $\mathcal{E}_0$ for the sequential data assimilation (DA) approach is sampled from the prior parameter distribution given the patient's covariates. Then for the initial state $s_0$ (pretreatment) the first dose is selected according to the prior expected long-term return $q_{\pi_0} :=\hat{q}_{\pi_\text{UCT}}$ calculated beforehand in the prior planning phase (Monte Carlo Tree Search (MCTS) with upper confidence bound applied to trees (UCT)). The selected dose is given to the patient and patient-specific therapeutic drug/biomarker monitoring (TDM) data $y_c$ is collected within cycle $c$. The TDM data is assimilated via a sequential DA approach (particle filter and smoother) creating a posterior particle ensemble $\mathcal{E}_{1:c}$. For subsequent dose decisions ($c=1,\dots,C$). The new patient state is inferred using $\mathcal{E}_{1:c}$, e.g., smoothed posterior expectation of the nadir concentration translated into the neutropenia grade of the cycle. Then a MCTS is started from the current patient state using $\mathcal{E}_{1:c}$ in the model simulations. Within the tree search we use the PUCT algorithm with prioritized exploration based on $\hat{q}_{\pi_\text{UCT}}$.}
\label{fig:DARLAlg}
\end{figure}

Since the refinement as well as the DA part has to run in real time (online), it has to be performed efficiently.
We do not need to take all possible state combinations into account, but only those that are still relevant for the \textit{remaining} part of the therapy. 
This reduces the computational effort, in particular for later cycles. 
The proposed DA-RL approach results in a sequence of estimated optimal dosing policies $\widehat{\pi}^{1},\widehat{\pi}^{1:2},\dots$ with $\widehat{\pi}^{1:c}$ denoting the estimated optimal dosing policy based on TDM data $y_{1:c}$, i.e. based on $\mathcal{E}_{1:c}$. 
In addition, we do not need to estimate the individualized action-value function from scratch, but can exploit $q_{\pi_0} :=\hat{q}_{\pi_\text{UCT}}$ as a prior determined by the RL scheme prior to any TDM data (see paragraph following Eq.~\eqref{eq:U}).
In PUCT (predictor+UCT \cite{Rosin2011,Silver2017GO}), the exploitation vs.\ exploration parameter $\epsilon_c$ in Eq.~\eqref{eq:U} is modified to prioritize doses with high a-priori expected long-term return:
\begin{equation}\label{eq:PriorProb}
U_{k}(s_c,d) = 
\underbrace{q^{1:c}_{{k}}(s_c,d) \vphantom{\frac{\sqrt{\sum_{d'}N^{(k)}(s_c,d')}}{N^{(k)}(s_c,d)+1}}}_\text{exploitation} +
\, \epsilon_c \cdot \underbrace{\frac{\exp\!\big(\hat{q}_{\pi_{UCT}}(s,d)\big)}{\sum_{d'} \exp\!\big(\hat{q}_{\pi_{UCT}}(s,d')\big)}}_\text{priortising}
\underbrace{\frac{\sqrt{N_{k}(s_c)}}{N_{k}(s_c,d)+1}}_\text{exploration}.
\end{equation}
Finally, we define  $\widehat{\pi}^{1:c}_\text{PUCT} = \pi^{1:c}_{K}$ based on $\mathcal{E}_{1:c}$ as an estimate of the optimal \textit{individualized} dosing policy \textit{in the training setting} (using Eqs.~\eqref{eq:UCT}+\eqref{eq:PriorProb}), and $\hat{q}_{\pi_\text{PUCT}^c}=q_K$ as an estimate of the associated expected long term return based on $\mathcal{E}_{1:c}$. 
For individualized dose recommendations \textit{in a clinical TDM setting}, we again use $\pi^*=\text{arg max } \hat{q}_{\pi_\text{PUCT}^{1:c}}$, i.e., $\epsilon_c=0$  in Eq.~\eqref{eq:PriorProb}. See Figure~\ref{fig:DARL} and \ref{fig:DARLAlg}, and 
Section~S~7. 
%142 words

%Word counts
% Methods 	333
% MIPD 		820
% RL 		613
% DA 		300
% DA-RL 		387
%SimStudy: 	472 words

% Methods total :   2925  words 

% =====================================================================================
\section{RESULTS}\label{sec:Results}
% =====================================================================================

% -----------------------------------------------------------------------------------------------------------------------------------------------------
\subsection{Novel individualized dosing strategies decreased the occurrence of \\grade 4 and grade 0 neutropenia compared to existing approaches}
% -----------------------------------------------------------------------------------------------------------------------------------------------------

We compared our proposed approaches with existing approaches for MIPD based on simulated TDM data in paclitaxel-based chemotherapy. 
The design was chosen to correspond to the CEPAC-TDM study \cite{Joerger2016a}: neutrophil counts at day 0 and 15 of each cycle were simulated for virtual patients employing a PK/PD model for paclitaxel-induced cumulative neutropenia (Figure~S~1) \cite{Henrich2017a}.
We focused only on paclitaxel dosing; we did not take into account drop-outs, dose reductions due to non-haematological toxicities, adherence and comedication.
Occurrence of grade 4 neutropenia, therefore, differed between our simplified simulation study and the clinical study (as might be expected), see Section~S~8.2.
This should be taken into account when interpreting the results.
%We compared our simulation study results (for standard and PK-guided dosing) with the observed occurrence of grade 4 in the CEPAC-TDM study, see Section~S~\ref{Supplsec:ComparisonCEPACTDM}.
To obtain meaningful statistics, all analyses were repeated $N=1000$ times with covariates sampled from the observed covariate ranges in the CEPAC-TDM study.
Detailed discussions and further analyses are provided in Section~S~2 and S~8.\\

\begin{figure}[H]
\centering
\includegraphics[width =1\linewidth]{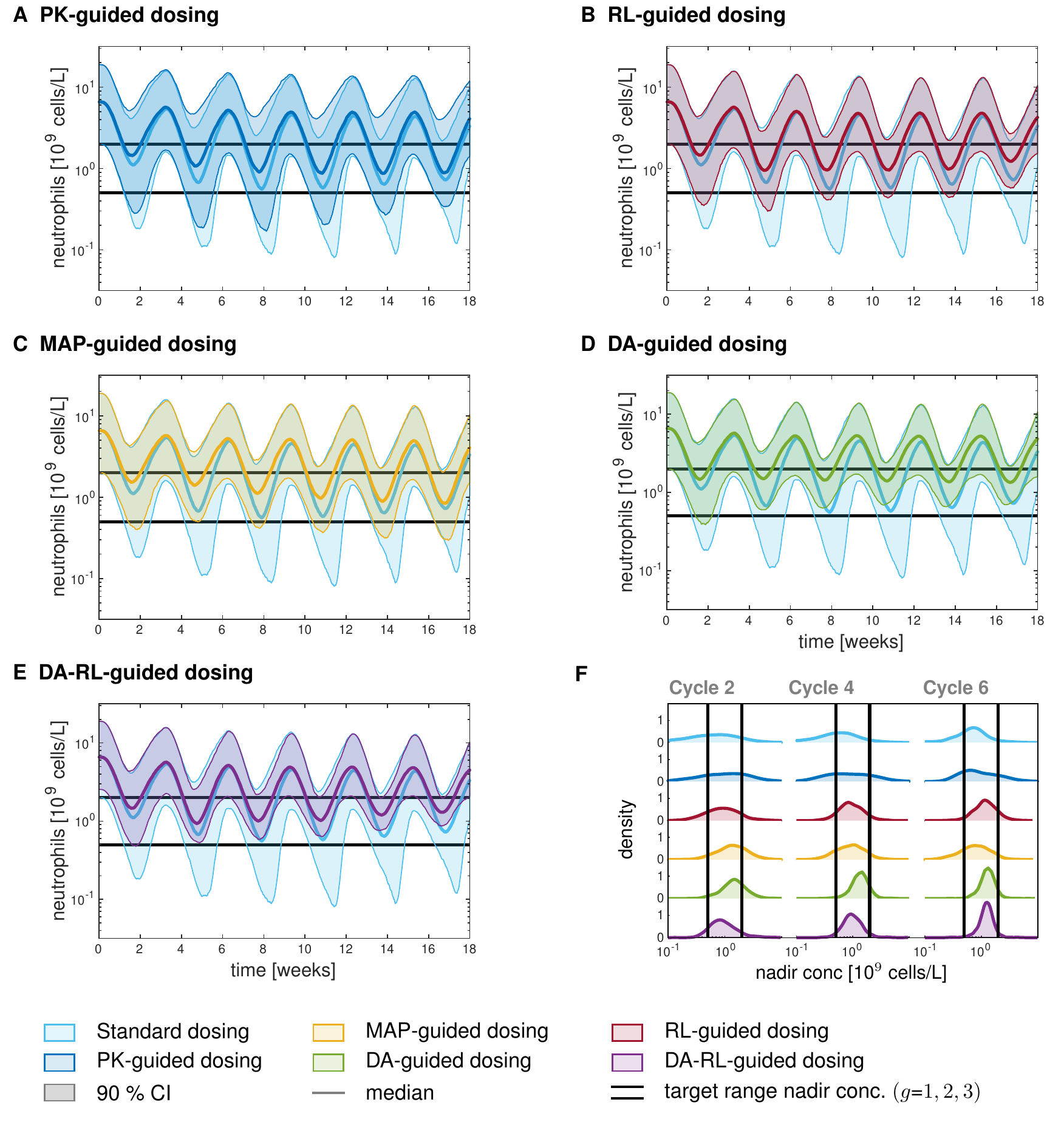}
\caption{\textbf{Comparison of different dosing policies for paclitaxel dosing}. Comparison of the 90~\% confidence intervals and median of the neutrophil concentration for the test virtual population ($N=1000$) using (\textbf{A}) PK-guided dosing (\textbf{B}) RL-guided dosing (\textbf{C}) MAP-guided dosing, (\textbf{D}) DA-guided dosing, and  (\textbf{E}) DA-RL-guided dosing, each in comparison to the standard dosing (BSA-based dosing). 
PK-guided dosing is the only approach that also takes into account exposure data ($T_{C_\text{drug} \geq 0.05\,\mu\text{mol}/\text{L}}$). 
(\textbf{F}) Comparison of the distributions of model-predicted nadir concentrations (smooth by kernel density estimation) for the test virtual population ($N=1000$).}
\label{fig:Sim1Comparison}
\end{figure}

\begin{figure}[H]
\centering
\includegraphics[width =0.65\linewidth]{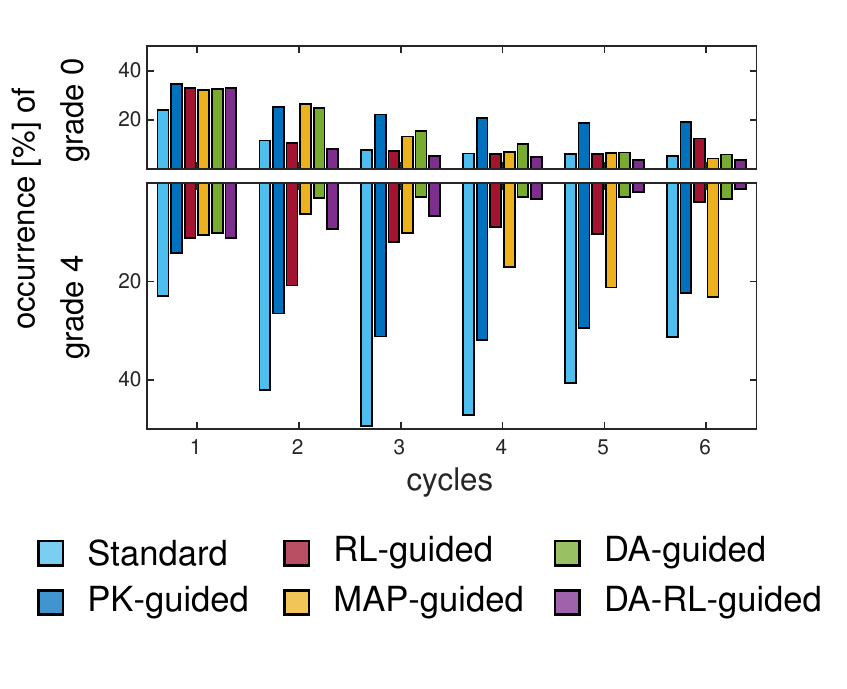}
\caption{\textbf{Occurrence of grade 0 and grade 4 for the different dosing policies}. The percentage is based on a test virtual population ($N=1000$) and six cycles (inferred from the model predicted nadir concentration). Additional analyses is provided in the supplement, Figure~S~22.
}
\label{fig:Bar04}
\end{figure}

\noindent
Figure~\ref{fig:Sim1Comparison} shows the predicted neutrophil concentrations---median \& 90\% confidence interval (CI)---over six cycles of three weeks each. 
A successful neutrophil-guided dosing should result in nadir concentrations within the target range (grades 1-3, between black horizontal lines). 
%93
%As a reference, we included the performance of the standard (BSA-guided) and PK-guided dosing strategies \cite{Joerger2012a}, see Figure \ref{fig:Sim1Comparison} A. 
%We observed that the standard dosing led to a higher occurrence of grade 4 than observed in the CEPAC-TDM study, potentially because we did not take into account comedications nor additional dose reductions due to non-haematological toxicities, see Section~S~\ref{Supplsec:ComparisonCEPACTDM} for a comparison with the CEPAC-TDM study results.
In all cycles, PK-guided dosing prevented the nadir concentrations (90\% CI)  to drop as low as for the standard dosing (Figure~\ref{fig:Sim1Comparison} A). 
However, PK-guided dosing also increased the occurrence of grade 0 (Figure~\ref{fig:Bar04}). 

RL-guided dosing controlled the neutrophil concentration well across the cycles (Figure~\ref{fig:Sim1Comparison} B) and the distribution of nadir concentrations over the whole population was increasingly concentrated within the target range (panel F). 
The occurrence of grade 0 and 4 neutropenia was substantially reduced compared to standard and PK-guided dosing (Figure~\ref{fig:Bar04}). 

For MAP-guided dosing, the occurrence of grade 4 neutropenia increased over the cycles (Figure~\ref{fig:Bar04}), showing the typical cumulative trend of neutropenia \cite{Henrich2017a}, despite inclusion of TDM data.
%44
In contrast, DA steadily guided nadir concentrations into the target range (Figure~\ref{fig:Sim1Comparison} D and F), thereby substantially decreasing the variance, i.e., the variability in outcome.
The occurrence of grade 0 and 4 was reduced considerably in later cycles (Figure~\ref{fig:Bar04}), suggesting that individualized uncertainty quantification played a crucial role in reducing the variability in outcome.
%70

% 
Integrating individualized uncertainties and considering the model state of the patient in the RL approach (DA-RL-guided dosing) also moved nadir concentrations into the target range and clearly decreased the variance (Figure~\ref{fig:Sim1Comparison} B+F). 
The slight differences between DA and DA-RL (Figure~\ref{fig:Bar04}) might be related to the difference in weighting grade 0 and 4 in the respective reward functions (eq.~\eqref{eq:DAOpt} vs.\ eq.~\eqref{eq:Rewardfunc}). 
For additional comparisons, see Figure~S~22. 

In summary, individualized uncertainties as in DA- and DA-RL-guided dosing seemed to be crucial in bringing nadir concentrations into the target range and reducing the variability of the outcome, thus achieving the goal of therapy individualization.
For this specific example, both approaches showed comparable results, but DA-RL has the greater potential for long-term optimization in a delayed feedback environment as well as integrating multiple endpoints.
%93

% -----------------------------------------------------------------------------------------------------------------------------------------------------
\subsection{Identification of relevant covariates via investigating the expected long-term return in RL}
% -----------------------------------------------------------------------------------------------------------------------------------------------------
A key object in RL is the expected long-term return or action-value function $q_{\pi}(s,d)$ (see eq.~\ref{eq:av_fct}). 
We demonstrate that it contains important information to identify relevant covariates to individualize dosing.   

\begin{figure}[H]
\centering
\includegraphics[width =1\linewidth]{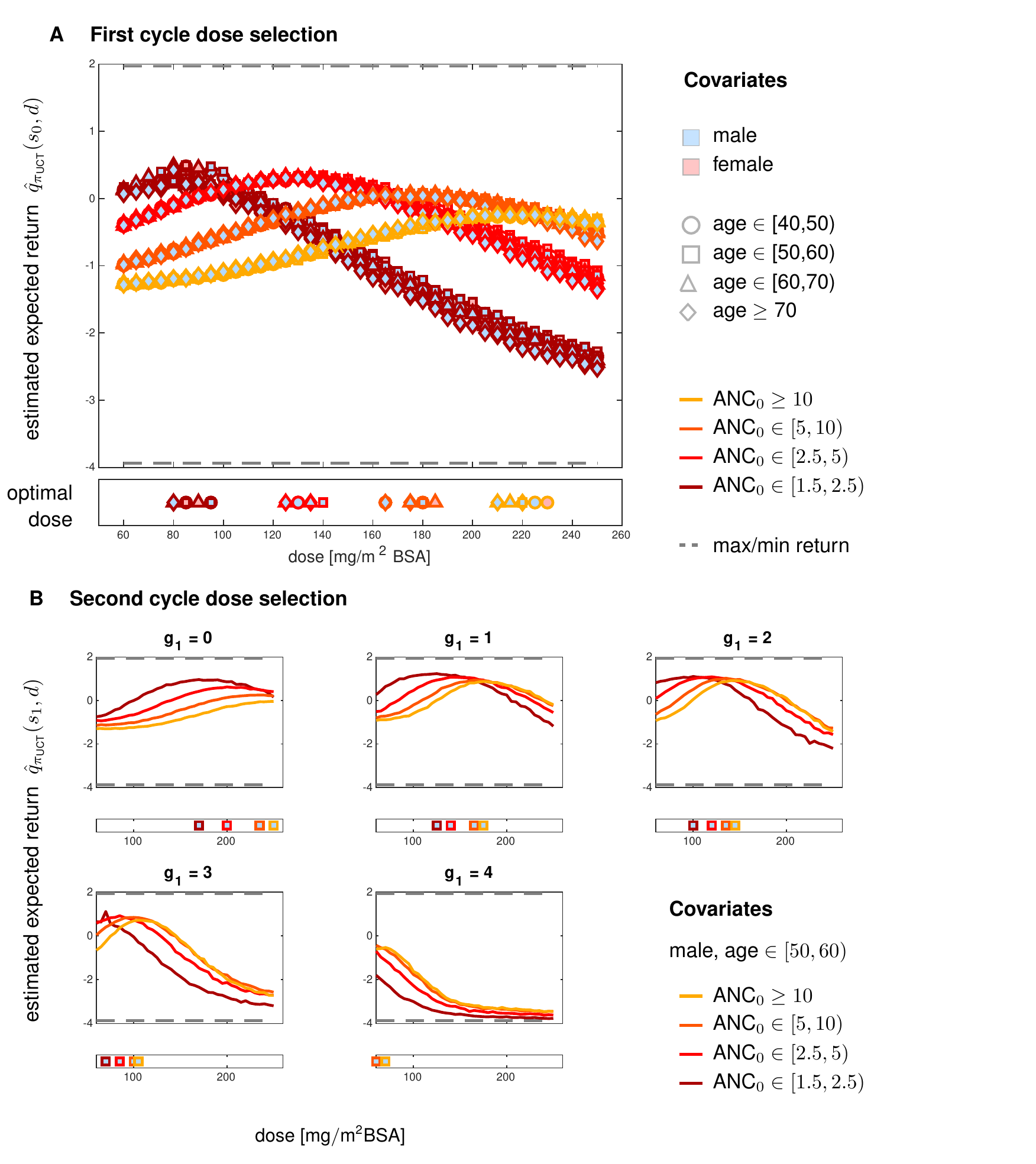}
\caption{\textbf{Expected long-term return across the dose range for dose selection}. \textbf{A}  across the considered covariate combinations for the dose selection in cycle 1. The symbols plotted below the x-axis show the optimal dose for the corresponding covariate class (i.e. the arg max of the plotted line). \textbf{B} for fixed sex and age class (here, male between 50 and 60  years) with different pre-treatment neutrophil values $\text{ANC}_0$ and observed neutropenia grades in cycle 1, i.e., $\text{g}_1$. The optimal dose for the second cycle depends on the neutropenia grade of the previous cycle and the pre-treatment neutrophil count $\text{ANC}_0$ in $[10^9 \text{cells}/\text{L}]$.
The grey dashed line shows the maximum and minimum possible return from the first cycle (A) and the second cycle (B) onwards, with $\gamma=0.5$.
The covariate classes were chosen based on the CEPAC-TDM study population: inclusion criteria were $\text{ANC}_0 >  \cdot 1.5 \cdot 10^9 \text{cells}/\text{L}$;  the typical baseline count for male was $\text{ANC}_0=6.48 \cdot 1.5 \cdot 10^9 \text{cells}/\text{L}$ (arm B). The median age was $63$ years ranging from 51 to 74 years (5th and 95th percentile of the population in arm B), see \cite{Henrich2017Thesis,Joerger2016a}. 
}
\label{fig:InvestigateQ}
\end{figure}

Figure~\ref{fig:InvestigateQ} A shows the estimated action-value function for RL-guided dosing stratified for the covariates, sex, age and baseline neutrophil counts $\text{ANC}_0$ (covariate classes are shown in the legend) for the first cycle dose selection. 
$\text{ANC}_0$ was found to be by far the most important characteristic for the RL-based dose selection at therapy start. 
Differences in age and sex played only minor roles. 
For comparison, the first cycle dose selection in the PK-guided algorithm is only based on sex and age.
The steepness of the curves gives an idea about the robustness of the dose selection.
%
%Figure~\ref{fig:InvestigateQ} B shows the action-value function for dose selection for the second cycle. 
For the second dose selection, the grade of neutropenia in the first cycle ($g_1$) has the largest impact, while larger $\text{ANC}_0$ led to larger optimal doses (Figure~\ref{fig:InvestigateQ} B).  
To illustrate the dose selection in RL, we extracted a similar decision tree to the one developed by Joerger et al. \cite{Joerger2012a}, see Figure~S~13. 

Similar investigations are not straightforward for MAP- or DA-guided dosing as no means is provided to investigate dose recommendations for an entire population; these approaches optimize doses for a single patient.
%

%In addition, the variance in return (Figure~S~\ref{Supplfig:Qvar_N} A)  provides additional information about the associated uncertainty of a dose selection. 

%303

%In the DA+RL guided approach a further individualization of the $q$ values is done whenever new patient specific data becomes available. Therefore, less iterations are possible which leads to less smooth curves, however, the search is focused on relevant doses through the prior probabilities, see Supplement Figure~S~\ref{Supplfig:Q_MCTS_DAMCTS}.

% Results total :  450 +303   = 756   words 

% =====================================================================================
\section{DISCUSSION}\label{sec:Discussion}
% =====================================================================================
We present three promising MIPD approaches employing DA and/or RL that substantially reduced the number of (virtual) patients in life-threatening grade 4 and grade 0, a surrogate marker for efficacy of the anticancer treatment.

RL-guided dosing in oncology has been proposed before \cite{YauneyShah2018}, however, only considering the mean tumor diameter. 
Since only a marker for efficacy was considered this led to a one-sided dosing scheme and resulted in very high optimal doses. 
The authors therefore introduced action-derived rewards, i.e., penalties on high doses.
In contrast, neutrophil-guided dosing considers toxicity and efficacy (link to median survival) simultaneously.
Ideally, dosing decisions should also include other adverse effects, e.g., peripheral neuropathy, tumor response or  long-term outcomes, e.g., overall or progression free survival.
Notably, RL easily extends to multiple adverse/beneficial effects and is especially suited for time-delayed feedback environments \cite{Yu2019,Zhavoronkov2020}, as typical in many diseases.

Using MCTS with UCT, we employed an RL framework that exploits the possibility to simulate until the end of therapy and evaluate the return.
Consequently, it requires less approximations as temporal difference approaches (e.g., Q-learning, used in \cite{YauneyShah2018}) that avoid computation of the return via a decomposition (Bellman equation).
Moreover, exploration via UCT allows to systematically sample from the dose range (as opposed to an $\epsilon$-greedy strategy) and allows to include additional information, e.g., uncertainties or prior information (as in PUCT). 
This becomes key when combined with direct RL based on real-world patient data, see e.g. \cite{Sutton1991,Silver2008}, which would allow to compensate for a potential model bias.
At the end of a patient's therapy the observed return can be evaluated and used to update the expected return $\hat{q}_{\widehat{\pi}}$.
This update would even be possible if the physician did not follow the dose recommendation (off-policy learning) and could be implemented across clinics, as it could be done locally without exchanging patient data.
Thus, the presented approach builds a basis for continuous learning post-approval, which has the potential to substantially improve patient care, including patient subgroups underrepresented in clinical studies.

Overall we have shown that DA and RL techniques can be seamlessly integrated and combined with existing NLME and data analysis frameworks for a more holistic approach to MIPD.
Our study demonstrates that incorporation of individualized uncertainties (as in DA) is favorable over state-of-the-art online algorithms such as MAP-guided dosing.
The integrated DA-RL framework allows not only to consider prior knowledge from clinical studies but also to improve and individualize the model and the dosing policy simultaneously during the course of treatment by integrating patient-specific TDM data. 
Thus, the combination provides an efficient and meaningful alternative to solely DA-guided dosing, as it allocates computational resources between online and offline and the RL part provides an additional layer of learning to the model (in form of the expected long-term return) that can be used to gain deeper insights into important covariates for the dose selection. 
%\hui{dabei spielt die Individualisierung durch TDM Data ja keine Rolle. Sollte man noch hervorheben, dass man die Covariate-Analyse also a-priori machen kann?}. \cm{da die word count so knapp ist wuerde ich das mal weglassen. In dem Ergebnisteil koennte ich allerdings nochmal hervorheben, welches $q$ wir verwendet haben}
Therefore showing that RL approaches can be well interpreted in clinically relevant terms, e.g., highlighting the role of ANC$_0$ values. 
%This kind of evaluation of machine learning algorithms in health care is particularly critical and has to be performed with great care \cite{Gottesman2018}. 
% INFO: im Gespraech mit Jana hatten wir eigentlich gesagt, dass bisher keine behauptet hat, dass RL als black box angesehen wird, wo wie es fuer Machine Learning der Fall ist. Daher habe ich diesen Satz rausgenommen.
%
Well-informed and efficient MIPD bears huge potential in drug-development as well as in clinical practice as it could (i) increase response rates in clinical studies \cite{Peck2015}, (ii) facilitate recruitment by relaxing exclusion criteria \cite{Darwich2017}, (iii) enable continuous learning post-approval and thus improve treatment outcomes in the long-term.

\newpage
% =====================================================================================
\section*{Acknowledgement}
% =====================================================================================

C.M. kindly acknowledges financial support from the Graduate Research Training Program PharMetrX: Pharmacometrics \& Computational Disease Modelling, Berlin/Potsdam, Germany. This research has been partially funded by 
Deutsche Forschungsgemeinschaft (DFG) - SFB1294/1 - 318763901 (associated project).
Fruitful discussions with Sven Mensing (AbbVie, Germany), Alexandra Carpentier (Otto-von-Guericke-Universitaet Magdeburg) and Sebastian Reich (University of Potsdam, University of Reading) are kindly acknowledged.

% =====================================================================================
\section*{Author Contributions}
% =====================================================================================

C.M., N.H., C.K., W.H., J.dW. designed research, C.M. mainly performed the research, C.M., N.H., C.K., W.H., J.dW. analyzed data and wrote the manuscript.

% =====================================================================================
\newpage
\bibliographystyle{cptpspref}
\bibliography{RL_project}

\end{document}

\maketitle
$^1$Institute of Mathematics, University of Potsdam, Germany,\\[1ex]
$^2$Graduate Research Training Program PharMetrX: Pharmacometrics \& Computational Disease Modelling, Freie Universit\"at Berlin and University of Potsdam, Germany\\[1ex]
$^3$Department of Clinical Pharmacy and Biochemistry, Institute of Pharmacy, Freie Universit\"at Berlin, Germany\\[1ex]
$^\ast$corresponding author (huisinga@uni-potsdam.de,)

\tableofcontents

\newpage

% =====================================================================================
\section{Introduction}
% =====================================================================================

In this supplementary text we provide more details about the employed methods, implemented algorithms and applied models. 
We also recap some general concepts in RL to support the reading of the manuscript and discuss tuning and robustness aspects. 
In combination with the main manuscript, it should be self-contained and together with the MATLAB code provided, it should enable reproduction of the simulation studies and implementation of the methods for own applications.

\begin{table}[H]
\begin{center}
\begin{tabular}{lll}
\multicolumn{2}{l}{Abbreviations}\\
\hline
ANC			&		& Absolute neutrophil counts\\
BSA 			& 		& Body surface area\\
CEPAC-TDM 	& 		& Central European Society for Anticancer Research (CESAR)\\
			&		& Study of Paclitaxel Therapeutic Drug Monitoring\\
DA			&		& Data assimilation\\
IIV			&		& Inter-individual variability\\
IOV			&		& Inter-occasion variability\\
MAP			&		& Maximum a-posteriori\\
MCTS		&		& Monte Carlo Tree Search\\
MIDT		& 		& Model-informed dosing table\\
MIPD		&		& Model-informed precision dosing \\
NLME		&		& Nonlinear mixed effects analysis\\
PK			&		& Pharmacokinetics\\
PD		  	&		& Pharmacodynamics\\
PUCT 		& 		& Predictor + UCT\\
RL 			&	        & Reinforcement learning\\
RUV 		&		& Residual unexplained variability\\
TDM			&		& Therapeutic drug/biomarker monitoring\\
TV 			&		& Typical values\\
UCT			&		& Upper confidence bound applied to trees\\
\hline
\end{tabular}
\end{center}
\caption{Abbreviations used throughout the manuscript and supplement.
}
\label{tab:Abbreviations}
\end{table}

% =====================================================================================
\section{Pharmacokinetic/pharmacodynamic (PK/PD) model for paclitaxel-induced cumulative neutropenia} \label{sec:PKPDmodel}
% =====================================================================================

We have employed published models describing the pharmacokinetics of paclitaxel as well as one of its side effects on the hematopoietic system. 
The simulation framework was previously described in the supplementary material of \cite{Maier2020}.

% -----------------------------------------------------------------------------------------------------------------------------------------------------
\subsection{Paclitaxel PK model}
% -----------------------------------------------------------------------------------------------------------------------------------------------------
Paclitaxel is a widely used anticancer drug in the treatment of ovarian, mammary and lung cancer \cite{Joerger2007,Kampan2015,Henrich2017Thesis}.
In this study, we investigated its use as first-line treatment against advanced non-small cell lung cancer in platinum-based combination therapy \cite{Belani2005,Joerger2016a}. 
Paclitaxel pharmacokinetics were previously described by a three compartment model with nonlinear distribution to one of the peripheral compartment and nonlinear elimination \cite{Joerger2012a}.
For our analysis we used the re-estimated parameter values in \cite{Henrich2017a}, see also \STabref{tab:PaclitaxelParameterEstimates}. 
The PK model includes a covariate model on the maximum elimination capacity
{\footnotesize\begin{equation*}
\mathrm{VM}_\mathrm{EL,TV,i} = \mathrm{VM}_\mathrm{EL,pop} \cdot \Big(\frac{\mathrm{BSA}_i}{1.8 \text{m}^2}\Big)^{\theta_{\mathrm{VM}_\mathrm{EL}\text{-}\mathrm{BSA}}} \cdot \Big(\theta_{\mathrm{VM}_\mathrm{EL}\text{-}\mathrm{SEX}}\Big)^\mathrm{SEX_i} \cdot \Big(\frac{\mathrm{AGE}_i}{56 \mathrm{y}}\Big)^{\theta_{\mathrm{VM}_\mathrm{EL}\text{-}\mathrm{AGE}}} \cdot \Big(\frac{\mathrm{BILI}_i}{7 \mu \M}\Big)^{\theta_{\mathrm{VM}_\mathrm{EL}\text{-}\mathrm{BILI}}}\,,
\end{equation*}}
where $\mathrm{BSA}$ denotes the body surface area of individual $i$, $\mathrm{SEX}$ the patient's gender ($0/1$ for female/male),  $\mathrm{AGE}$ the patient's age (in years) and $\mathrm{BILI}$ the bilirubin concentration.
In addition to inter-individual variability and residual variability, interoccasion variability was included on two parameters, $\mathrm{VM}_\mathrm{EL}$ as well as the central volume of distribution $V_1$. An occasion was defined as the start of a  chemotherapeutic cycle $c$,
\begin{equation*}
\theta_{i,c} = \theta^\mathrm{TV}_i(\text{cov}) \cdot e^{\eta_i + \kappa_{i,c}}\,, \qquad \eta_i \iidsim \mathcal{N}(0,\Omega), \ \kappa_{i,c} \iidsim \mathcal{N}(0,\Pi)\,.
\end{equation*}

The system of ordinary differential equations (odes) describing rate of change of the amount in [$\mu$mol] of paclitaxel is given by
\begin{align*}
\frac{d \Cent}{dt} &= u(t) - \frac{\mathrm{VM}_\mathrm{EL}\cdot C_1 }{\mathrm{KM}_\mathrm{EL}+C_1} + k_{21} \text{Per1} - \frac{\mathrm{VM}_\mathrm{TR}\cdot C_1 }{\mathrm{KM}_\mathrm{TR}+C_1}  + k_{31}\text{Per2} - k_{13}\Cent\,, &\Cent(0) = 0 \\
\frac{d \text{Per1}}{dt} &= \frac{\mathrm{VM}_\mathrm{TR}\cdot C_1}{\mathrm{KM}_\mathrm{TR}+C_1} - k_{21}\text{Per1}\,, &\text{Per1}(0)=0\\
\frac{d \text{Per2}}{dt} &= k_{13}\Cent - k_{31}\text{Per2}\,, &\text{Per2}(0)=0\\
\end{align*}
where Cent refers to the central compartment, and Per1, Per2 to the first and second peripheral compartment, respectively, and $C_1(t) = \Cent/V_1$ denotes the concentration in plasma, $k_{13} = Q/V_1$ and $k_{31}=Q/V_3$, where $V_3$ denotes the volume of Per2, $u(t)$ is the dosing input, see also \SFigref{fig:VirtualPatient} (left part) for a schematic representation of the model.

\begin{table}
\begin{center}
\begin{tabular}{lll}
\begin{tabular}{lll}
\hline
\multicolumn{3}{c}{structural submodel}\\
\hline
$\mathrm{V_1}$ 				& 	10.8 		& $[\L]$\\
$\mathrm{V_3}$ 				& 	301 		& $[\L]$\\
$\mathrm{KM}_\mathrm{EL}$ 		& 	0.667 	& $[\mu \M]$\\
$\mathrm{VM}_\mathrm{EL,pop}$ 	& 	35.9  	& $[\mu \mathrm{mol}/h]$\\
$\mathrm{KM}_\mathrm{TR}$ 		& 	1.44	 	& $[\mu \M]$\\
$\mathrm{VM}_\mathrm{TR}$ 		& 	175   	& $[\mu \mathrm{mol}/h]$\\
$k_{21}$						&	1.12 	 	& $[1/h]$\\
$Q$							&	16.8		& $[1/h]$ \\
\hline
\multicolumn{3}{c}{covariate submodel}\\
\hline
$\theta_{\mathrm{VM}_\mathrm{EL}\text{-}\mathrm{BSA}}$ &  1.14 &\\
$\theta_{\mathrm{VM}_\mathrm{EL}\text{-}\mathrm{SEX}}$ &  1.07 &\\
$\theta_{\mathrm{VM}_\mathrm{EL}\text{-}\mathrm{AGE}}$ &  -0.447 &\\
$\theta_{\mathrm{VM}_\mathrm{EL}\text{-}\mathrm{BILI}}$  &  -0.0942 &\\
\end{tabular}
\hphantom{xxxx}&
&
\begin{tabular}{lll}
\hline
\multicolumn{3}{c}{statistical submodel IIV}\\
\hline
\\
$\omega^2_{V_3}$ 					&  0.1639 &\\
\\
$\omega^2_{\mathrm{VM}_\mathrm{EL}}$ &  0.0253 &\\
$\omega^2_{\mathrm{KM}_\mathrm{TR}}$ &  0.3885 &\\
$\omega^2_{\mathrm{VM}_\mathrm{TR}}$ &  0.077 &\\
$\omega^2_{k_{21}}$ 					&   0.008  &\\
$\omega^2_{Q}$ 					&  0.1660 &\\

\hline
\multicolumn{3}{c}{statistical submodel IOV}\\
\hline
$\pi^2_{V_1}$ 					&   0.1391 &\\
$\pi^2_{\mathrm{VM}_\mathrm{EL}}$ 					&  0.0231&\\
\hline
\multicolumn{3}{c}{statistical submodel RV}\\
\hline
$\sigma^2$ 						& 0.0317& \\
\end{tabular}
\end{tabular}
\end{center}
\caption{Pharmacokinetic parameter estimates of the previously published PK model \cite{Joerger2012a} for the anticancer drug paclitaxel (re-estimated by \cite{Henrich2017a}).}
\label{tab:PaclitaxelParameterEstimates}
\end{table}

% -----------------------------------------------------------------------------------------------------------------------------------------------------
\subsection{Bone Marrow Exhaustion model (PD model)}
\begin{figure}[bt]
\centering
\includegraphics[width =0.75\linewidth]{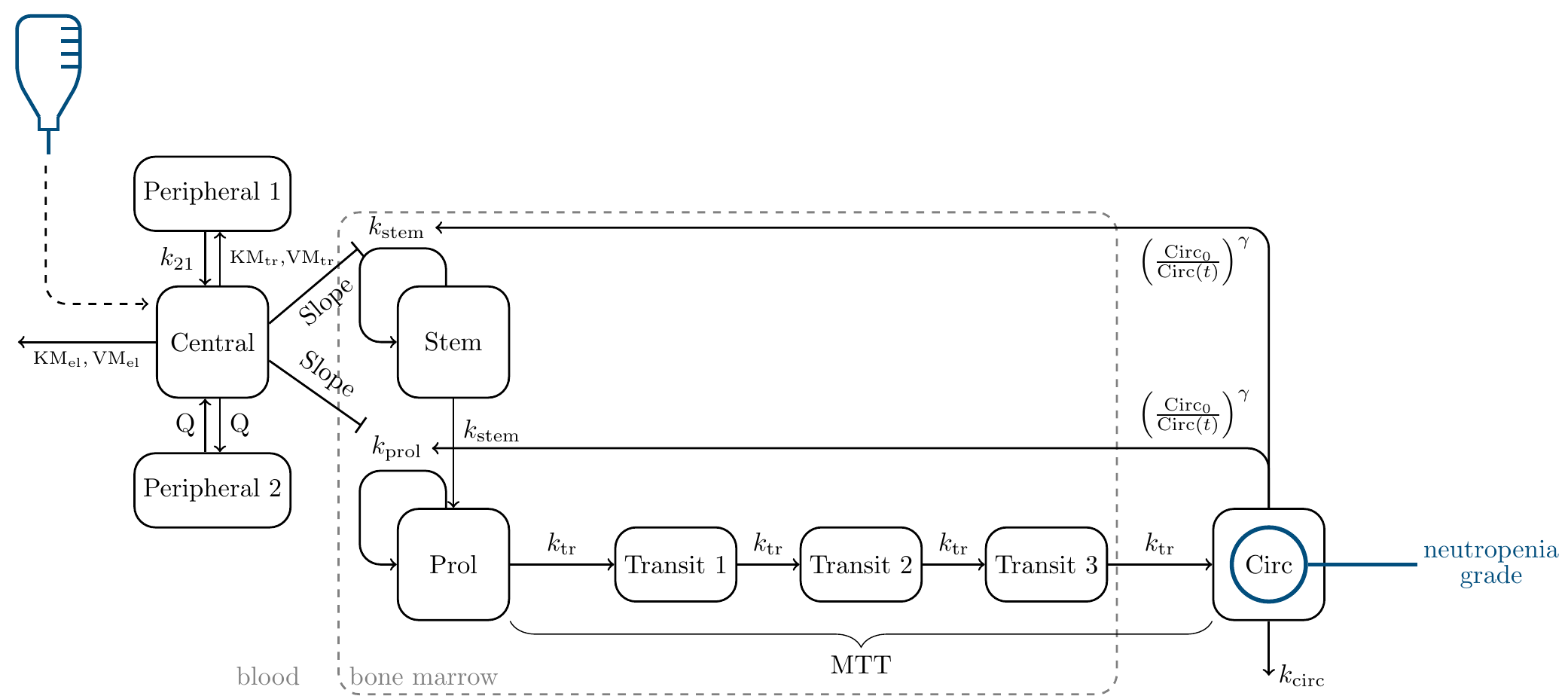}
\caption{\textbf{PK/PD model scheme for paclitaxel-induced neutropenia}. The mathematical model includes a PK model describing the pharmacokinetics of paclitaxel \cite{Joerger2012a} as well as a PD model describing the cytotoxic effect of the drug on the hematopoietic system \cite{Henrich2017a}. The proliferation rate of the stem cells $k_\text{stem}$ as well as of the proliferating cells $k_\text{prol}$ is affected by the drug. The observable is the concentration of neutrophils in the systemic circulation $\Circ$ (blue circle). From the lowest neutrophil concentration (nadir) the grade of neutropenia is inferred.}
\label{fig:VirtualPatient}
\end{figure}

% -----------------------------------------------------------------------------------------------------------------------------------------------------
In the CEPAC-TDM study \cite{Joerger2016a} cumulative neutropenia was observed, i.e., the lowest neutrophil concentration (nadir) as well as the maximum neutrophil concentration decreased over the course of treatment. 
A potential hypothesis for this cumulative behavior is that the drug also affects the long-term recovery of the bone marrow (bone marrow exhaustion). 
The gold-standard model for neutropenia by Friberg et al. \cite{Friberg2002} does not describe this long-term effect and was shown to overpredict neutrophil concentration at later cycles \cite[section 3.3]{Henrich2017Thesis}. 
Therefore, Henrich et al. \cite{Henrich2017a} extended the model to include a stem cell compartment, representing pluripotent stem cells with slower proliferation, which are also affected by the drug, see \SFigref{fig:VirtualPatient}.

The proliferation rates for the two compartments $\Prol$ and $\Stem$ are given by
\begin{align*}
\kprol &= \ftr \cdot \ktr \\
k_\mathrm{stem} &= (1-\ftr) \cdot \ktr \,, \\
\end{align*}
respectively, where $\ktr=4/\mathrm{MTT}$ denotes the transition rate constant of the maturation chain and $\ftr$ the fraction of input in Prol via replication. The baseline neutrophil count Circ$_0=$ANC$_0$ was inferred from the baseline data point $y_0$ (baseline method 2 in \cite{Dansirikul2008})
\begin{equation*}
\Circ_{0,i}= y_{0} \cdot e^{\sigma\cdot \eta_{\Circ_0,i}}\,, \qquad \eta_{\Circ_0,i} \sim \mathcal{N}(0,1)\,.
\end{equation*}

The system of odes describing the structural model reads
 \begin{align*}
 \frac{d \Stem}{dt} &= k_{\text{stem}}\Stem \cdot (1-\Edrug) \cdot \Big( \frac{\Circ_0}{\Circ} \Big)^\gamma - k_\text{stem} \Stem \,, &\Stem(0) = \Circ_0 \\
\frac{d \Prol}{dt} &= k_{\text{prol}}\Prol \cdot (1-\Edrug) \cdot \Big( \frac{\Circ_0}{\Circ} \Big)^\gamma + k_\text{stem} \Stem - k_\text{tr} \Prol\,, &\Prol(0) = \Circ_0\\
\frac{d \text{Transit1}}{dt} &= k_\text{tr}\Prol - k_\text{tr} \text{Transit1}\,, &\text{Transit1}(0)=\Circ_0\\
\frac{d \text{Transit2}}{dt} &= k_\text{tr}\text{Transit1} - \ktr \text{Transit2}\,, &\text{Transit2}(0)=\Circ_0\\
\frac{d \text{Transit3}}{dt} &= k_\text{tr}\text{Transit2} - \ktr \text{Transit3}\,, &\text{Transit3}(0)=\Circ_0\\
\frac{d \Circ}{dt} &= \ktr \text{Transit3} - k_\text{circ} \Circ \,, &\text{Circ}(0)=\Circ_0\\
\end{align*}
where $k_\text{circ}=\ktr$ and $\Edrug = \Slope \cdot C_1$ the linear drug effect.
Note that the model implicitly assumes that the volumes of all compartments are identical.
Henrich et al. \cite{Henrich2017a} calibrated this model in a population analysis to the CEPAC-TDM study data \cite{Joerger2016a}.

\begin{table}[bt]
\begin{center}
\begin{tabular}{lll}
\hline
\multicolumn{3}{c}{Structural submodel}\\
\hline
$\Circ_0$	&	baseline method & $[10^9 \cells/\L]$\\
$\MTT$	&	145 & $[h]$ \\
$\Slope$	&	13.1 & $[1/\mu\M]$\\
$\gamma$		&	0.257 &\\
$\mathrm{ftr}$     	& 	0.787 & \\
\hline
\multicolumn{3}{c}{Statistical submodel}\\
\hline
$\omega^2_{\Slope}$ & 0.2007& \\
$\sigma^2$		& 0.2652 &\\
\end{tabular}
\end{center}
\caption{Parameter estimates for the bone marrow exhaustion model and the anticancer drug paclitaxel, taken from \cite{Henrich2017a}. }
\label{tab:HenrichParameterEstimates}
\end{table}

\newpage
% =====================================================================================
\section{PK-guided dosing}\label{sec:JoergerAlg}
% =====================================================================================

In order to put our proposed approaches into context, we compared the results with previously applied dosing algorithms.
For the sake of completeness, we repeat here the algorithm developed and applied by Joerger et al. \cite{Joerger2012a,Joerger2016a} (called \textit{PK-guided dosing}) as we used it for comparison throughout the manuscript. 
In the \textit{PK-guided dosing}, the dose of the first cycle is determined based on age and sex. 
For subsequent cycles the dose is adjusted according to exposure (time during which the drug concentration is above 0.05\,$\mu\M$) and neutropenia grade observed in the previous cycle (inferred from observation at day 15).
Thus, the algorithm is not completely \textit{offline}, since the exposure measure is inferred online using the PK model. The PD-based adaptations, which were the focus in this study, however, were performed offline.

\begin{figure}[H]
\centering
\includegraphics[width =1\linewidth]{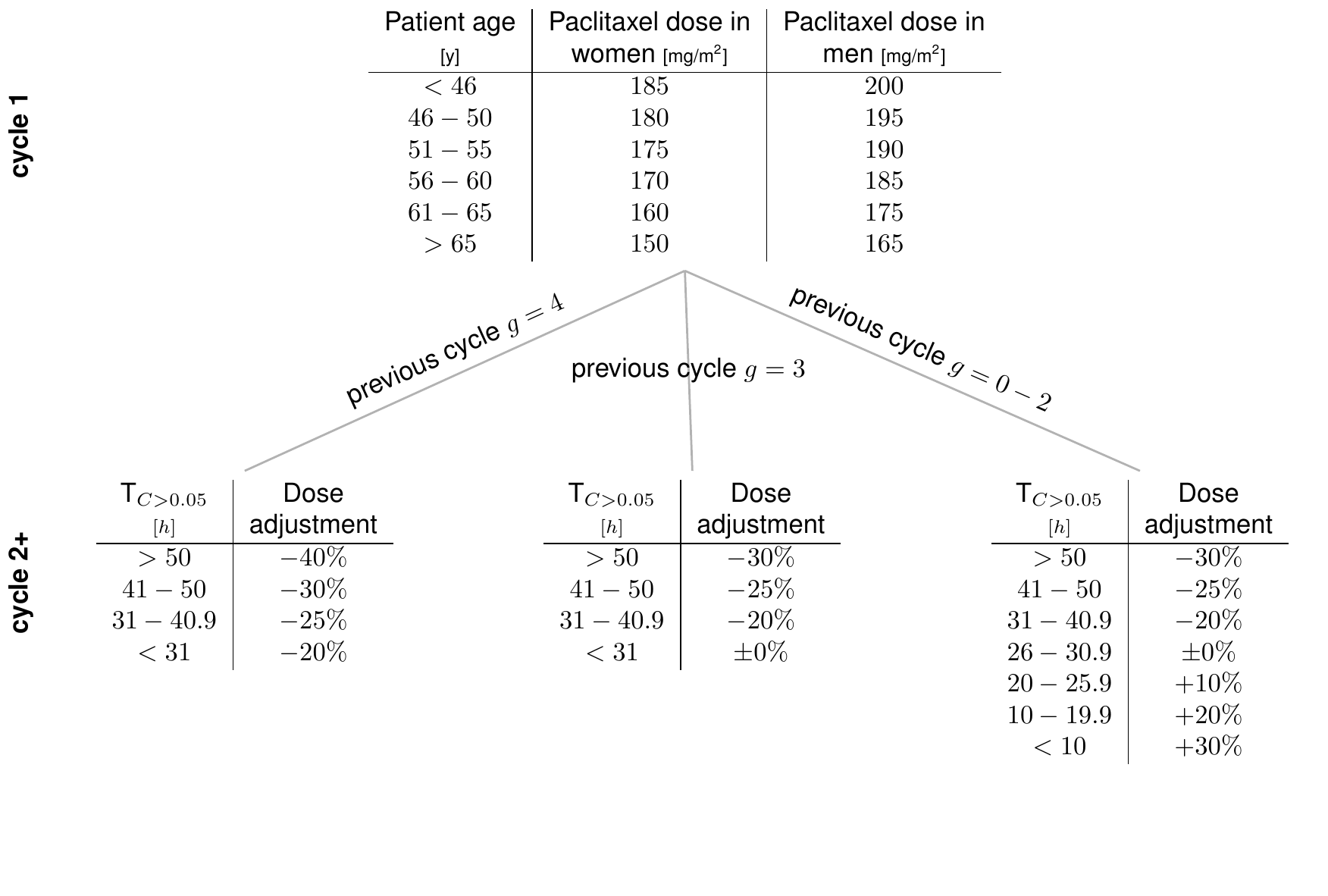}
\caption{\textbf{Diagram of the PK-guided dosing algorithm adopted from \cite{Joerger2012a}}.}
\label{fig:JoergerTree}
\end{figure}

\newpage
% =====================================================================================
\section{MAP-guided dosing}\label{sec:MAPMIPD}
% =====================================================================================

MAP-guided dosing is widely applied in various therapeutic areas for online therapy individualization \cite{Bleyzac2001,Wallin2009} and implemented in assorted software tools, e.g. TDMx \cite{TDMx}, InsightRX \cite{InsightRX}.
The optimal dose is determined in a two-step procedure:
\begin{enumerate}
\item The MAP-estimate $\hat{\theta}_{1:c}^\text{MAP}$ is computed by minimizing the negative log-posterior over the observed data points
\begin{equation*}
\hat{\theta}^\text{MAP}_{1:c} = \underset{\theta}{ \arg \min}\, -\!\log p(\theta|y_{1:c})\,,
\end{equation*}
where $y_{1:c}$ denotes the patient-specific TDM data collected up to the end of cycle $c$. The posterior is proportional to the likelihood times the prior, i.e., $p(\theta|y_{1:c}) \propto p(y_{1:c}|\theta)\cdot p(\theta)$.
Often, the prior $p(\theta)$ is chosen to be lognormal and the likelihood is chosen to be gaussian for log-transferred data (additive error model on log-scale). 
The method is outlined in more detail in \cite{Sheiner1979,Maier2020}.
\item Then, the MAP estimate is used to generate model predictions for solving the objective problem in the dose selection at start of cycle $c+1$
\begin{equation*}
d_{c+1}^* = \underset{d}{ \arg \min} - R\big(c_\text{nadir}\big(\hat{\theta}^\text{MAP}_{1:c},d\big)\big),
\end{equation*}
where we investigated different evaluation functions $R(\cdot)$ based on the MAP-based nadir concentration, see Section~S~\ref{sec:ResultsMAP}. Typically, a utility function is maximized or the deviation to a target concentration is minimized.
\end{enumerate}
MAP-guided dosing depends largely on the reliability of MAP-based predictions which, however, do not necessarily represent the most probable therapeutic outcomes and neglect model uncertainties \cite{Maier2020}. 
In particular, any distributional information like, e.g., the tails of the posterior distribution which describe sub-therapeutic as well as toxic ranges, are completely neglected.
These tails provide crucial information for dose selection and might be pronounced due to the often chosen lognormal prior parameter distributions.
In addition, the choice of the target or utility function has a crucial impact on the optimal dose selection. 
While the concept of an utility would be quite desirable, the definition is rather challenging since clinically rather therapeutic ranges are observed.

% =====================================================================================
\section{DA-guided dosing}\label{sec:DAMIPD}
% =====================================================================================

DA-guided dosing is based on Bayesian forecasting of the therapeutic outcome using sequential Bayesian DA approaches \cite{Maier2020}.
There exist many sequential but also variational DA algorithms which could be used, see e.g., \cite{ReichCotter2015,Law2015,Sarkka2013}.
We employed in the simulation studies a basic particle filtering and smoothing algorithm \cite{Gordon1993,Arulampalam2002,Doucet2008} with resampling and rejuvenation ($M=100$).
% TOO MUCH DETAIL HERE: , if the effective ensemble size was lower than half the initial ensemble size . 
Particle filters are well suited to the setting in pharmacometrics with nonlinear models and non-gaussian prior distributions as they do not make an assumption on the distribution of the posterior and provide a correct estimation of the posterior in the limit of large sample sizes ($M \rightarrow \infty$, without rejuvenation) \cite{SmithGelfand1992,Gordon1993,Law2015}.
We considered the setting of an augmented state space, adding the model parameters to the model states to simultaneously update the states and parameters sequentially.
Smoothing was realised via resampling or reweighting of the state history.
The used DA approach is outlined in all detail in \cite{Maier2020}. 

At decision time points $T_c$ the current posterior ensemble is used to predict the therapy outcome under posterior uncertainty. 
Thus, the probabilities of subtherapeutic/toxic outcomes can be computed and integrated in the optimal dose selection allowing to simultaneously integrate efficacy and safety aspects into the dose selection.
Note that the dose finding problem is a multi-objective optimization problem, and that it is not possible to simultaneously  decrease the probability of grade 4 and the probability of grade 0.

\newpage
% =====================================================================================
\section{RL-guided dosing}\label{sec:RLOverview}
% =====================================================================================

There exist many different RL algorithms for computing the action-value function \cite{SuttonBarto2018,Bertsekas2019,Yu2019}. We have restricted our analysis to two popular and powerful methods with relevance in clinical decision-making: Monte Carlo tree search is a popular choice for episodic tasks (finite time horizon) \cite{KocsisSzepesvari2006}, and Q-learning \cite{WatkinsDayan1992} for continuous and episodic tasks (infinite time horizon). As both types of tasks---episodic and continuous---are of relevance in dosing policies we considered both here. Due to our specific application setting (episodic task of six treatment cycles), however, our focus is on MCTS.

\begin{table}
\begin{tabular}{ll}
\multicolumn{2}{l}{Common notation}\\
\hline
$\pi$ & generic policy \\
$q_\pi$ & expected return given generic policy $\pi$\\
\\
Notation specific to MCTS+UCT & \\
\hline
$\pi_{k}$ & policy in training phase\\
$q_k$ & sample approximation of $\hat{q}_{\pi_k}$ in training phase\\
$\hat{\pi}_\text{UCT}:=\pi_K$ & policy after training phase (incl. exploration)\\
$\hat{q}_{\pi_\text{UCT}}:=q_K$ & sample approx at the end of training phase\\
$\pi^* = \arg \max \hat{q}_{\pi_\text{UCT}}$ & RL-guided dosing policy (clinical setting, no exploration)\\
\\
Notation specific to MCTS+PUCT & \\
\hline
$q_{\pi_0} :=\hat{q}_{\pi_\text{UCT}}$ & prior estimated return\\
$\pi^{1:c}_k$ & policy in training phase using ensemble $\mathcal{E}_{1:c}$\\
$q^{1:c}_k$ & sample approximation in training phase using ensemble $\mathcal{E}_{1:c}$\\
$\hat{\pi}^{1:c}_\text{PUCT}:=\pi^{1:c}_K$ & policy after training phase using ensemble $\mathcal{E}_{1:c}$ (incl. exploration)\\
$\hat{q}_{\pi^{1:c}_\text{PUCT}}:=q^{1:c}_K$ & sample approx at the end of training phase using ensemble $\mathcal{E}_{1:c}$\\
$\pi^{*} = \arg \max \hat{q}_{\pi^{1:c}_\text{PUCT}}$ & DA-RL-guided dosing policy (clinical setting, no exploration)\\
\end{tabular}
\caption{Notation related to the reinforcement learning approaches used in RL-guided and DA-RL-guided dosing.}
\end{table}

% -----------------------------------------------------------------------------------------------------------------------------------------------------
\subsection{Monte Carlo tree search (MCTS) with UCT}\label{sec:MCTSAlg}
% -----------------------------------------------------------------------------------------------------------------------------------------------------
Monte Carlo tree search combines the Monte Carlo method with tree search. It was mainly developed and applied to game-tree search, e.g., AlphaGo \cite{Silver2016}. To efficiently explore the search tree (Figure~2) we used as tree policy the upper confidence bound applied to trees (UCT) \cite{Auer2002,KocsisSzepesvari2006}. MCTS comprises four recursive steps which are repeated in each episode $k=1,\dots,K$ for building a search tree based on random samples in the decision space \cite{Browne2012} (see Figure~S~\ref{fig:MCTSTree}):

\begin{figure}[H]
\centering
\includegraphics[width =0.75\linewidth]{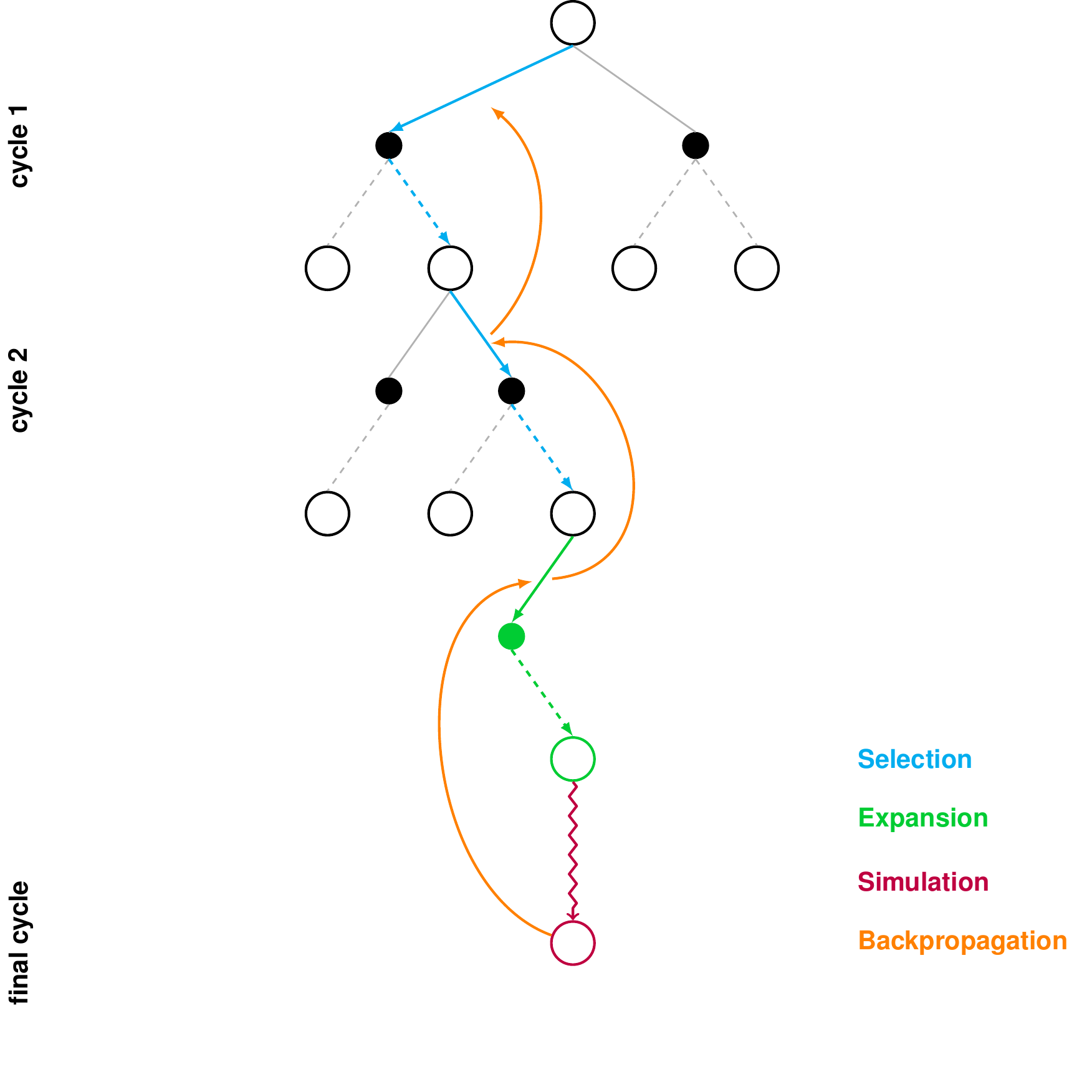}
\caption{\textbf{Illustration of Monte Carlo tree search (MCTS)}. 1.) Selection: within the search tree, actions are selected according to the tree policy, until a not yet fully expanded node is reached. 2.) Expansion: The action is chosen among the unvisited actions according to a roll-out policy (e.g., random) and the new node is added to the search tree. 3.) Simulation: From the new child node, a simulation is started using the roll-out policy until the end of the episode/therapy. 4.) Back-propagation: The return is back-propagated through the tree, i.e., action-values functions and number of visits are updated for each selected state-action pair within the search tree. }
\label{fig:MCTSTree}
\end{figure}

\begin{itemize}
\item \textit{Selection:} Starting at the root node $s_0$ actions are selected according to the \textit{tree policy} ($\pi_k$) until a not yet fully expanded node, i.e., a node with an unvisited action, is reached.
\item \textit{Expansion:} If the selected node is expandable (nonterminal state), one child node is added by selecting an unvisited action.
\item \textit{Simulation:} Following a default policy (often random) a single simulation is run until a terminal state is reached. The return of this episode is calculated. 
\item \textit{Backpropagation:} The simulated return is backpropagated to the selected nodes in the search tree. More specifically, \textit{back up} means we compute incrementally the expected return ($q_k$) via a running mean, i.e., in episode $k$ 
\begin{equation*}
q_{k}(s_c,d) = q_{k-1}(s_c,d)+\frac{1}{N_k(s_c,d)}\Big( G_c^{(k)}-q_{k-1}(s_c,d)\Big)\,.
%M_{2,k}(s_c,d) &= M_{2,k-1}(s_c,d)+(g_c^{(k)}-q_{k-1})(G_c^{(k)}-q_{k})\,, 
\end{equation*}
%where the variance can be calculated via $s^2_k(s_c,d)=M_{2,k}(s_c,d)/(k-1)$ or $\sigma^2_k(s_c,d)=M_{2,k}(s_c,d)/k$.
\end{itemize}
Note that we consider a basic version of MCTS and various modifications are possible, e.g., adding several child nodes in the expansion step or running multiple simulations in parallel.
More details about the implementation and chosen tuning parameters are provided in Section~S~\ref{sec:SimStudyRL}.
In brief, in our setting for each episode $k$, a virtual patient is generated with covariates cov$^{(k)}$  in the corresponding covariate class, say $\mathcal{COV}_l$, sampled according to the obsefved covariate distributions of the CEPAC-TDM study, and the individual model parameters $\theta^{(k)}\sim p_\Theta\big(\hatthetaTV(\cov^{(k)}),\Omega\big)$ are sampled from the corresponding prior parameter distributions.

The UCT algorithm is based on Hoeffding's inequality \cite{Hoeffding1963}. We consider the setting of  \cite[Theorem 2]{Hoeffding1963}: Let $X_1,\dots,X_n$ be independent, bounded random variables, i.e., $a_i\leq X_i \leq b_i$ with sample mean $\bar{X}= (X_1+\ldots+X_n)/n$, then
\begin{equation}
P\big[E[X]>\bar{X}+u\big] \leq \exp\left(\frac{-2n^2u^2}{\sum_{i=1}^n (b_i-a_i)^2}\right)\,.
\end{equation}

Translating this into the RL setting for $X=G$ \cite{Auer2002,Silver2017GO} 
\begin{equation}
P[q_\pi(s,d)>q_k(s,d)+U_k(s,d)] \leq \exp\left(- \frac{2N_k(s,d)^2U_k(s,d)^2}{\sum_i (b_i-a_i)^2}\right)\,,
\end{equation}
Choose probability $\alpha$ that the true value exceeds the upper bound
\begin{equation}
\exp\left(- \frac{2N_k(s,d)^2U_k(s,d)^2}{\sum_i (b_i-a_i)^2} \right) = \alpha
\end{equation}
leads to the upper bound
\begin{equation}
U_k(s,d) = \sqrt{ (b-a)^2}\cdot \sqrt{ \frac{ - \log \alpha }{2N_k(s,d)}}\,,
\end{equation}
with $b=b_i$ and $a=a_i$ for all $i=1,\ldots,N_k(s,d)$, i.e., the maximum and minimum return does not change.
In the bandit literature, it was shown that  $\alpha = N_k(s)^{-4}$ ensures  logarithmic regret \cite{Auer2002}, where $N_k(s):=\sum_{d'} N_k(s,d')$ the number of visits of state $s$ and leads to the UCB1 algorithm for the case $[a_i,b_i]=[0,1]$ for $i=1,\dots,n$ \cite{Auer2002} with upper bound
\begin{equation}\label{eq:U_UCB}
U_k(s,d) = \sqrt{\frac{2 \log N_k(s)}{N_k(s,d)}}\,.
\end{equation}
In \cite{Silver2017GO} the bound
\begin{equation}\label{eq:U_Silver}
U_k(s,d) = \sqrt{\frac{ N_k(s)}{1+N_k(s,d)}}
\end{equation}
was used, which corresponds to 
\begin{displaymath}
\alpha = \exp\left(-2N_k(s,d)\cdot \frac{N_k(s)}{1+N_k(s,d)}\right) \approx \exp\left(-2N_k(s)\right)
\end{displaymath}
for the bounds $[0,1]$. 
Note, the adding 1 in the denominator of Eq.~\eqref{eq:U_Silver} avoids division by zero, i.e., if the dose has not been taken before. 
The UCB1 Algorithm is actually initialized by taking each action once; this is, however, not possible in our tree setting.
The bound in Eq.~\eqref{eq:U_Silver} encourages exploration as the numerator is larger than in Eq.~\eqref{eq:U_UCB}.
As we are in a pure model-based learning setting, we want to encourage exploration as we do not have to pay a price for a suboptimal outcome, i.e., we are not interested in maximizing the cumulative reward but in learning an optimal policy.
Therefore, we considered (in our learning setting) the bound 
\begin{equation}
U_k(s,d) = \sqrt{ (b-a)^2}\cdot \sqrt{\frac{ N_k(s)}{1+N_k(s,d)}}\,,
\end{equation}
where the first factor is included in the exploration-exploitation parameter $\epsilon$, see also Eq.~\eqref{eq:epsilon}. 
When model-based learning is completed and one would aim at continued learning from real patients one should rather choose a more cautious exploration strategy as in~\eqref{eq:U_UCB}.

% -----------------------------------------------------------------------------------------------------------------------------------------------------
\subsection{Q-learning}\label{sec:Qlearn}
% -----------------------------------------------------------------------------------------------------------------------------------------------------
As we considered an episodic task of six treatment cycles, we employed MCTS. For long-term therapy plans, however, temporal difference approaches based on a one step look-ahead approach could be beneficial as they do not require the computation of the total return (requiring simulation until the terminal state). The choice of algorithm is therefore problem dependent. 
Q-learning is based on the decomposition of the action-value function into an immediate reward and adiscounted action-value of successor state and action (Bellman)
\begin{equation*} 
q_\pi(s,d) = \mathbb{E}_\pi[R_{c+1}+\gamma q_\pi(S_{c+1},D_{c+1})|S_c=s,D_c=d]\,. 
\end{equation*}
In Q-learning the action-value function is also learned iteratively
\begin{equation}
q_{k+1}(s_c,d) \leftarrow q_k(s_c,d)+ \alpha \cdot \big(R_{c+1}+\gamma \cdot \underset{d'}{\text{max}} \ [q_k(s_{c+1},d')] - q_k(s_c,d)\big)\,,
\end{equation}
where $s_{c+1}$ is the next state when giving dose $d$ in state $s_c$.
Parameters which need to be specified are
\begin{itemize}
\item learning rate $\alpha$
\item exploitation and exploration parameter $\epsilon$
\end{itemize}
To ensure convergence, the learning rate needs to decay appropriately with the number of iterations, e.g., as a Robbins-Monro sequences \cite{RobbinsMonro1951}. Similarly as in MCTS, the learner is confronted with a trade-off between exploration and exploitation. In Q-learning, frequently the $\epsilon$-greedy approach is used: With probability $\epsilon$, a random action is chosen and with probability $1-\epsilon$, the greedy action, i.e., the current argmax of $q_\pi$, is chosen. Also $\epsilon$ can be chosen in a decreasing manner, to encourage exploration in the beginning and exploitation at later training episodes.

% =====================================================================================
\section{Combining DA and RL: DA-RL-guided dosing}\label{sec:DARLbackground}
% =====================================================================================

DA can be integrated into RL in two ways, (i) by improving the state representation, and (ii) by using the posterior ensemble in a decision-time planning procedure to update and individualize the estimate of the $q_\pi$-values reflecting the posterior uncertainty.

First, before any patient is treated, a prior dosing policy $\hat{q}_{\pi_\text{UCT}}$ is planned, i.e., determined, via model-based RL, e.g., via MCTS+UCT as in Section~S~\ref{sec:MCTSAlg}.

When a patient is to be treated, the ensemble $\mathcal{E}_0$ for the sequential DA algorithm, e.g., particle filter/smoother, is initialized.
The patient-specific TDM data $y_{1:c}$ is integrated, leading to an updated posterior particle ensemble $\mathcal{E}_{1:c}$. 
At a decision time point $T_c$, the posterior expectation is computed for an improved estimate of the current patient state, e.g., a sample approximation to the posterior expectation of a nadir concentration
\begin{equation*}\label{eq:postexp}
\hat{c}_\text{nadir}=\sum_{m=1}^M w_c^{(m)} \cdot  c_\text{nadir}\left(x_{1:c}^{(m)},\theta^{(m)}\right) \,,
\end{equation*}
where $c_\text{nadir}(x_{1:c}^{(m)},\theta^{(m)})$ denotes the minimum neutrophil concentration of the $m$-th particle within the cycle.
The posterior expected nadir $\hat{c}_\text{nadir}$ is translated to the corresponding neutropenia grade of the cycle $g_c$ and used to update the current patient state $s_c$.
A MCTS search tree is initialized at the current patient state $s_c$ and the search within the tree is guided by the PUCT algorithm \cite{Silver2017GO}, where prior probabilities of choosing a dose are computed from the prior $\hat{q}_{\pi_\text{UCT}}$-values, see Eq.~(17). 
For model simulations within each episode in the MCTS the model state parameter vector $x^{(k)}$ and $\theta^{(k)}$ is sampled from the posterior particle ensemble $\mathcal{E}_{1:c}$.

% =====================================================================================
\section{Simulation study: neutrophil guided dosing of paclitaxel over multiple cycles}\label{sec:SimStudy}
% =====================================================================================

% -----------------------------------------------------------------------------------------------------------------------------------------------------
\subsection{Setting} \label{ssec:setting}
% -----------------------------------------------------------------------------------------------------------------------------------------------------
The simulation study was performed in MATLAB R2017b/2018b using the previously described PK/PD model \cite{Henrich2017a}. 
%58 words
In the offline approaches, the second neutrophil measurement (at day 15 of the cycle) is used to infer the grade of neutropenia (according to the common terminology criteria for adverse events \cite{CTCAE}, see also Figure~3).
The generated virtual patient populations for training and testing were sampled based on the reported covariate ranges in the CEPAC-TDM study \cite{Joerger2016a}. For the standard dosing approach, we employed the rules applied in the CEPAC-TDM study arm A, i.e., $200\,\text{mg}/\text{m}^2 \, \text{BSA}$ and a 20 \% dose reduction if grade 4 neutropenia was observed \cite{Joerger2016a}. Note that in clinical practice, the dose was also reduced if other severe adverse effects were observed, which was not included in our simulation setting.
To save computational time, we only optimized over the next cycle (rather than over all remaining cycles) in the MAP-guided and DA-guided dosing. 
In MAP-guided dosing the sensitivities for gradients used in the MATLAB solver \texttt{fmincon} were computed using the Toolbox AMICI \cite{AMICI_doi,Frohlich2016a} which requires a MATLAB version $<$ R2018.
In RL-guided dosing we employed MCTS to exploit the characteristics of an episodic task (six treatment cycles) instead of Q-learning. 

% -----------------------------------------------------------------------------------------------------------------------------------------------------
\subsection{Comparison with reported CEPAC-TDM study outcomes}\label{sec:ComparisonCEPACTDM}
\begin{figure}[H]
\centering
\includegraphics[width =1\linewidth]{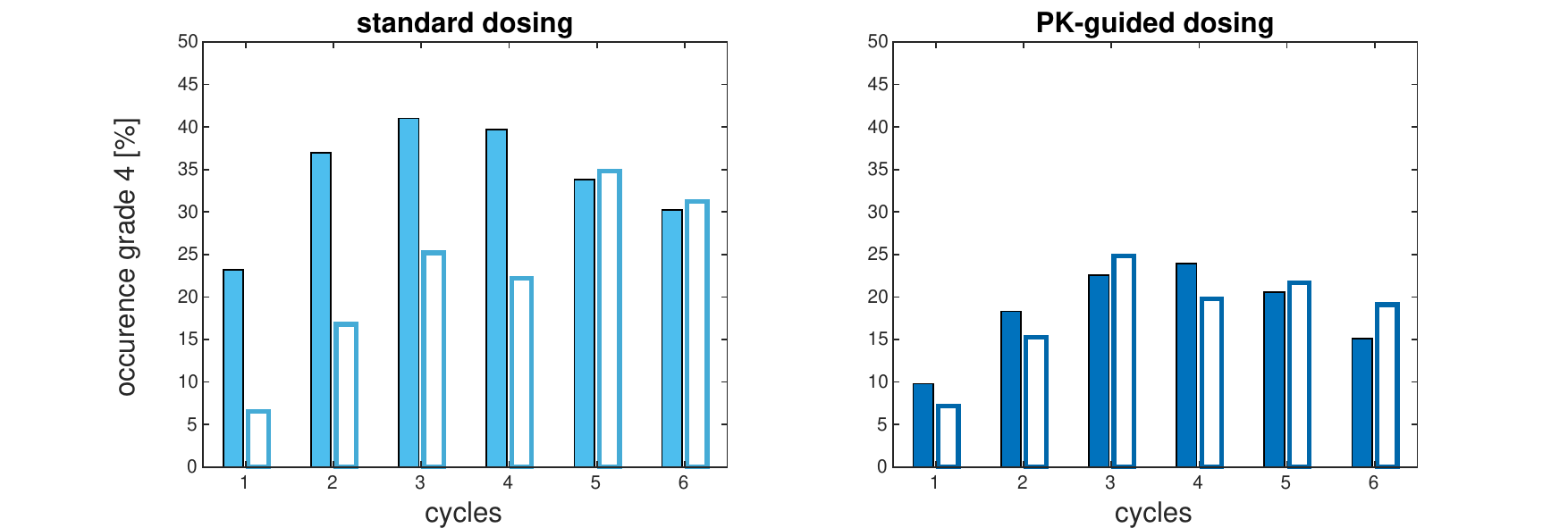}
\caption{\textbf{Comparison of  predicted grade 4 neutropenia with observed occurrence in the CEPAC-TDM study}. Blue bars show the results from the simulation study (based on day 15 observation) and white bars show the results from the CEPAC-TDM study. Clearly, the occurrence of grade 4 is overestimated for the standard dosing in the simulation study, but comparable for the PK-guided dosing. The results for the CEPAC-TDM study were retrieved from \cite{Henrich2017Thesis}. }
\label{fig:CEPAC_TDM_Comparison}
\end{figure}

% -----------------------------------------------------------------------------------------------------------------------------------------------------
In our simulation study, we followed the design of the CEPAC-TDM study. 
To put the simulation results into perspective, we compared the simulated occurrence of grade 4 neutropenia (based on simulated observations on day 15 including residual variability) with the observed occurrence in the CEPAC-TDM study for the standard dosing (arm A) and the PK-guided dosing algorithm (arm B), see Figure~S~\ref{fig:CEPAC_TDM_Comparison}. 
We observed that we overpredict the occurrence of grade 4 neutropenia for standard dosing (left panel). 
This was attributed to the fact that the standard dosing in arm A, the dose was also decreased if non-haematological toxicities occurred (see also comment at the end of Section~\ref{ssec:setting}). 
Since our model did only allow to simulate neutropenia we could not take further aspects into account. 
For the PK-guided dosing algorithm (right panel), the simulation results were well aligned with the observed results in the CEPAC-TDM study.
The occurrence of grade 4 neutropenia was comparable across all cycles. 
Additional characteristics of the clinical study that we did not take into account in our simulation study are drop-outs, adherence to the dosing instructions (e.g. the given relative dose in the first cycle in study arm A ranged from $150\,\text{mg}/\text{m}^2$ to $215\,\text{mg}/\text{m}^2$) and comedication (e.g. therapeutic GCSF). 

% -----------------------------------------------------------------------------------------------------------------------------------------------------
\subsection{Observation time points}
% -----------------------------------------------------------------------------------------------------------------------------------------------------
Observation time points of neutrophil concentrations were chosen in accordance with the CEPAC-TDM study \cite{Joerger2016a}:  day of the dose administration as well as day 15 of each cycle.
However, in the evaluation of the dosing algorithm the average model predicted nadir time (based on the Friberg model) was found to be on day 11.5 \cite{Joerger2012a}.
Therefore, we investigated also day 12 als alternative sampling time point.

\begin{figure}[H]
\centering
\includegraphics[width =1\linewidth]{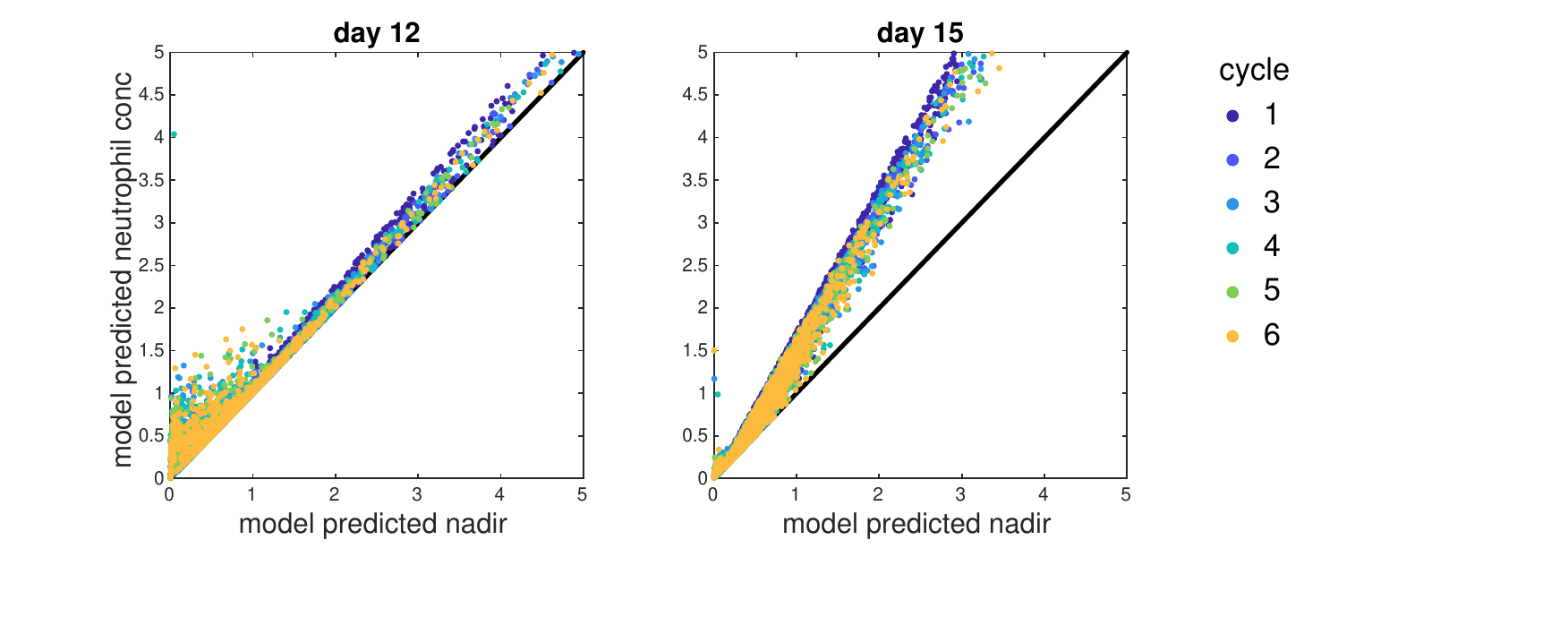}
\caption{\textbf{Comparison of the model predicted nadir concentration compared to the model predicted neutrophil concentration at days 12 and 15}. The standard dosing was used for simulation of the neutropenia time-courses for 1000 virtual patients.  }
\label{fig:Day12_15}
\end{figure}

In Figure~S~\ref{fig:Day12_15}, we examined the correlation between the model predicted nadir (based on the BME model) and the simulated neutrophil concentration at day 12 and 15. 
For larger nadir concentrations (nadir $> 1\cdot 10^9 \text{cells}/\text{L}$) the neutrophil concentrations at day 15 clearly overpredict the true nadir, i.e., underpredicts the severity of neutropenia. 
For small nadir concentrations (nadir $\leq 1\cdot 10^9 \text{cells}/\text{L}$) the correlation between model predicted nadir and model predicted neutrophil concentration at day 15 seems to be better.
This information could be relevant for future studies and demonstrates the importance of optimal sampling time points and the benefit of an model-informed analysis. 

\begin{figure}[H]
\centering
\includegraphics[width =1\linewidth]{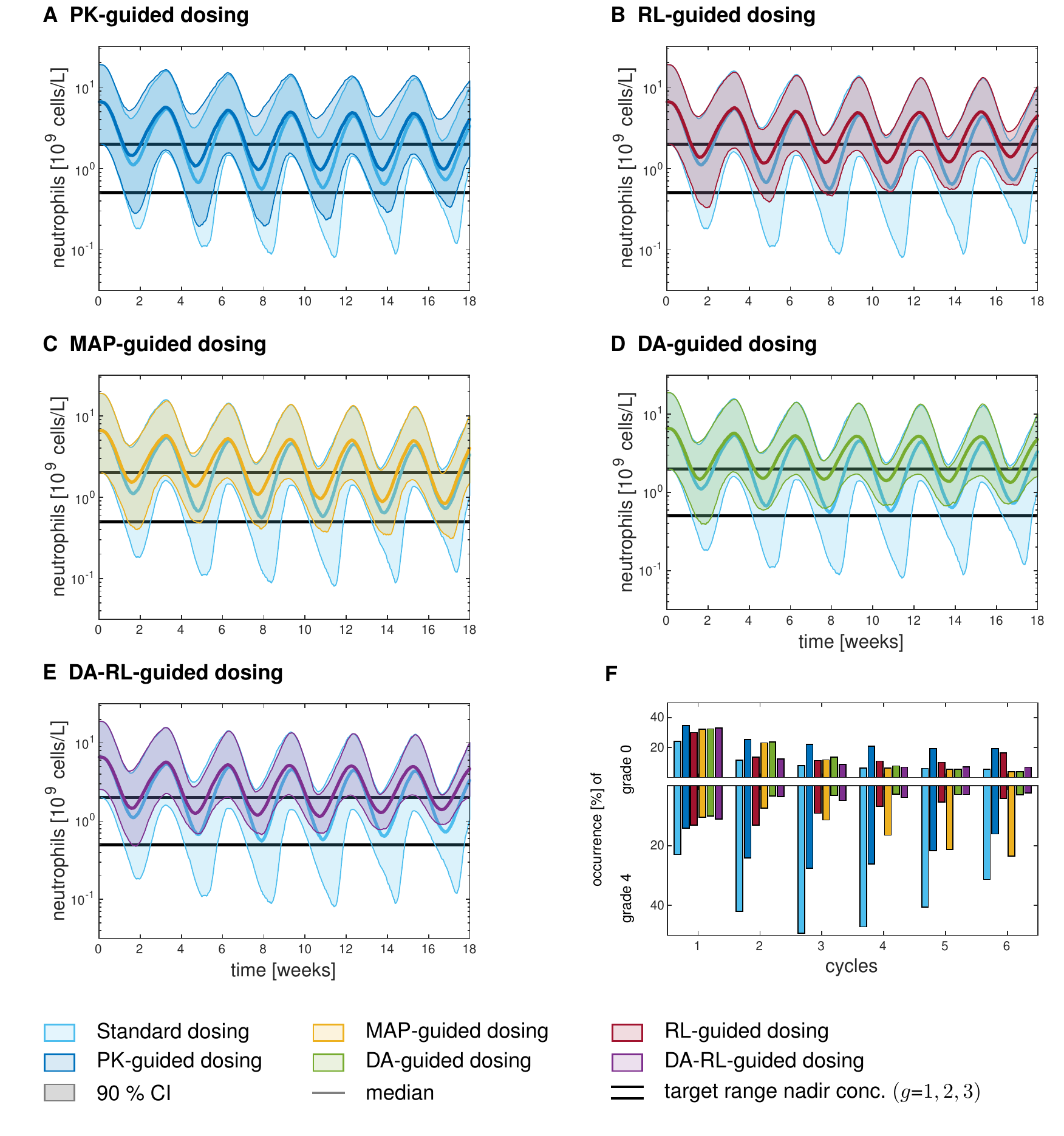}
\caption{\textbf{Comparison of different dosing policies for paclitaxel dosing based on observations at day 12}. Comparison of the 90 \% confidence intervals and median of the neutrophil concentration for the test virtual population ($n_\text{test}=1000$) using (\textbf{A}) the standard dosing and PK-guided dosing (\textbf{B}) MAP-guided dosing, (\textbf{C}) DA-guided dosing, (\textbf{D}) RL-guided dosing and  (\textbf{E}) DA+RL-guided dosing. (\textbf{F}) Occurrence of grade 0 and grade 4 across the different dosing policies for the test population over the six cycles.}
\label{fig:Sim1ComparisonDay15}
\end{figure}

% -----------------------------------------------------------------------------------------------------------------------------------------------------
\subsection{PK-guided dosing}\label{sec:SimJoerger}
\begin{figure}[H]
\centering
\includegraphics[width =1\linewidth]{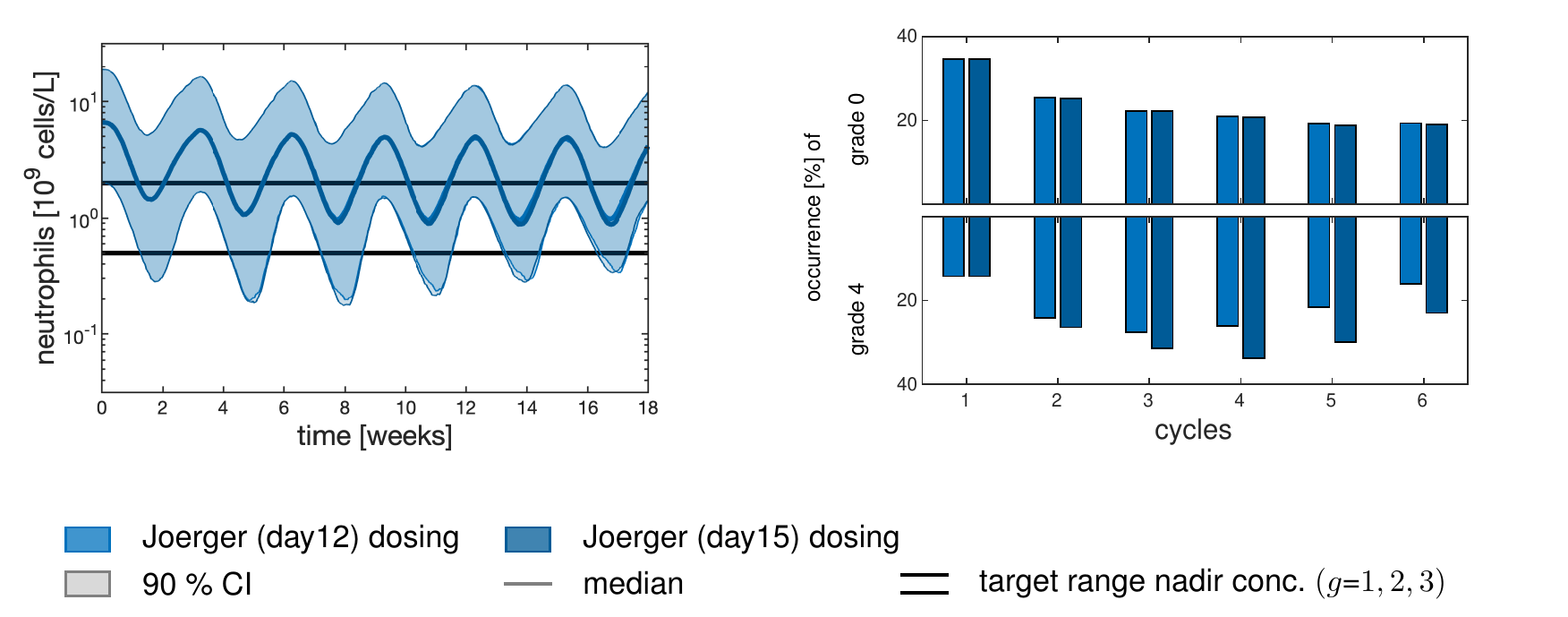}
\caption{\textbf{Comparison of the results when PK-guided dosing was based on the neutrophil measurement at day 12 or day 15}.   }
\label{fig:Joerger15}
\end{figure}

% -----------------------------------------------------------------------------------------------------------------------------------------------------
We also compared the effect of the sampling time point on the PK-guided dosing algorithm by applying the algorithm to the test virtual population, see Figure~S~\ref{fig:Joerger15}. 
For this the neutropenia grade of the previous cycle was inferred either based on simulated neutrophil measurements at day 12 or day 15 in the previous cycle (including residual variability). 
The occurrence of neutropenia grade 4 (evaluated based on model predicted nadir) was slightly higher if the previous cycle grade was inferred from the measurement at day 15 compared to day 12. 
Thus, the sampling time point has an effect on the PK-guided dosing algorithm and a sampling time point around day 12 is advantageous.

% -----------------------------------------------------------------------------------------------------------------------------------------------------
\subsection{MAP-guided dosing}\label{sec:ResultsMAP}
\begin{figure}[H]
\centering
\includegraphics[width =1\linewidth]{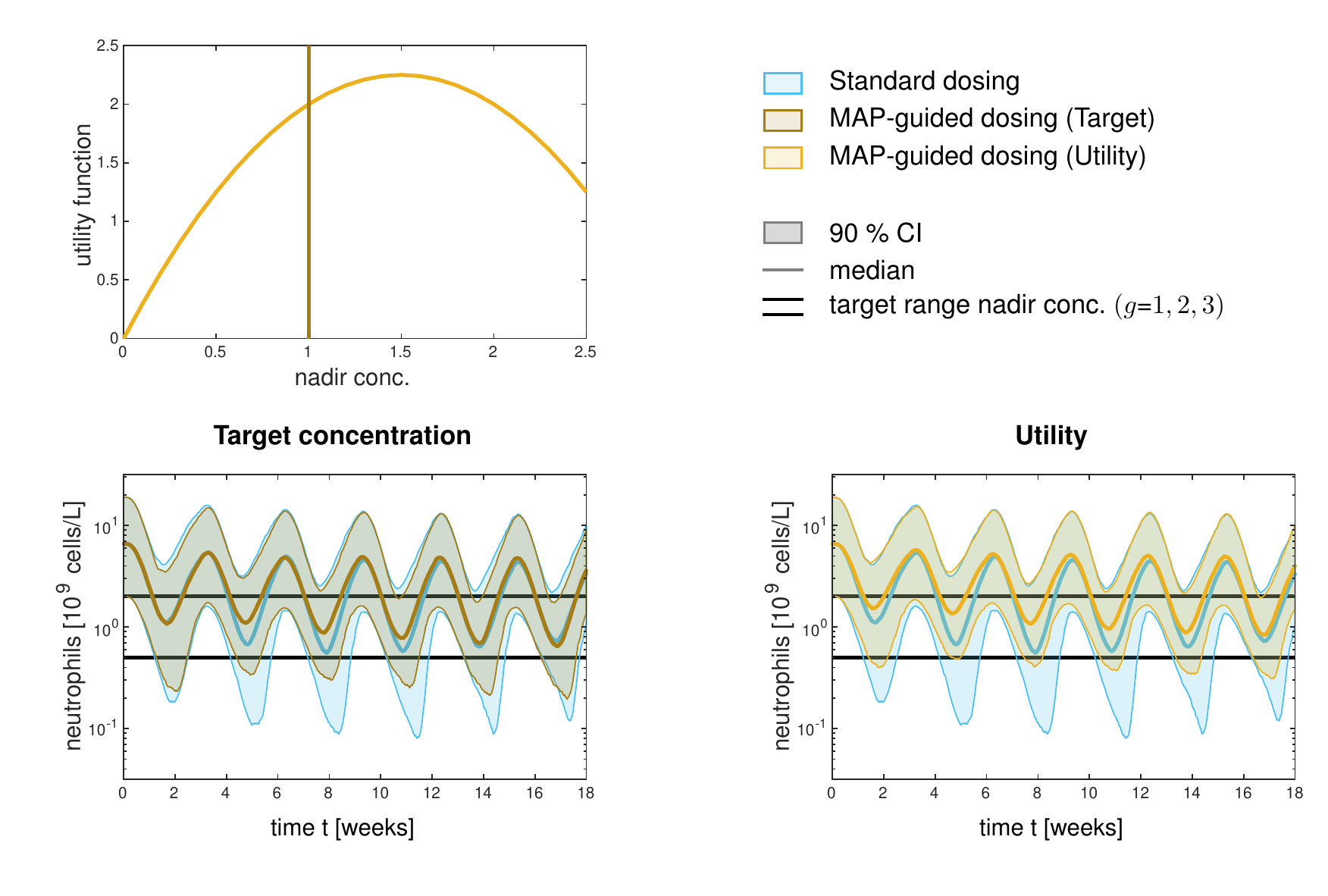}
\caption{\textbf{Comparison of different objective functions for MAP-guided dosing.} Two different objective functions were considered for MAP-guided dosing: (i) the least squared differences to a target concentration of $c_\text{nadir} = 1 \cdot 10^9 \text{cells}/ \text{L}$, (ii) a utility function which penalizes low nadir concentration (in the range of grade 4 neutropenia) higher compared to high neutrophil concentrations (in the range of grade 0 neutropenia).}
\label{fig:Utility}
\end{figure}

% -----------------------------------------------------------------------------------------------------------------------------------------------------
In literature, a utility function based on a hypothetical survival probability across the different neutropenia grades was investigated \cite{WallinPage2009}.
We proposed a utility function that was designed to mirror the essence of the reward function which we employed in RL-guided dosing to enable a fair comparison, see \SFigref{fig:Utility}.
In order to also offer a comparison to the often used concept of a target concentration, we also performed target concentration intervention with a target of $c_\text{nadir} = 1 \cdot 10^9 \text{cells}/ \text{L}$ \cite{Wallin2009}. 
For this, we minimized the squared difference, i.e., $R(s_c) = (s_c - 1)^2$, where the model state of the patient state was given by 
\begin{equation}
s_c = \underset{t \in [T_c,T_{c+1}]}{\text{min}} \ c_\text{neutrophils}\left(t;\hat{\theta}^\text{MAP},d\right)\,.
\end{equation}
For the target concentration intervention, the 90\% CI of the neutropenia time-courses of the virtual test population reached lower neutrophil concentrations compared to the utility function and increased the occurrence of grade 4 neutropenia across all cycles, see Table~S~\ref{tab:Utility} and \SFigref{fig:Utility}.
\begin{table}[H]
\centering
\begin{tabular}{c|cccccc}
cycle & 1 & 2& 3& 4& 5& 6 \\
\hline
Utility function & 10.5\% & 7.4\% &  11.4\% & 16.5\% & 21.3\% & 23.5\% \\
Target deviation &  22\% & 15.3\% & 22.3\% &29.9\% & 34.6\% & 41.5\% \\
\end{tabular}
\caption{Occurence of grade 4 neutropenia across cycles for MAP-guided dosing.}
\label{tab:Utility}
\end{table}

% -----------------------------------------------------------------------------------------------------------------------------------------------------
\subsection{DA-guided dosing}\label{sec:ResultsDA}
\begin{figure}[H]
\centering
\includegraphics[width =1\linewidth]{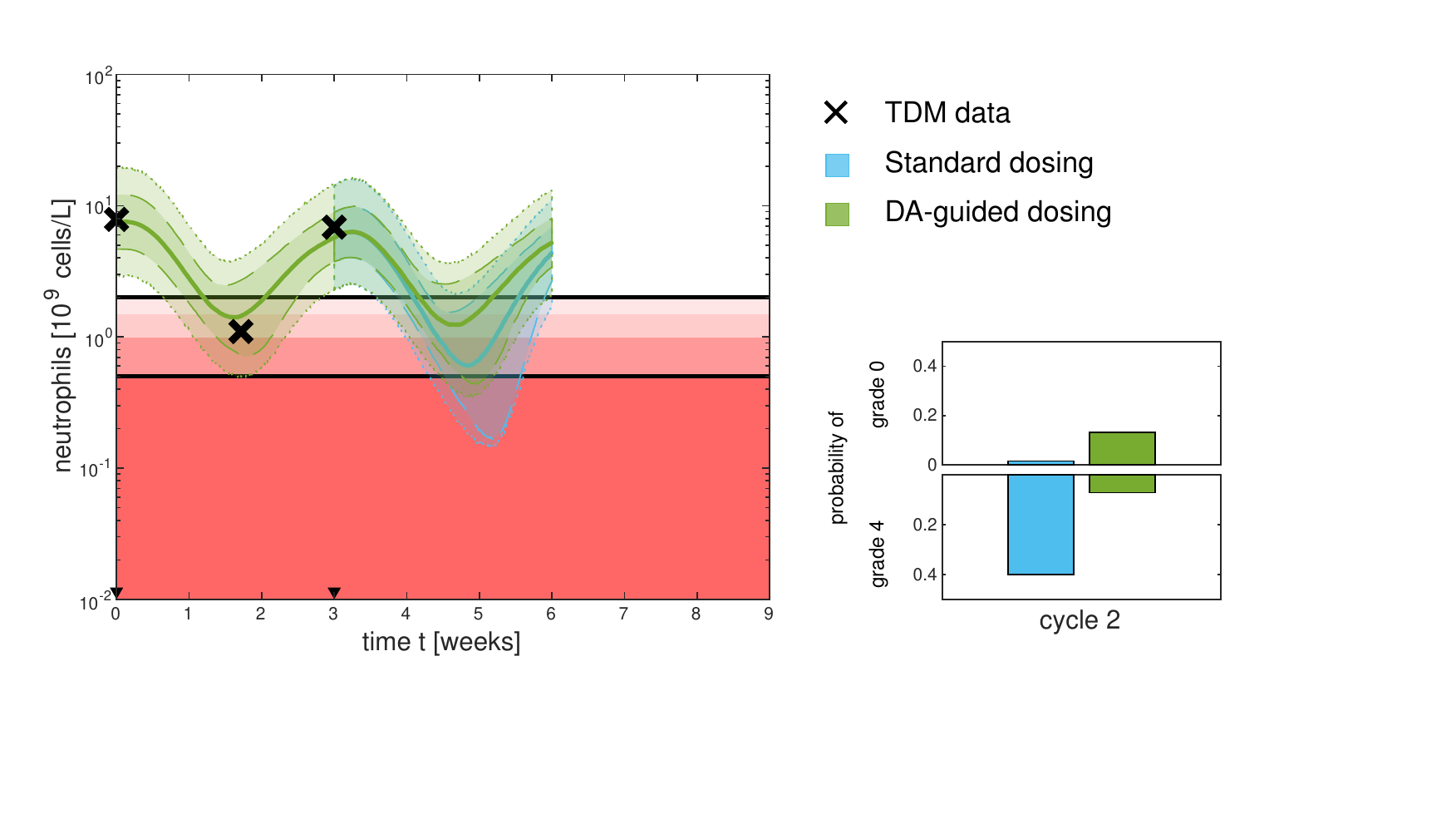}
\caption{\textbf{Exemplary dose selection for cycle 2 in DA-guided dosing.} The optimal dose in DA-guided dosing was defined to be the dose that minimizes the weighted sum of risk of grade 0 and grade 4, i.e., probability of grade 0 and grade 4 neutropenia. Note for illustration purposes a larger number of particles was chosen $M=10^3$. For comparison we also propagated the particle ensemble for the case if the standard dose was chosen (blue).}
\label{fig:DAMIPDIllustration}
\end{figure}

% -----------------------------------------------------------------------------------------------------------------------------------------------------
For the DA-guided approach, we chose the \textit{optimal dose} to be the dose that minimizes the a-posteriori probability of being outside the target range, i.e., the weighted sum of the predicted risk of the patient having neutropenia grade $g_c=0$ or $g_c=4$ in the next cycle, see Eq.~(16).
We illustrated the DA-guided dosing approach exemplarily for the second cycle dose selection for a virtual patient, see Figure~S~\ref{fig:DAMIPDIllustration}.
We solved the one-dimensional optimization problem (16) using the \texttt{fminbnd} function in MATLAB R2018b (golden section search and parabolic interpolation). 
Each objective function evaluation correspond to $M$ model simulations for the corresponding cycle. 
Therefore, we chose a rather small ensemble size $M=100$ and only considered one cycle. Note that a larger ensemble size could be chosen for the DA step, while subsequently solving the optimization problem only for a subset of the ensemble.

% -----------------------------------------------------------------------------------------------------------------------------------------------------
\subsection{RL-guided dosing}\label{sec:SimStudyRL}
% -----------------------------------------------------------------------------------------------------------------------------------------------------
For the considered patient state representation $s_c=(\text{sex},\text{age},\text{ANC}_0, g_{1:c})$, we obtained in total $L=32$ covariate classes $\mathcal{COV}_1,\ldots,\mathcal{COV}_L$ (2 genders $\times$ 4 age classes $\times$ 4 baseline neutrophil count classes), per covariate class we have $19531$ possible grade combinations for the 6 cycles, thus leading to a dimension of the discrete state space of $|\mathcal{S}| = 624992$. 
As we do not need to make a dose decision after the last cycle we can exclude the leafs of the tree (grade of last cycle), reducing the total number of states to $|\mathcal{S}| = 124992$. 

The discrete dose steps of $5\,\text{mg}/\text{BSA}$ were chosen within the range of given doses in the CEPAC-TDM study $d_\text{min} = 60 \,\text{mg}/\text{m}^2\, \text{BSA}$ and $d_\text{max} = 250\,\text{mg}/\text{m}^2\,\text{BSA}$ leading to $|\mathcal{D}|=39$.

We chose a discount factor for future rewards: $\gamma = 0.5$, see Section~S~\ref{sec:RLTuning}. 
Note that this implies that the current grade of neutropenia was higher weighted than future grades. 
Yet, $\gamma$ is sufficiently large to factor in the impact of the current dose choice on the grade of neutropenia in future cycles.

The exploration-exploitation parameter $\epsilon_c$ was chosen cycle-varying, since the expected return changes over time (cycles), due to the intermediate rewards.
Based on Hoeffding's inequality for random variables that can take values in the interval $ [a,b]$ we get,
\begin{equation}\label{eq:epsilon}
\epsilon(c) = c_{UCT} \cdot \sqrt{\sum_{k=1}^{C-c} \gamma^{k-1}\cdot (b_k-a_k)^2}\,,
\end{equation}
with respect to our chosen reward function $a_k=-2$ and $b_k=1$ for all $k$. We chose $c_{UCT}=3$.
Note that the maximum and minimum return is dependent on the current cycle as it is computed as the cumulative reward from the current cycle onwards.
The choice of all tuning parameters/reward function was further investigated in Section~S~\ref{sec:RLTuning}.

\begin{figure}[H]
\centering
\includegraphics[width =1\linewidth]{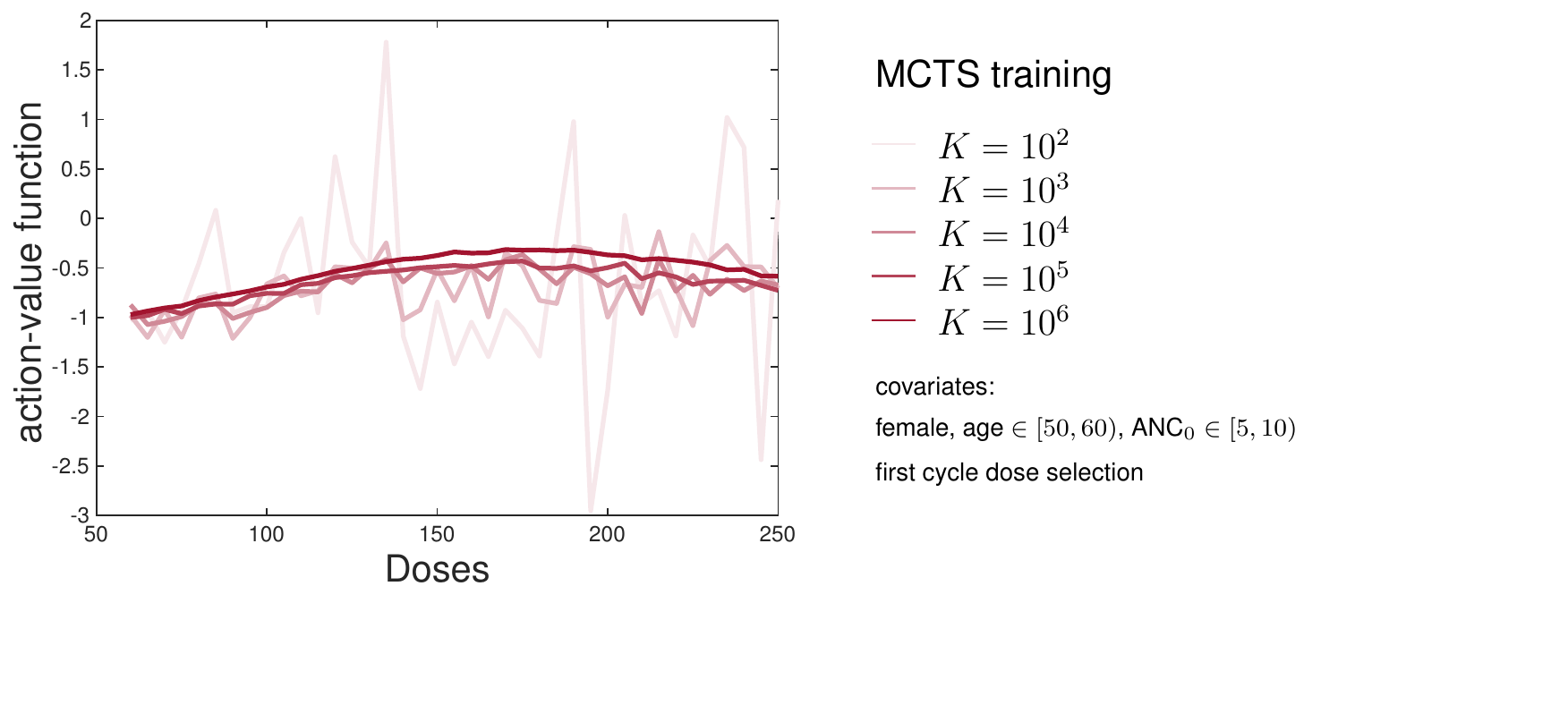}
\caption{\textbf{Training stages of Monte Carlo tree search (MCTS): Approximation of $q_\pi$}. The same virtual test patient population was dosed according to the current estimate of the action-value function $q_K$ after $K$ episodes of the planning steps for the covariate class: female, age between 50 and 60 years, and pre-treatment neutrophil counts $\text{ANC}_0 \in [5,10)$ in $[10^9 \text{cells}/\text{L}]$. }
\label{fig:MCTSTraining}
\end{figure}

\begin{figure}[H]
\centering
\includegraphics[width =1\linewidth]{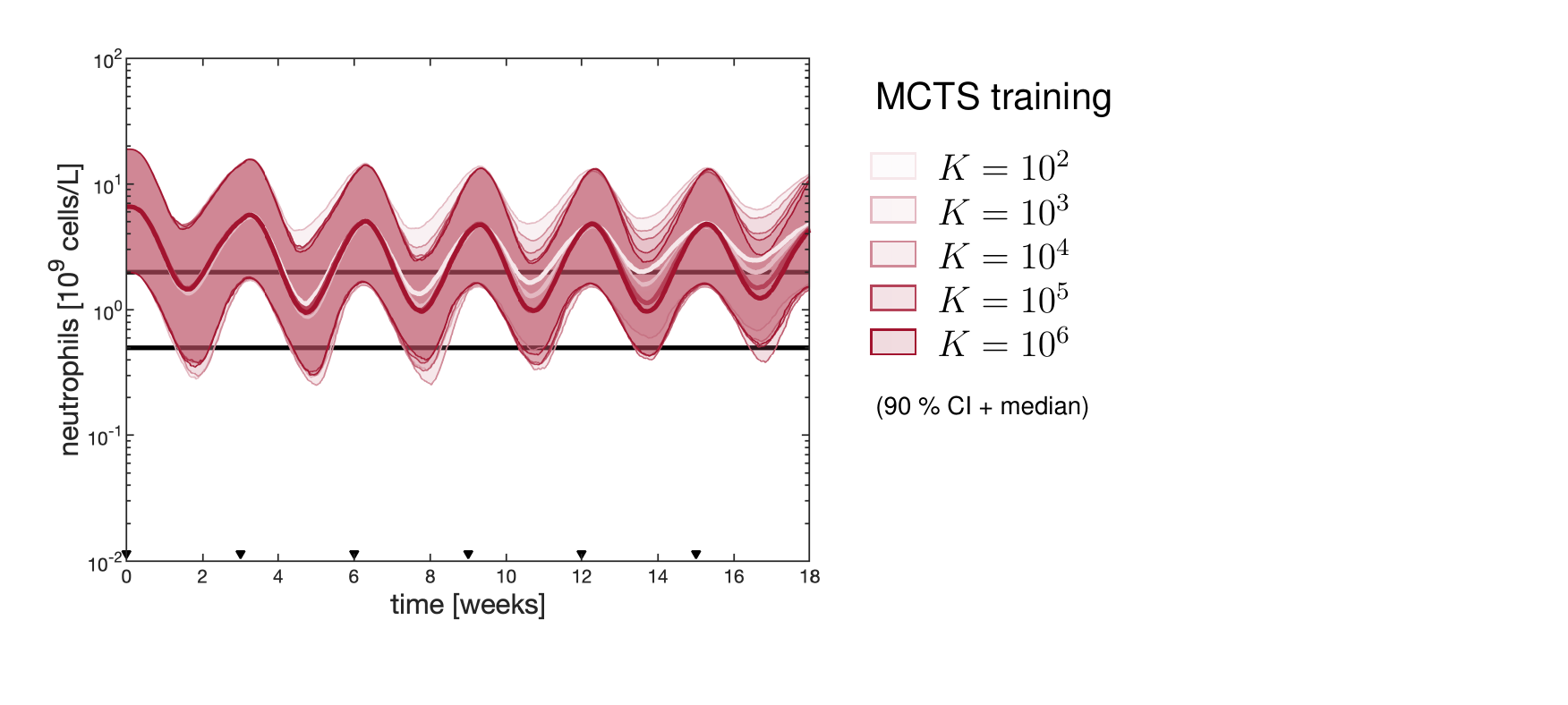}
\caption{\textbf{Training stages of Monte Carlo tree search (MCTS)}. The same virtual test patient population (as in Figure~\ref{fig:MCTSTraining}) was dosed according to the current estimate of the action-value function $q_K$ after $K$ iterations of the planning steps per covariate class. }
\label{fig:MCTSQTraining}
\end{figure}

The training phase of MCTS is visualized in Figures~S~\ref{fig:MCTSTraining} and S~\ref{fig:MCTSQTraining}.

\begin{figure}[H]
\centering
\includegraphics[width =1\linewidth]{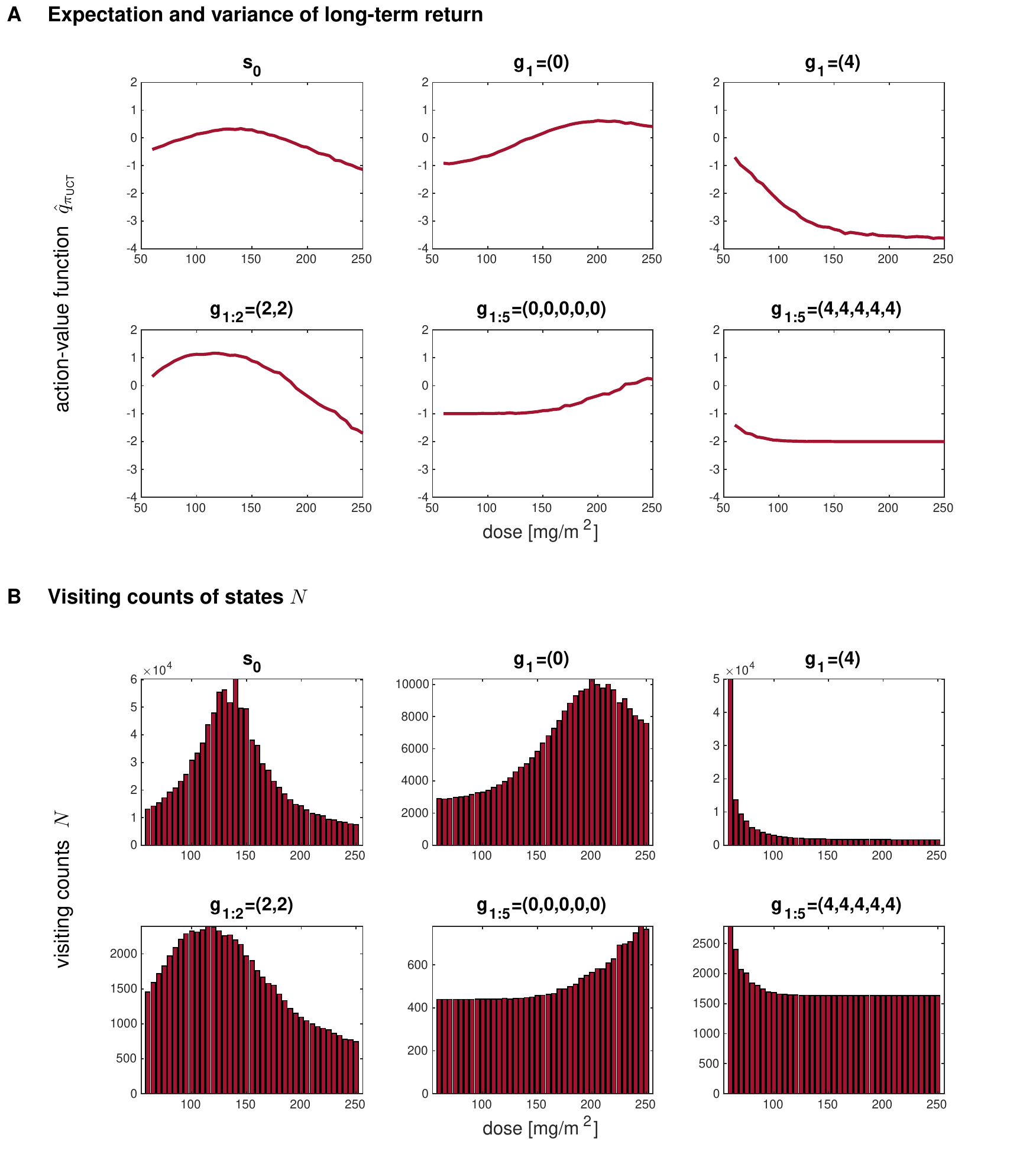}
\caption{\textbf{Monte Carlo Tree Seach}. \textbf{A} Expectation of the long-term return for exemplary states for the covariate class: male, age $\in [50,60)$, ANC0 $\in [2.5,5) \cdot 10^9 \text{cells}/\text{L}$. \textbf{B} Visiting counts of states in training phase for $K= 10^6$. }
\label{fig:Qvar_N}
\end{figure}

\begin{figure}[H]
\centering
\includegraphics[width =1\linewidth]{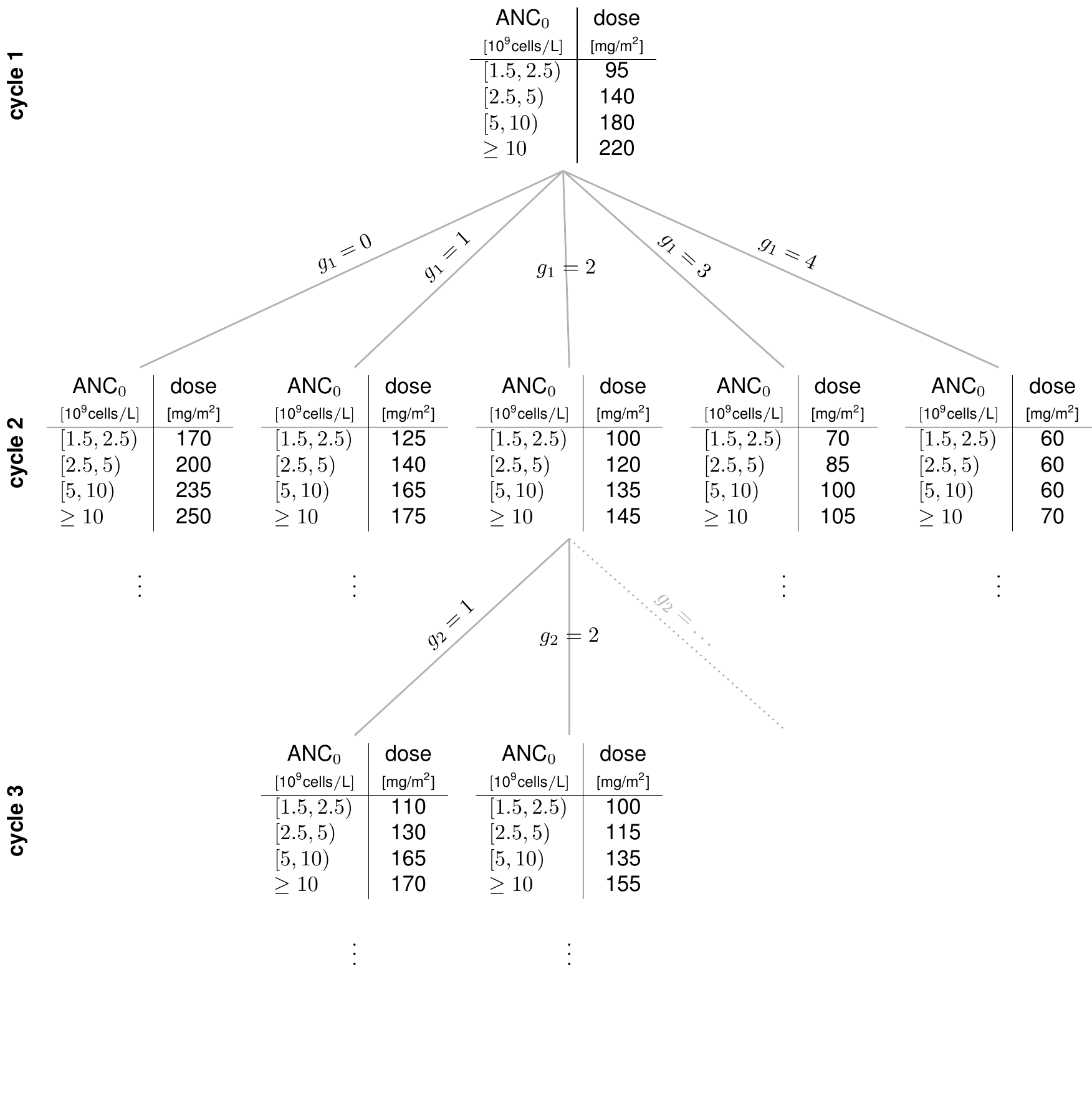}
\caption{\textbf{Illustration of the decision tree procedure for the dose selection}.  From the action-value function values $\hat{q}_\pi$ a decision tree/look-up table can be extracted. Here shown for fixed covariates: male, age $\in [50,60)$. For example, the optimal dose for the second cycle depends on the neutropenia grade of the previous cycle and the pre-treatment neutrophil count $\text{ANC}_0$. }
\label{fig:SecondCycleQ}
\end{figure}

The computed $\hat{q}_{\pi_\text{UCT}}$-Matrix can be used as a look-up table. 
For a certain patient state we need to determine the corresponding row in the matrix and then select the dose corresponding to the maximal $\hat{q}_{\pi_\text{UCT}}$-value. 
This procedure can be visualized in a diagram structure similar to the one developed by Joerger et al. \cite{Joerger2012a}, see \SFigref{fig:SecondCycleQ}.
Since RL allows to deal with a large amount of information regarding patient state/dose combinations, we just depict a small subtree. 

% -----------------------------------------------------------------------------------------------------------------------------------------------------
\subsubsection{Investigating the effect of the tuning parameters}\label{sec:RLTuning}

\begin{figure}[bt]
\centering
\includegraphics[width =1\linewidth]{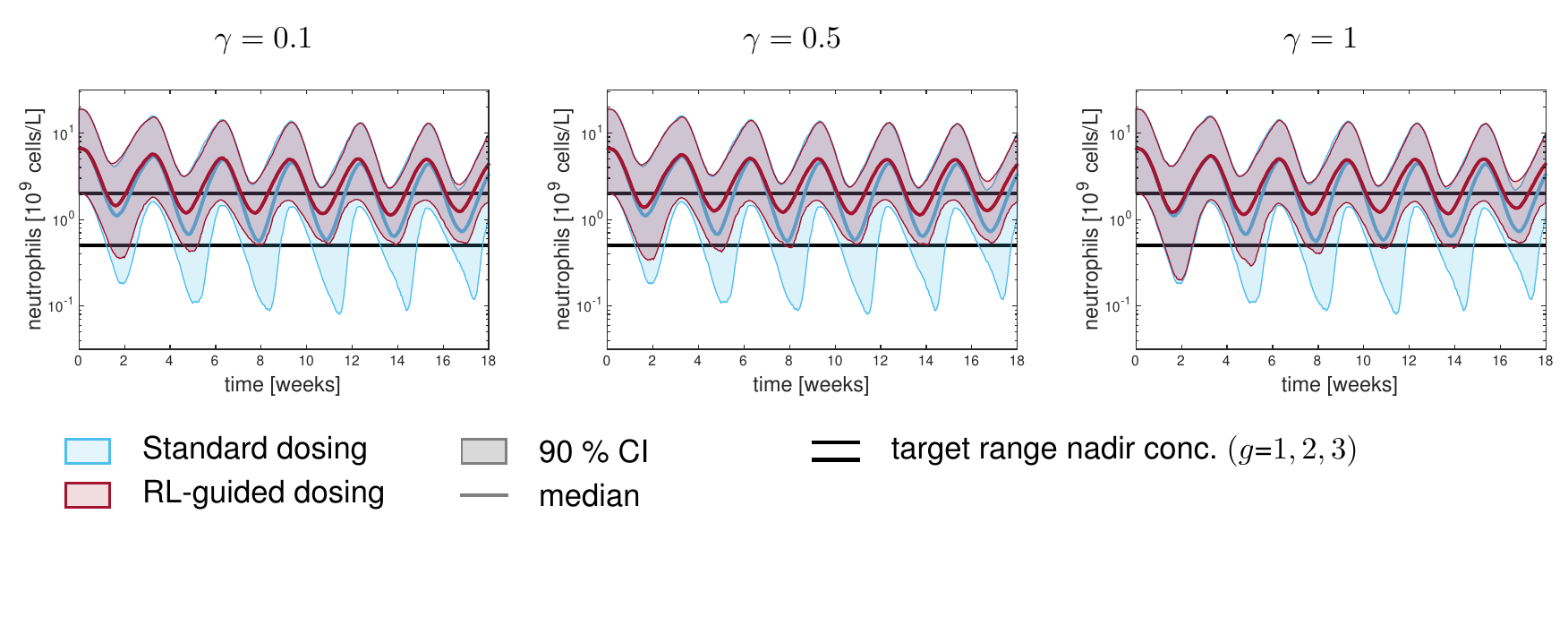}
\caption{\textbf{Choice of the discount parameter $\gamma$}. The discount parameter $\gamma \in [0,1]$ weighs the relation between short term and long-term goals. Since no discount is applied to the current cycle, small $\gamma$ values give the current cycle a much higher weighting compared to later cycles, i.e., prioritizing the short term return rather than the long-term outcome. On the contrary, large $\gamma$ values put more weight on the long-term goals. Note, this analysis was done with sampling time points day 0 \& 12 and a virtual test population of $N=1000$.  }
\label{fig:RLgamma}
\end{figure}

We investigated the choice of the discount parameter $\gamma \in [0,1]$.
In our setting, the long-term goal (median survival) is already included in the immediate reward, since neutropenia grade 0 was also evaluated with $-1$ in the reward function.
Therefore, $\gamma$  does not have such a strong impact on the results, see Figure~S~\ref{fig:RLgamma}. 
The parameter is expected to be more relevant if efficacy is not 'measured' by the surrogate marker of neutropenia, but rather evaluated based on a tumour growth model or a survival model. This would result in rewards with larger time lag since the choice of a dose impacts tumour growth and in particular survivial only at (much) later time points. 

\begin{figure}[H]
\centering
\includegraphics[width =1\linewidth]{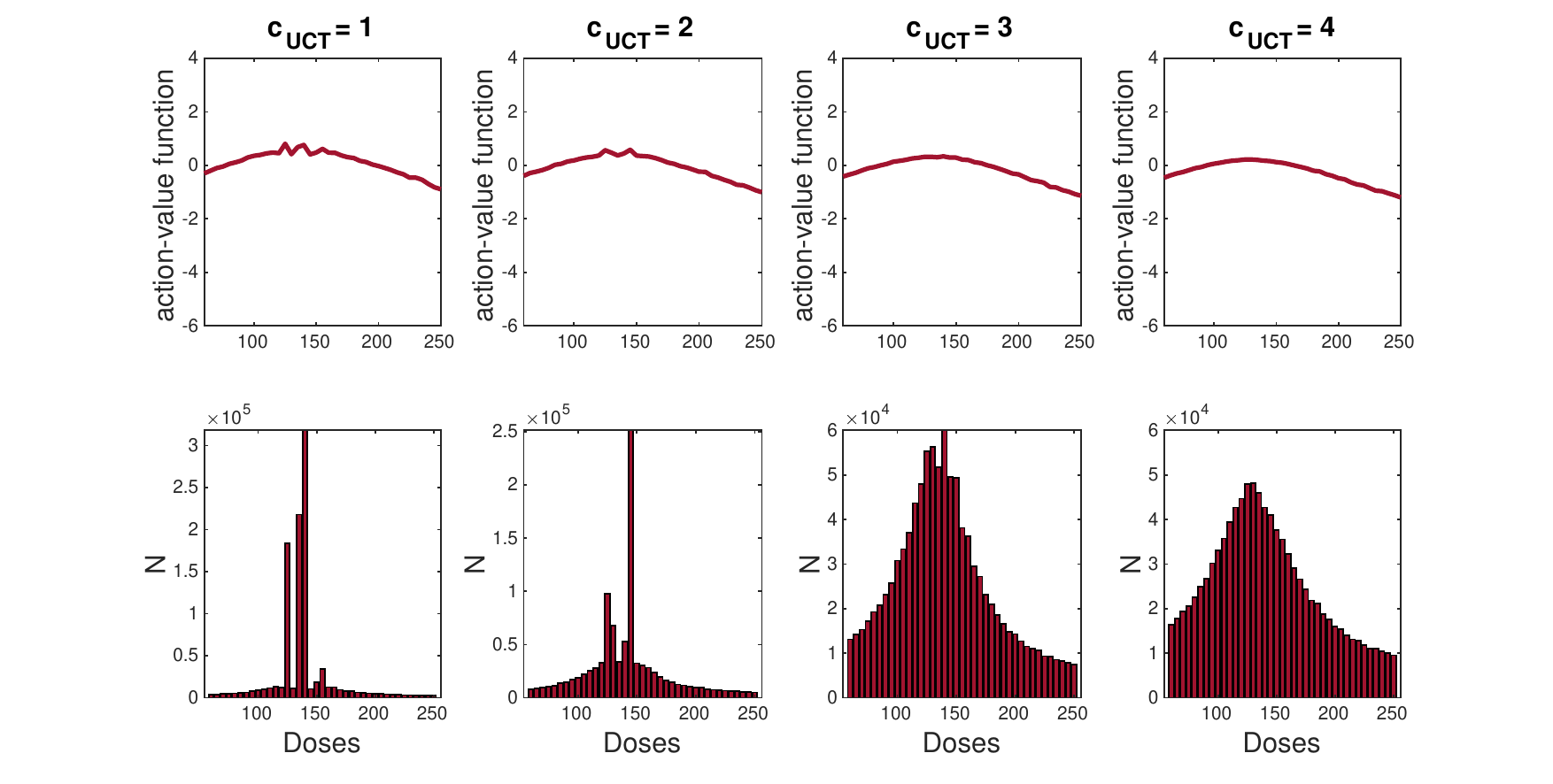}
\caption{\textbf{Monte Carlo tree search: exploration/exploitation parameter}.  Expectation and variance of the long-term return for exemplary states of the covariate class: male,age $\in [50,60)$, ANC0 $\in [2.5,5)$ (top panel) and visiting counts of dose selections in the initial state $s_0$, $N(s_0,d)$ in training phase for $K= 10^6$ (bottom panel) for different exploration/exploitation parameter $c_\text{UCT}$ in eq.~\eqref{eq:epsilon}.}
\label{fig:Qvar_N123}
\end{figure}

We further examined the trade-off between exploration and exploitation. 
For this, we varied the constant $c_\text{UCT}$ in eq.~\eqref{eq:epsilon} in the MCTS with UCT approach. 
We found that for smaller $c_\text{UCT}$ values, the algorithm selected only a small number of doses with relative high probability, see e.g., Figure~S~\ref{fig:Qvar_N123} for the initial dose selection. 
This led to an un-smooth action-value function, which is not expected in the considered scenario. Therefore, we chose $c_\text{UCT}=3$, as this choice showed a balanced exploration of the dose space while still prioritizing doses with high expected return.

\begin{figure}[H]
\centering
\includegraphics[width =1\linewidth]{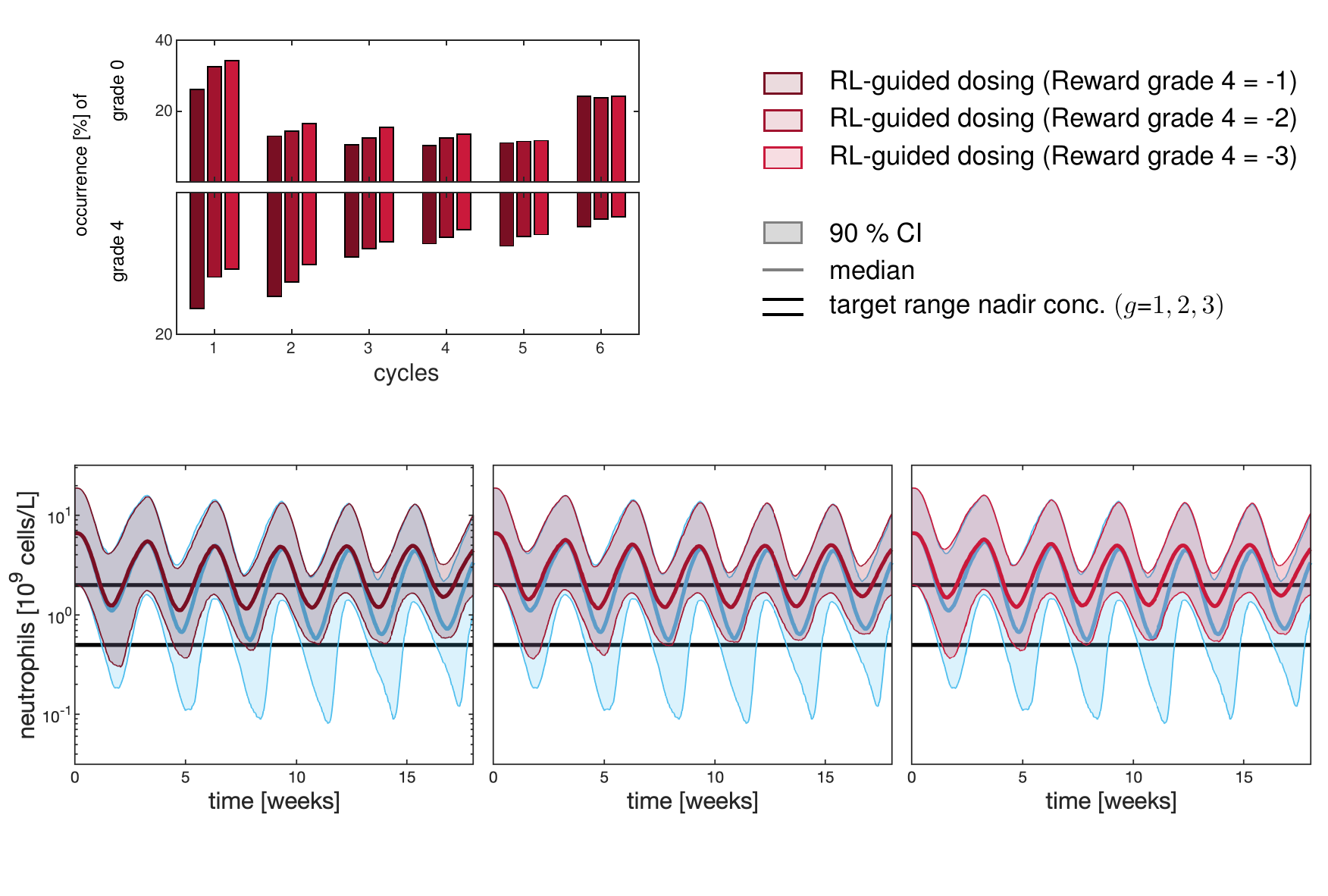}
\caption{\textbf{Comparison of RL-guided dosing results for changes in the reward function.} In the three scenarios the rewards for grades 0-3 remained the same and only the reward for grade 4 was changed. In the first panel the reward value of grade 4 was set to -1, thus equal to the reward of grade 0. The second reward function corresponds to the scenario presented in the main manuscript (-2) and in the last scenario a larger penalty is put on grade 4 in comparison to grade 0, i.e., R=-3 if grade 4 was observed. }
\label{fig:RL_Reward}
\end{figure}

Finally, we investigated the effect of changes in the reward functions. 
For this, we exemplary changed the reward---here corresponding to a penalization---of grade 4 neutropenia. 
In the first scenario, the reward of grade 4 neutropenia was set equal to the reward of grade 0 neutropenia ($R_{c+1}=-1\,, \text{if} \ g_c=4$). 
Thus, subtherapeutic and toxic ranges result in the same reward value.
The second scenario, $R_{c+1}=-2\,, \text{if} \ g_c=4$ corresponds to the scenario presented in the main manuscript. 
In the the third scenario, neutropenia grade 4 was even more strongly penalized, reflecting the potential of  exposing patients to immediate life-threatening conditions ($R_{c+1}=-3\,, \text{if} \ g_c=4$). 
As expected, the occurrence of grade 4 decreased the stronger grade 4 neutropenia was penalized, see Figure~S~\ref{fig:RL_Reward}. Due to the uncertainty, at the same time the incidence of grade 0 is increased. 
Thus, it is crucial to have a clear therapeutic goal prior to defining the evaluation function. 
The choice of the evaluation function should be examined in comparison with potential alternatives, as in Figure~S~\ref{fig:RL_Reward} and the results should be compared with the desired therapeutic outcome.

% -----------------------------------------------------------------------------------------------------------------------------------------------------
\subsubsection{Q-planning}

As an alternative to Monte Carlo Tree Search, Q planning can be performed to learn the action-value function.
We employed the same state representation and the same reward function as for the MCTS approach.

\begin{figure}[H]
\centering
\includegraphics[width =1\linewidth]{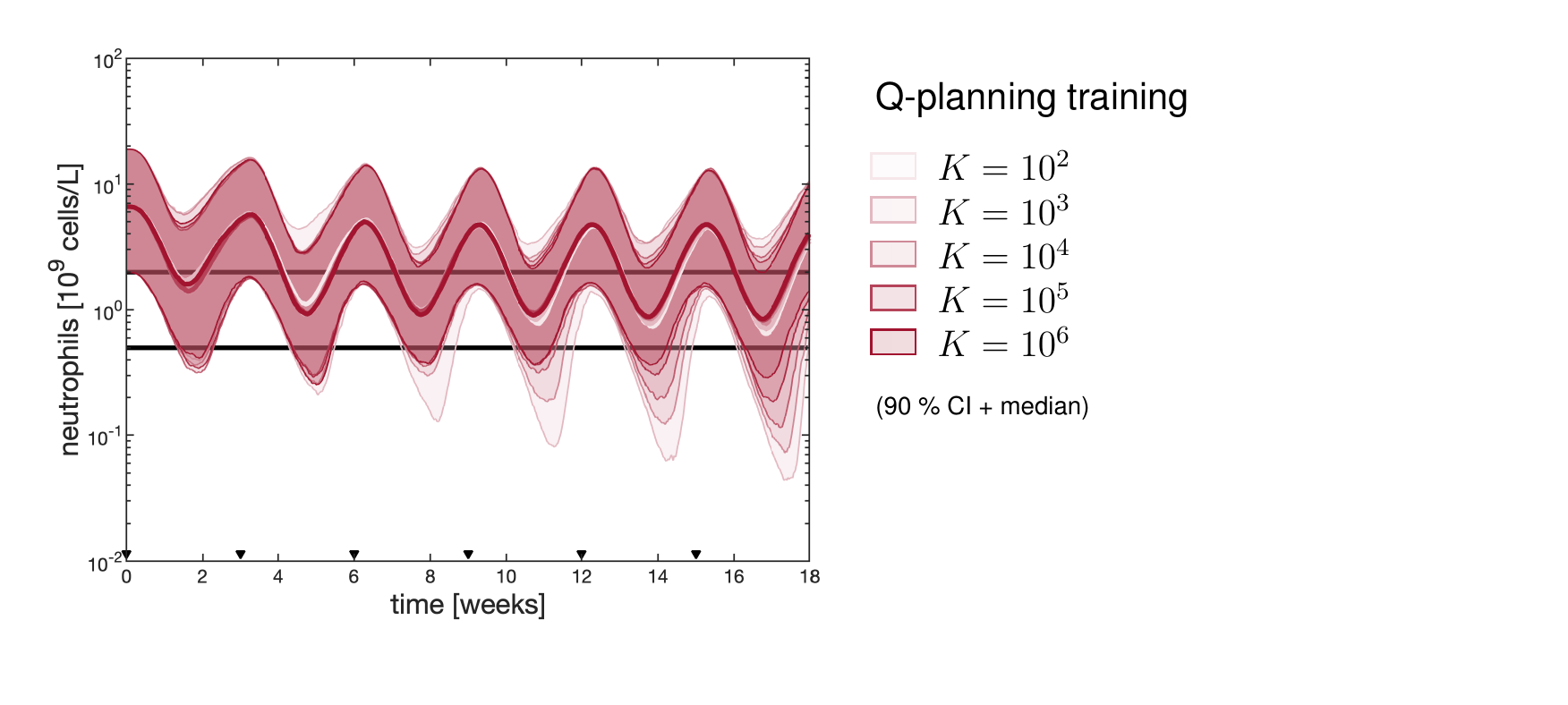}
\caption{\textbf{Training stages of Q-planning}. The same virtual test patient population was dosed according to the current estimate of the action-value function $q_K$ after $K$ iterations of the planning steps per covariate class.}
\label{fig:QplanTraining}
\end{figure}

We also visualized the training phase for Q-planning, see Figure~S~\ref{fig:QplanTraining}.
For this specific example and selected patient state representation, the results using MCTS are more promising and could better reduce the incidence of grade 0 \& 4 neutropenia in later cycles.

% -----------------------------------------------------------------------------------------------------------------------------------------------------
\subsection{DA-RL-guided dosing}
% -----------------------------------------------------------------------------------------------------------------------------------------------------

% -----------------------------------------------------------------------------------------------------------------------------------------------------
\subsubsection{Different approaches to estimate the grade of neutropenia $\hat{g}_c$ in cycle $c$}
\begin{figure}[H]
\centering
\includegraphics[width =1\linewidth]{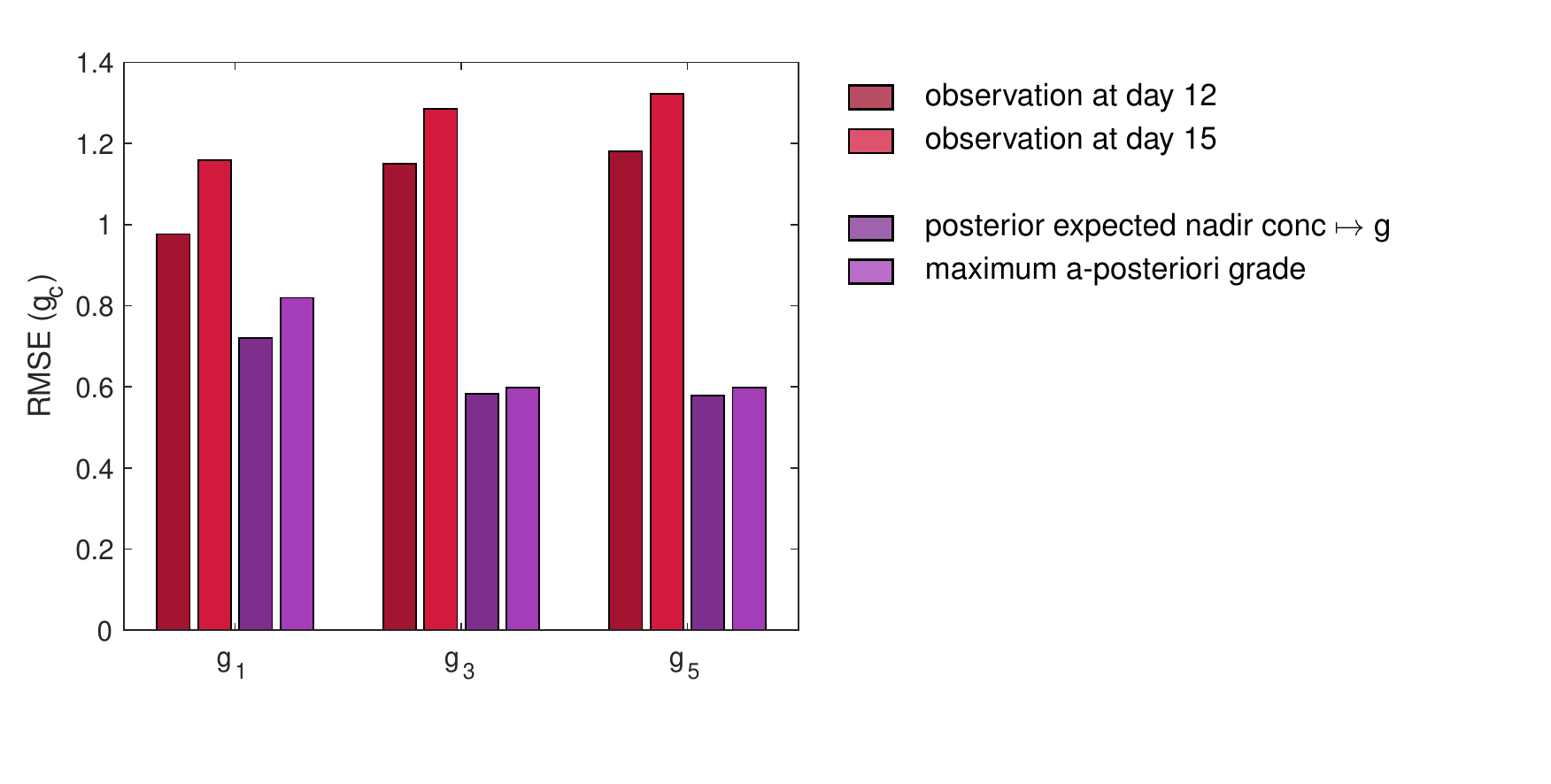}
\caption{\textbf{RMSE of estimating the grade of neutropenia $\hat{g}_c$ in cycle $c$ using different approaches.} In RL-guided dosing, the neutrophil measurement is used to infer the neutropenia grade. We investigated the two sampling time points day 12 (typical nadir time) and day 15 (as in the CEPAC-TMD study). For DA-RL-guided dosing, the particle ensemble $\mathcal{E}_{1:c}$ can be used to infer an improved patient state. There are two options: (i) the posterior expected nadir concentration is computed and then translated into a discrete grade; or (ii) the probability of each grade is determined by summing the weights of particle giving raise to that grade;  then, the maximum a-posteriori grade is defined as the grade with highest sum of weights.}
\label{fig:ComparisonStateEst}
\end{figure}

In DA-RL-guided dosing, the particle ensemble $\mathcal{E}_{1:c}$ is used to estimate the patient state more reliably than just using the observed neutrophil concentration at day 12 or 15. Figure~S~\ref{fig:ComparisonStateEst} shows the root means squared error (RMSE) between the estimated neutropenia grade $\hat{g}_c$ and the \textit{true} grade $g_c$ from the underlying ``truth" used to simulate the data. Note, that we neglected the `$\hat{\hphantom{g}}$' in the main manuscript for ease of notation.
Overall, the RMSE is lower for day 12 than for day 15. Moreover, using a model-based state representation reduced the RMSE substantially---and much more than the difference between day 12 and 15.
We further compared the posterior expected nadir concentration, see Eq.~(S~\ref{eq:postexp}), translated into discrete grades, with first computing the probabilities of the different grades and then using the maximum a-posteriori grade, i.e., the grade with highest sum of weight.
The posterior expected nadir concentration performed slightly better and was therefore used in the main manuscript for approximating the patient state using the particle ensemble $\mathcal{E}_{1:c}$.

% -----------------------------------------------------------------------------------------------------------------------------------------------------
\subsubsection{RL-guided dosing based on DA state}
\begin{figure}[H]
\centering
\includegraphics[width =1\linewidth]{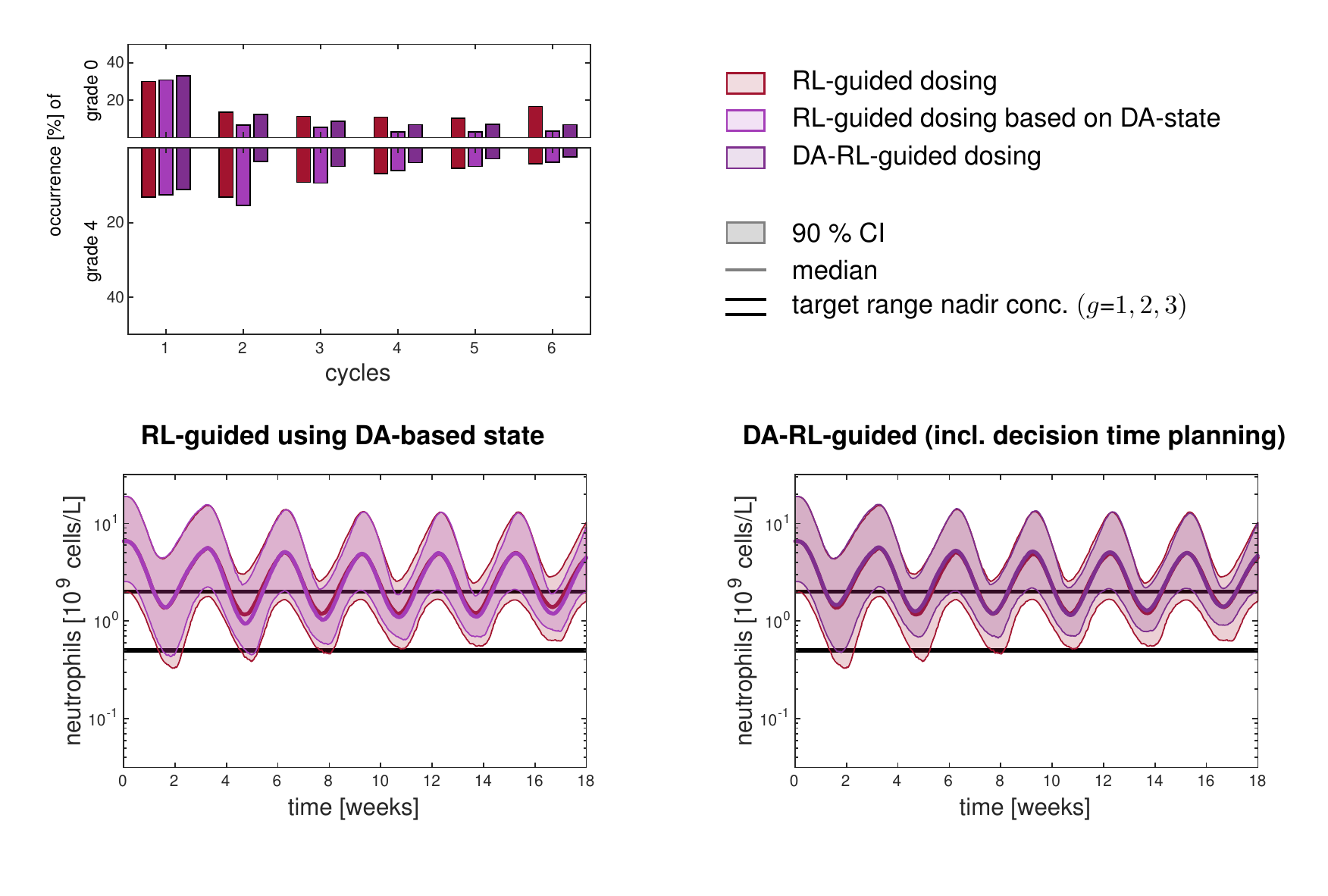}
\caption{\textbf{RL-guided dosing based on the DA-based model state of the patient}. The virtual test population was dosed with RL-guided dosing adding the different aspects of DA. Lower left panel: RL-guided dosing using the improved DA-based model state of the patient. Here we used the smoothed posterior expected nadir concentration translated to the discrete neutropenia grades. Lower right panel: DA-RL-guided dosing as presented in the main manuscript. Note, that we used the scenario with sampling time points day 0 \& 12 for this analysis.}
\label{fig:DA1MCTS}
\end{figure}

In the main manuscript, we discussed that DA can be used in two ways to improve RL-guided dosing: (i) providing an improved state estimate (as in the previous section); and (ii) by using the posterior particle ensemble $\mathcal{E}_{1:c}$ to update the $\hat{q}_{\pi_\text{UCT}}$ values in relevant and promising dose-state-pairs.
In Figure~S~\ref{fig:DA1MCTS}, we investigate the scenario (i) alone, i.e., if we only use the improved state estimate in RL-guided dosing (without decision time planning based on the posterior particle ensemble).
We observed a one-sided improvement, only the occurrence of grade 0 was reduced compared to RL alone. This indicates again the key role of individualized uncertainties for MIPD. 
In short: if the quality of estimating the grade of neutropenia is improved, also the corresponding dosing table should be updated,since the RL dosing table accounted for the potential ``bias" in the state estimation.
If not, improved estimates are used in decision trees that have been determined based on the less accurate estimated of the grade of neutropenia. 
Such a mismatch should be avoided. 

% -----------------------------------------------------------------------------------------------------------------------------------------------------
\subsubsection{PUCT algorithm}
In the PUCT algorithm, the pre-calculated action-value function values $\hat{q}_{\pi_\text{UCT}}$ have to be translated to probabilities.
As described in the main text, we used the Boltzmann distribution (see Eq.~(17) in the main text) to convert the expectation values in $\mathbb{R}$ to probabilities  in $[0,1]$. 
In addition, we performed a kernel density estimation to further smooth the function in case of a rough action-value function due to small visiting counts (this step is more relevant if less pre-training steps were possible, e.g., in larger state spaces), compare small $K$ values (rough) to large $K$ values in Figure~\ref{fig:MCTSTraining}.

\begin{figure}[H]
\centering
\includegraphics[width =1\linewidth]{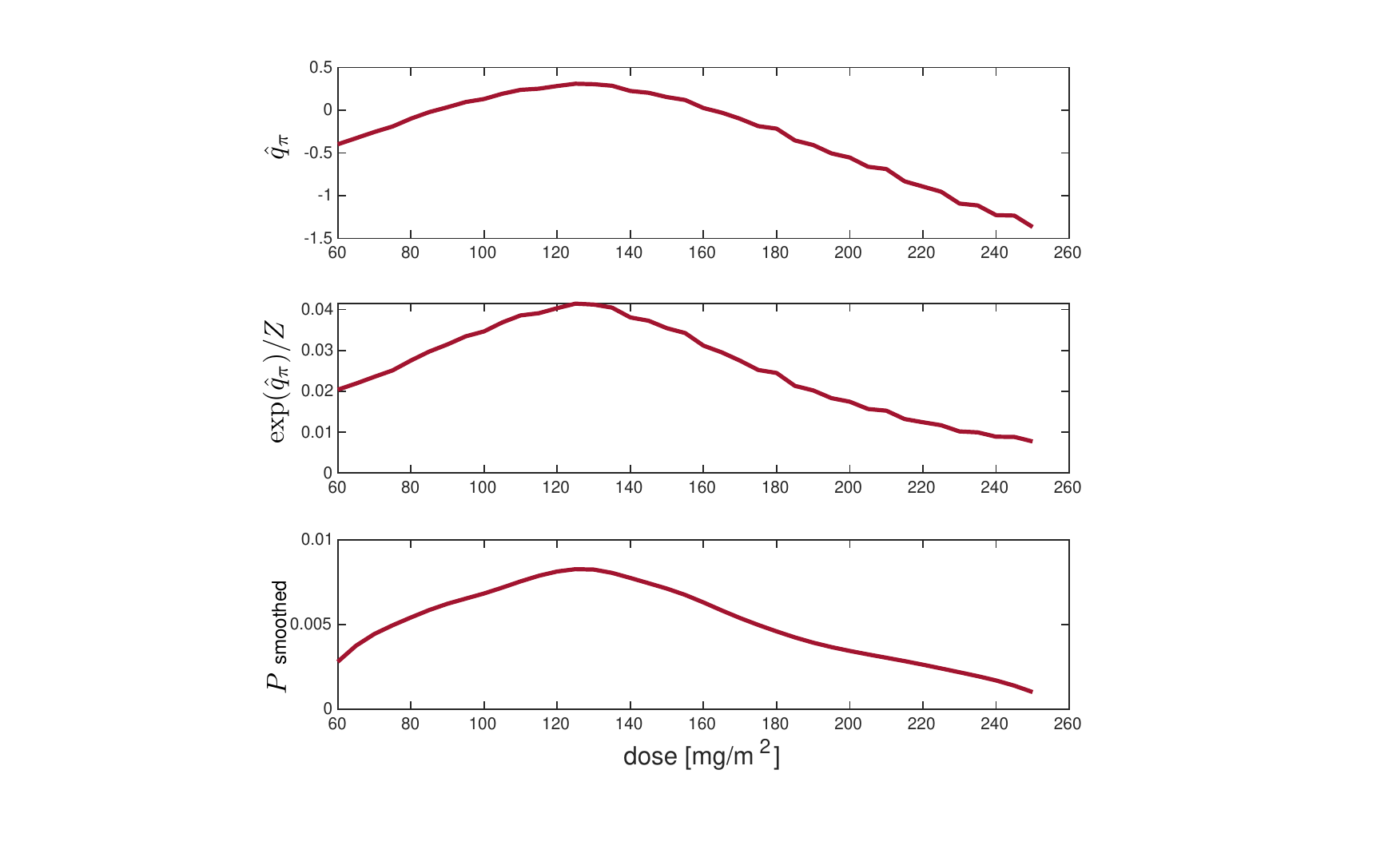}
\caption{\textbf{Prior probabilities for DA-RL-guided dosing}. The action-value function $\hat{q}_\pi$ was translated into probabilities $P=\exp(\hat{q}_\pi)/Z$, where $Z=\sum_d \exp(\hat{q}_\pi(s,d))$ using the Boltzmann function, and subsequently smoothed (P smooth) using kernel density estimation. The smoothed probabilities were used in Algorithm~1 to guide the dose selection in the individual search tree. }
\label{fig:PriorProbs}
\end{figure}

\begin{figure}[H]
\centering
\includegraphics[width =1\linewidth]{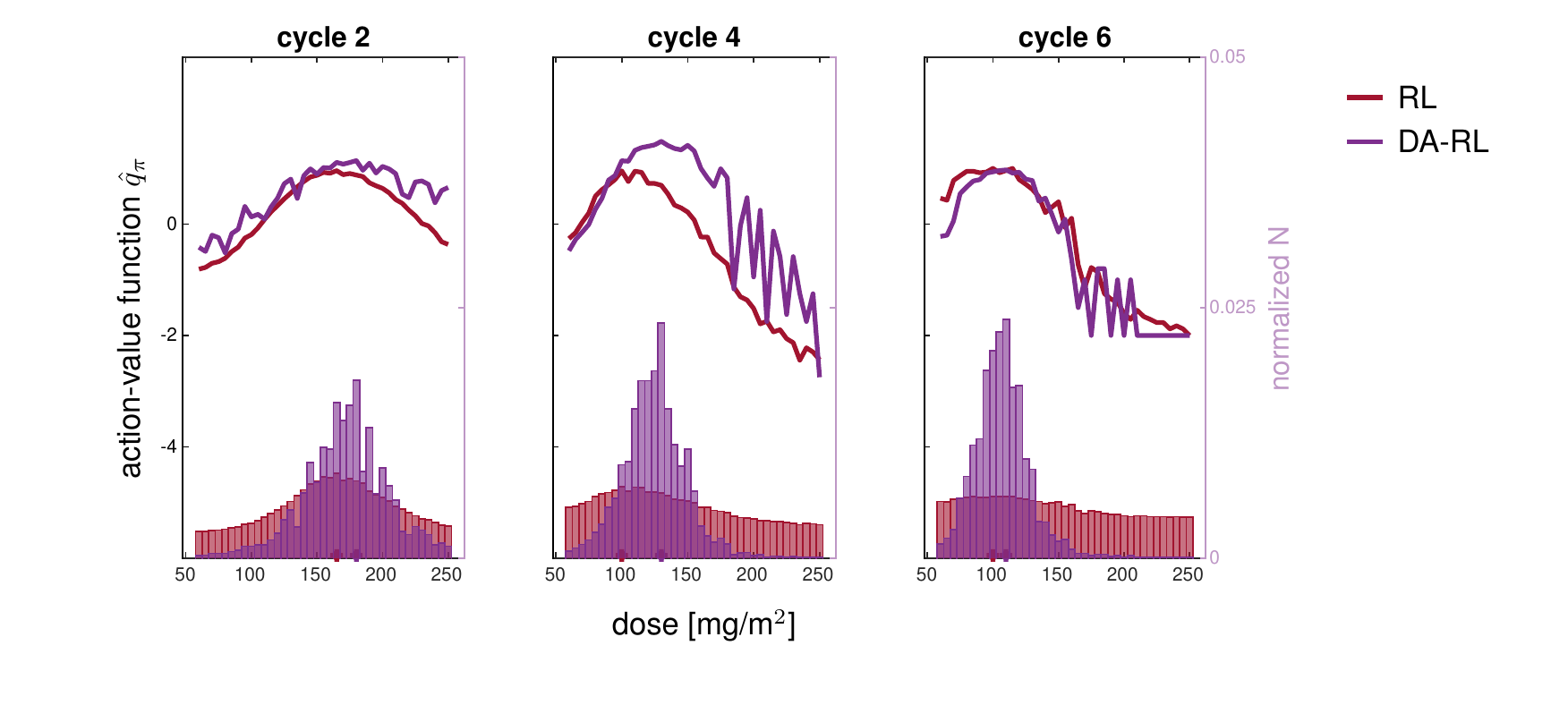}
\caption{\textbf{Comparison of the action-value function values $\hat{q}_\pi$ and visiting counts $N$ (normalized) for RL-guided dosing and DA-RL-guided dosing}. The individual action-value function values (DA-RL) differ from the population values (RL). Doses are chosen more frequently in regions where the prior probabilities, defined by the population action-value (red line), are large, see histogram (right axis). Note that it can be observed that the $\hat{q}_\pi$ values are smoother in regions with high number of samples (e.g. cycle 2 for lower doses). Regions with low prior probabilities (computed from RL $\hat{q}_\pi$ values(red line)) are chosen less often. Overall, this results in improved sampling of important regions. 
The visiting counts were divided by the maximum number of visits for one dose (i.e. scaled to one, $N/max(N)$) to allow for comparison since $K_\text{RL}>>K_\text{DA-RL}$.
}
\label{fig:Q_MCTS_DAMCTS}
\end{figure}

As a result of the updated uncertainties, the action-value function $\hat{q}_{\pi_\text{PUCT}}$ (DA-RL-guided dosing) differ from the static $\hat{q}_{\pi_\text{UCT}}$ 
(RL-guided dosing), see Figure~S~\ref{fig:Q_MCTS_DAMCTS}. This also led to different optimal doses (markers at the x axis). The purple bars show the visiting counts $N$ of the different doses in the given state, showing that doses are chosen more often which have high $\hat{q}_{\pi_\text{UCT}}$ (red line) as enforced via the PUCT algorithm. 
It can be also seen that the $\hat{q}_{\pi_\text{PUCT}}$ -curve (purple line) is not very smooth in dose regions which have low $\hat{q}_{\pi_\text{UCT}}$ values as these values are not chosen often.
In PUCT, the search focused more on promising regions of the dose space.
In practical applications, deviations from this highly focused search need to be discussed depending on how much one wants to trust the prior knowledge or how much we expect the new patients to deviate.

% -----------------------------------------------------------------------------------------------------------------------------------------------------
\subsection{Comparison across all considered evaluation functions}
% -----------------------------------------------------------------------------------------------------------------------------------------------------
The different methods towards the optimal dose problem considered in the manuscript are based on different evaluation/reward function.
For a more in-depth comparison, we also show the results of the different methods with respect to all considered evaluation functions: the utility (MAP-guided dosing), deviation from target concentration (MAP-guided dosing), the weighted sum of occurrence of grade 0/4 (DA-guided dosing), and the total reward (RL-guided dosing).

\begin{figure}[H]
\centering
\includegraphics[width =1\linewidth]{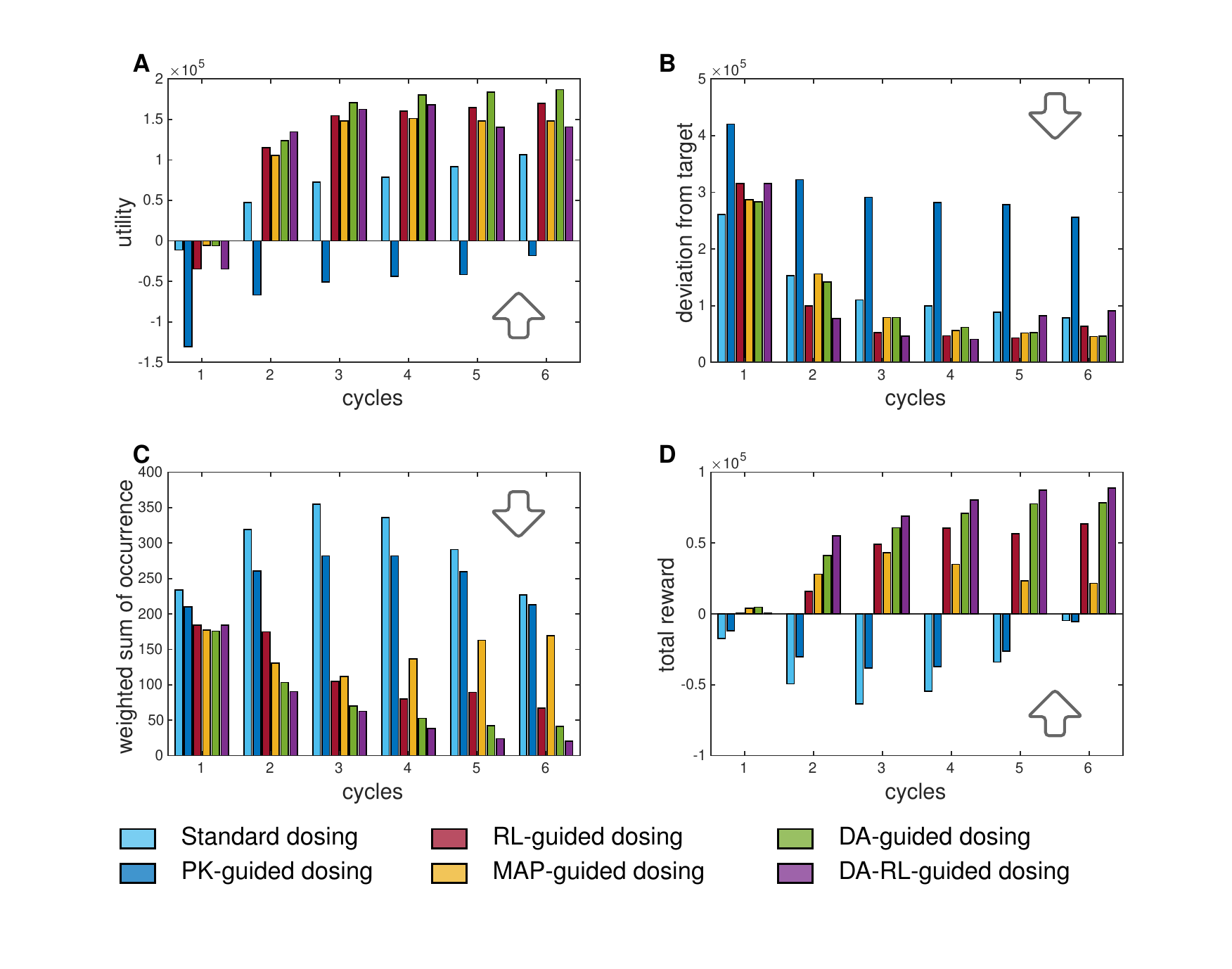}
\caption{\textbf{Comparison of methods across all evaluation functions}. \textbf{(A)} Comparison of the utility used in the MAP-guided dosing for all dosing strategies, see Figure~S~\ref{fig:Utility}. The higher the utility the better (upward arrow). \textbf{(B)} The deviation from a target concentration ($1\cdot 10^9 \cdot \text{cells}/\text{L}$) was proposed for MAP-guided dosing (target concentration intervention). The smaller the deviation from the target the better (downward arrow). \textbf{(C)} The weighted sum of occurrence was minimized in the DA-guided approach. The occurrence of grade 4 was penalized more strongly than the occurrence of grade 0. The smaller the weighted sum the better (downward arrow).\textbf{(D)} The total reward as defined for the RL-guided dosing is to be maximized. The higher the total reward the better (upward arrow).}
\label{fig:AllEvalFunc}
\end{figure}

\newpage
\bibliographystyle{cptpspref}
\bibliography{RL_project}